\documentclass{article}

\usepackage[dvipsnames]{xcolor}
\usepackage[nonatbib,final]{neurips_2020}
\usepackage{times}
\usepackage{epsfig}
\usepackage{graphicx}
\usepackage{amsmath}
\usepackage{amssymb}
\usepackage{newtxtt} 
\usepackage{longtable}
\usepackage{pgfplots}

\usepackage{algorithm,algorithmic} 
\usepackage{algorithmic,eqparbox,array}

\usepackage{booktabs} 
\usepackage{wrapfig} 
\usepackage{subcaption}
\usepackage{placeins} 
\newcommand{\pal}{\textit{PAL}}
\usepackage{amsthm}
 \usepackage{setspace}
\newcommand{\argmin}[1]{\underset{#1}{\operatorname{arg}\,\operatorname{min}}\;}
\newcommand{\xn}[1]{\mathbf{x}_n}
\newcommand{\xnt}[1]{\mathbf{x}^T_n}
\newcommand{\He}[1]{\mathbf{H}}

\usepackage{stfloats}
\usepackage{longtable}
\usepackage{listings}
\usepackage{fancyhdr}

\usepackage{nicefrac}       
\usepackage{microtype}      
\usepackage{cite}
\usepackage{multicol} 
\usepackage{wrapfig}
\usepackage{xargs} 
\usepackage[breaklinks=true,bookmarks=false]{hyperref}
\pgfplotsset{compat=1.3}
\usepgfplotslibrary{external} 
\tikzsetexternalprefix{tikz_figures/}
{neurips_2020}
\tikzsetfigurename{pal}

\begin{document}
\newcommandx{\BL}[1][1=t]{\mathcal{L}_{\mathbb{B}_{#1}}}
\newcommandx{\s}[1][1=\space]{s_{\text{upd}_{#1}}}
\renewcommandx{\vec}{\boldsymbol}
\renewcommandx{\mathbf}{\boldsymbol}

\title{Parabolic Approximation Line Search for DNNs}

%

\author{Maximus Mutschler and Andreas Zell\\
	University of T\"ubingen\\
	Sand 1, D-72076 T\"ubingen, Germany\\
	{\tt\small \{maximus.mutschler, andreas.zell\}@uni-tuebingen.de}
}

\maketitle
%

\begin{abstract}
	 A major challenge in current optimization research for deep learning is automatically finding optimal step sizes for each update step. The optimal step size is closely related to the shape of the loss in the update step direction. However, this shape has not yet been examined in detail. This work shows empirically that the mini-batch loss along lines in negative gradient direction is locally mostly convex and well suited for one-dimensional parabolic approximations. We introduce a simple and robust line search approach by exploiting this parabolic observation, which performs loss-shape-dependent update steps. Our approach combines well-known methods such as parabolic approximation, line search, and conjugate gradient to perform efficiently. It surpasses other step size estimating methods and competes with standard optimization methods on a large variety of experiments without the need for hand-designed step size schedules. Thus, it is of interest for objectives where step-size schedules are unknown or do not perform well. Our extensive evaluation includes multiple comprehensive hyperparameter grid searches on several datasets and architectures. Finally, we provide a general investigation of exact line searches in the context of batch losses and exact losses, including their relation to our line search approach.
\end{abstract}

\section{Introduction}
  Automatic determination of optimal step sizes for each update step of stochastic gradient descent is a major challenge in current optimization research for deep learning \cite{L4,L4_alternative,probabilisticLineSearch,gradientOnlyLineSearch,backtracking_line_search,backtracking_line_search_NIPS,BigbatchLineSearch,hypergradientdescent,ratesautomatic}. One default approach to tackle this challenge is to apply line search methods. Several of these have been introduced for Deep Learning \cite{probabilisticLineSearch,gradientOnlyLineSearch,backtracking_line_search,backtracking_line_search_NIPS,BigbatchLineSearch}. However, these approaches have not analyzed the shape of the loss functions in update step direction in detail, which is important since the optimal step size stands in strong relation to this shape. To shed light on this, our work empirically analyses the shape of the loss function along update step direction for deep learning scenarios often considered in optimization. We further elaborate on the properties found to define a simple and empirically justified optimizer. 
  
Our contributions are as follows:
\textbf{1:} Our empirical analysis suggests that the mini-batch loss in the negative gradient direction mostly shows locally convex shapes. Furthermore, we show that parabolic approximations are well suited to estimate the minima in these directions (Section \ref{sec:sample_loss_lines}).
\textbf{2:} Exploiting this parabolic observation, we build a simple line search optimizer that constructs its loss function-dependent learning rate schedule. The performance of our optimization method is extensively analyzed, including a comprehensive comparison to other optimization methods (Sections \ref{sec:algorithm},\ref{sec:evaluation}).
\textbf{3:} We provide a convergence analysis that backs our empirical results under strong assumptions (Section \ref{subsec_theory}).
\textbf{4:} We provide a general investigation of exact line searches on batch losses and their relation to line searches on the exact loss as well as their relation to our line search approach (Section \ref{sec:optimality}) and, finally, analyze the relation of our approach to interpolation (Section \ref{sec:interpolation_analysis}).

\indent The  full-batch loss  $\mathcal{L}$ is defined as the average over realizations of a mini-batch-wise loss function $\mathcal{L}_{\mathbb{B}}$: $\quad$
$ \mathcal{L} : \mathbb{R}^m \rightarrow \mathbb{R}$, $\boldsymbol{\theta} \mapsto n^{-1} \sum_{i=1}^{n} \mathcal{L}_{\mathbb{B}_i}(\boldsymbol{\theta})$ with $\mathbb{B}_i$ being a batch of input samples, and $n$ the amount of batches in the dataset. $\boldsymbol{\theta} \in \mathbb{R}^m$ denotes the parameters to be optimized. Note, that we consider a sample as one mini-batch of multiple inputs. In this work, we consider $\BL$ at optimization step $t$ in negative gradient direction:
\begin{equation}\label{eq:line}
	l_t(s): \mathbb{R} \rightarrow \mathbb{R},s \mapsto \BL(\boldsymbol{\theta}_t+s \frac{-\mathbf{g}_t}{||\mathbf{g}_t||}),
\end{equation}
where $\mathbf{g}_t$ is  $\nabla_{\boldsymbol{\theta}_t} \BL(\boldsymbol{\theta}_t)$. We denote $l_t$ as \textit{mini-batch loss along a line} and $s$ is the step on this line.
 The motivation of our work builds upon the following assumption:
\newtheorem{assumption}{Assumption}
\begin{assumption}(Informal) The position $\mathbf{\theta}_{min}=\mathbf{\theta}_t+s_{min} \frac{-\mathbf{g}_t}{||\mathbf{g}_t||}$ of a minimum of $l_t$ is a well enough estimator for the position of a minimum of the full-batch loss $\mathcal{L}$ on the same line to perform a successful optimization process.
\label{ass:line_min}
\end{assumption} \vspace{-0.1cm}We empirically analyze Assumption 1 further in section \ref{sec:optimality}.
\section{Related work}
Our optimization approach is based on well-known methods, such as line search, the non-linear conjugate gradient method, and quadratic approximation, which can be found in Numerical Optimization \cite{numerical_optimization}. In addition, \cite[\S 3.5]{numerical_optimization} describes a similar line search routine for the deterministic setting. The concept of parabolic approximations is also exploited by the well-known line search of More and Thunte \cite{more1994line}.
Our work contrasts common optimization approaches in deep learning by directly exploiting the local parabolic observation (see Section \ref{sec:sample_loss_lines}) of $l_t$. Similarly, \textit{SGD-HD} \cite{hypergradientdescent} performs update steps towards a minimum of $l_t$, by performing gradient descent on the learning rate.  Concurrently, \cite{empericalLineSearchApproximations}  explored a similar direction as this work by analyzing possible line search approximations for DNN loss landscapes but does not exploit these for optimization.

The recently published \textit{Stochastic Line-Search} (\textit{SLS}) \cite{backtracking_line_search_NIPS} is an optimized backtracking line search based on the Armijo condition, which samples, like our approach, additional mini-batch losses from the same batch and checks the Armijo condition on these. \cite{backtracking_line_search_NIPS} assumes that the model interpolates the data.  The latter states formally that the gradient with respect to each batch converges to zero at the optimum of the convex function. This implies that if $\mathcal{L}(\boldsymbol{\theta})$ is minimized at $\boldsymbol{\theta}^*$ and thus $\nabla_{\boldsymbol{\theta}^*}\mathcal{L}(\boldsymbol{\theta}^*)=0$ holds, then for all  $\mathcal{L}_{\mathbb{B}_i}$  we have that $\nabla_{\boldsymbol{\theta}^*}\mathcal{L}_{\mathbb{B}_i}(\boldsymbol{\theta}^*)=0$. \textit{SLS} exhibits competitive performance against multiple optimizers on several DNN tasks.  \cite{backtracking_line_search} introduces a related idea but does not provide empirical results for DNNs. \cite{BigbatchLineSearch} also uses a backtracking Armijo line search, but with the aim to regulate the optimal batch size.\\
The methodically appealing but complex \textit{Probabilistic Line Search} (\textit{PLS}) \cite{probabilisticLineSearch} approximates $\mathcal{L}$ along lines with a Gaussian Process posterior based on realizations of $\mathcal{L}_{\mathbb{B}_i}$ for different batches $\mathbb{B}_i$. The mean of this Posterior is a cubic spline. A suitable update step size is found by exploiting a probabilistic formulation of the Wolf conditions.\\
\textit{Gradient Only Line Search} (\textit{GOLS1}) \cite{gradientOnlyLineSearch} \textit{GOLS1} searches for a minimum on lines by looking for a sign change of the first directional mini-batch derivative in search direction. For each measurement, a different $\mathbb{B}_i$ is chosen. Both approaches show that they can optimize successfully on several machine learning problems and can compete against plain \textit{SGD}.
 
From the perspective of assumptions about the shape of the loss landscape,
second order methods such as \textit{oLBFGS} \cite{oLBFGS}, \textit{KFRA} \cite{gausnewton}, \textit{L-SR1} \cite{L-sr1}, \textit{QUICKPROP} \cite{quickprop}, \textit{S-LSR1} \cite{S-LBFGS}, and  \textit{KFAC} \cite{KFAC}
generally assume that the loss function can be approximated locally by a parabola of the same dimension as the loss function.
Adaptive methods such as \textit{SGD} with momentum \cite{grad_descent}, \textit{ADAM}\cite{adam}, \textit{ADAGRAD}\cite{adagrad}, \textit{ADABOUND}\cite{AdaBound}, \textit{AMSGRAD}\cite{AmsGrad} or \textit{RMSProp} \cite{rmsProp} focus more on the handling of noise than on shape assumptions. 
In addition, methods exist that approximate the loss function in specific directions: The \textit{L4} adaptation scheme \cite{L4} as well as \textit{ALIG} \cite{L4_alternative} estimate step sizes by approximating the loss function linearly in negative gradient direction, whereas our approach approximates the loss function parabolically in negative gradient direction. 
Finally,
\textit{COCOB} \cite{cocob} has to be mentioned, an alternative learning rate free approach, which automatically estimates step directions and sizes with a reward-based coin betting concept.

\vfill
\pagebreak

 \section{Empirical analysis of the shape of mini-batch losses along lines}
 \vspace{-0.2cm}
 \label{sec:sample_loss_lines}
  \begin{figure*}[t!]
  	\centering
  	\includegraphics[width=0.19\linewidth]{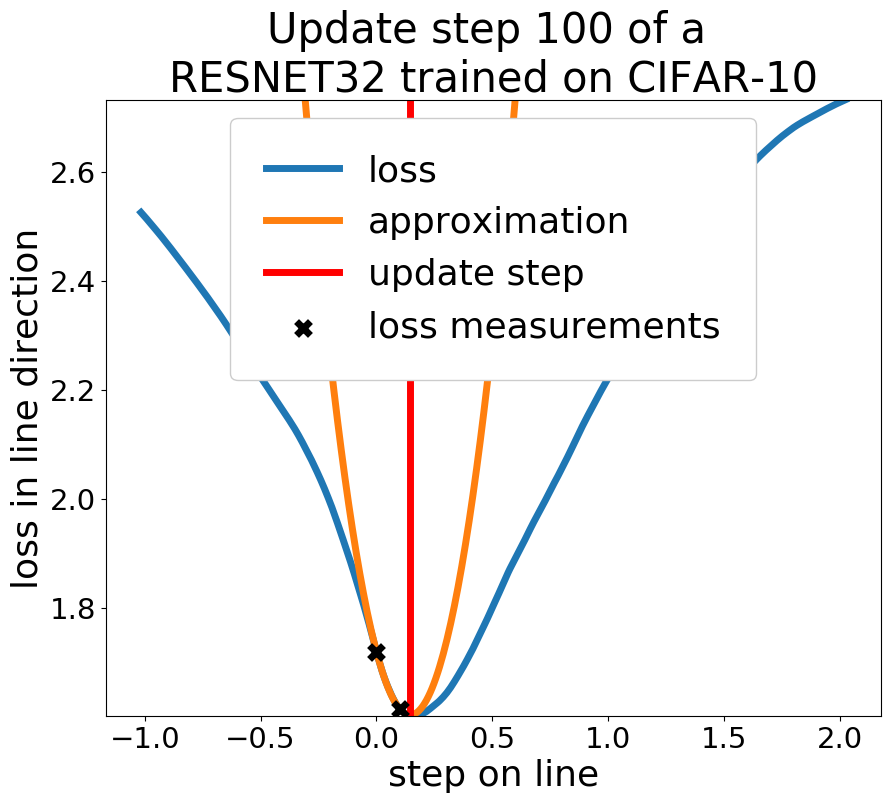}
  	\includegraphics[width=0.19\linewidth]{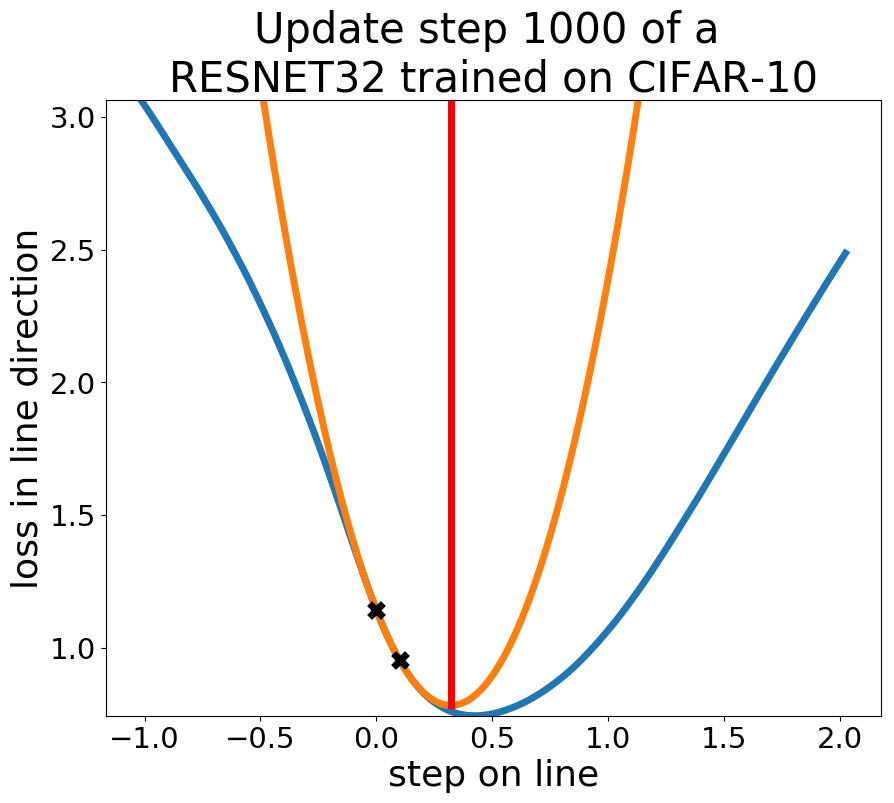}
  	\includegraphics[width=0.19\linewidth]{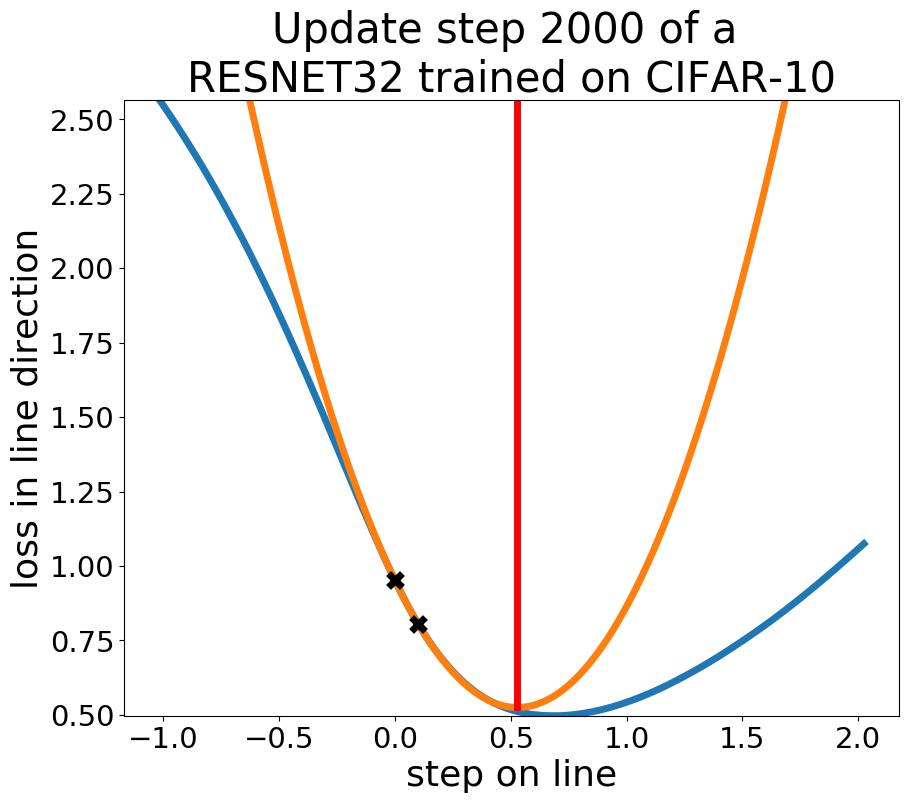}
  	\includegraphics[width=0.19\linewidth]{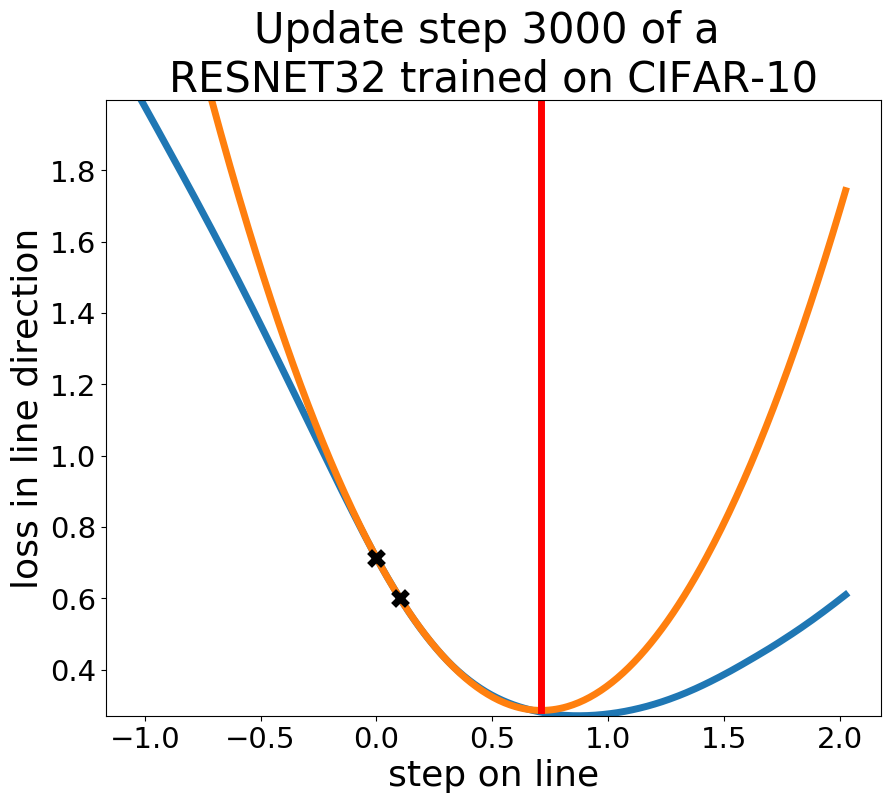}
  	\includegraphics[width=0.19\linewidth]{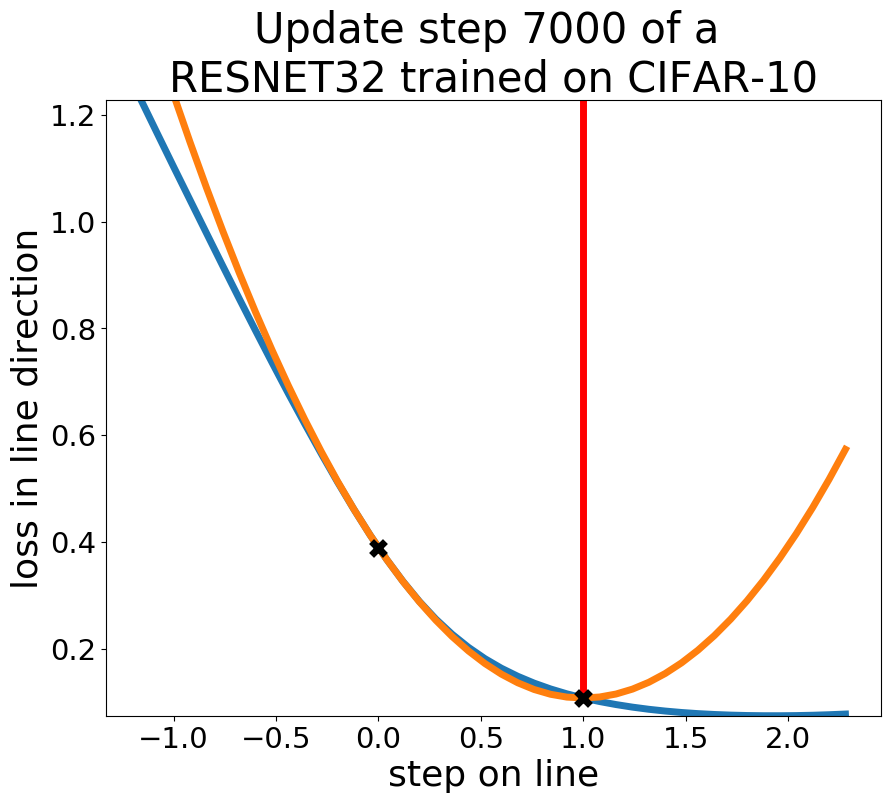}
  	\caption[]{Representative mini-batch losses along lines in negative normalized gradient direction (\textcolor{blue}{blue}), parabolic approximations (\textcolor{orange}{orange}) and the position of the approximated minima (\textcolor{red}{red}). Further plots are provided in Appendix \ref{sec:further_line_plots}. \vspace{-0.5cm}}
  	\label{fig.lines}
  \end{figure*}
In this section, we analyze mini-batch losses along lines (see Eq. \ref{eq:line}) during the training of multiple architectures and show that they locally exhibit mostly convex shapes, which are well suited for parabolic approximations.
We focus on  CIFAR-10, as it is extensively analyzed in optimization research for deep learning. However, we observed similar results on MNIST, CIFAR-100, and ImageNet, considering random samples. We analyzed mini-batch losses along the lines in the first 10000 SGD update step directions for four commonly used architectures (ResNet-32 \cite{resnet}, DenseNet-40 \cite{denseNet}, EfficientNet \cite{efficientnet}, MobileNetV2 \cite{mobilenet}). Along each line, we sampled 50 losses and performed a parabolic approximation (see Section \ref{sec:algorithm}).  An unbiased selection of our results on a ResNet32 is shown in Figure \ref{fig.lines}. Further results are given in Appendix \ref{sec:further_line_plots}. In accordance with \cite{walkwithsgd}, we conclude that the mini-batch losses along the analyzed lines tend to be locally convex. In addition, one-dimensional parabolic approximations of the form $f(s)= as^2+bs+c$ with $a\neq0$ are well suited to estimate the position of a minimum along such directions.

\begin{figure*}[h]
 	\centering
 	\includegraphics[trim={0 0cm 0cm,  0cm},clip,width=0.4\linewidth]{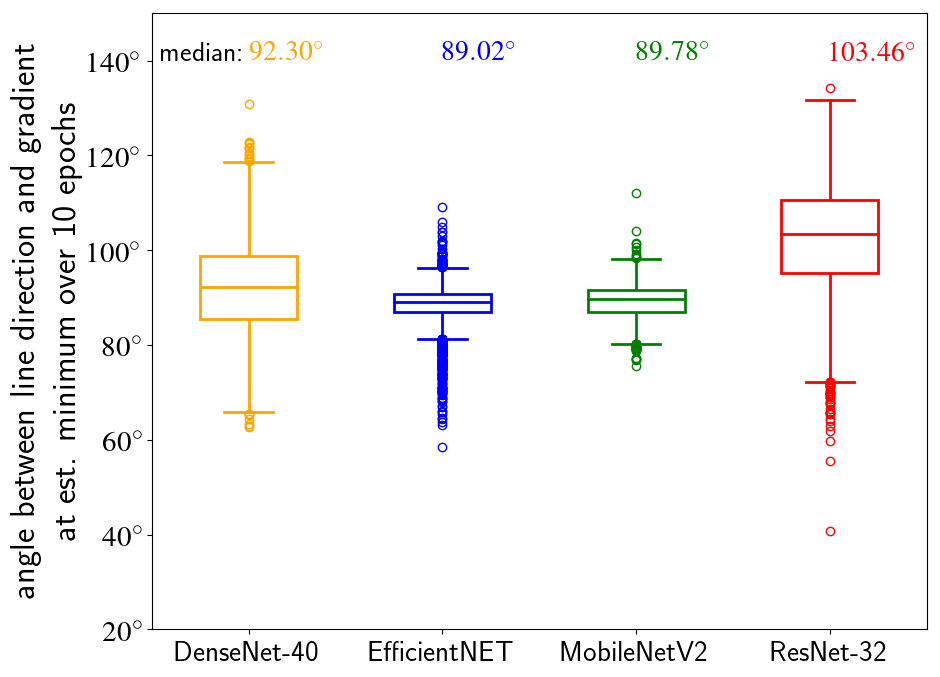}\quad
 	\includegraphics[trim={0 0cm 0cm 0cm},clip,width=0.4\linewidth]{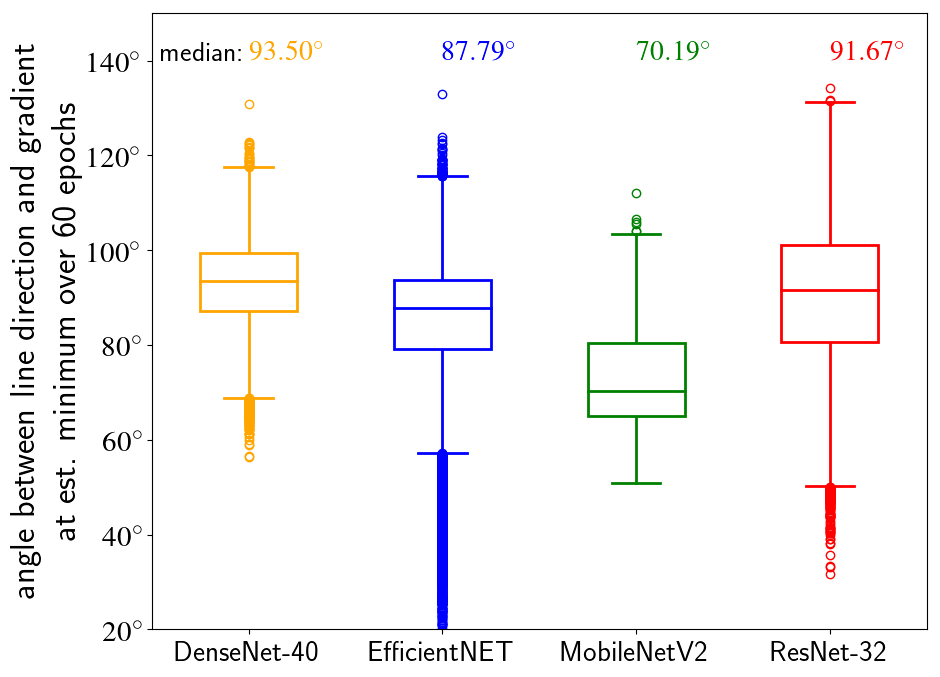}
 	\caption[]{Angles between the line direction and the gradient at the estimated minimum measured on the same mini-batch. If the angle is $90^\circ$, the estimated minimum is a real local minimum. We know from additional line plots that the found extrema or saddle points are minima. Left: measurement over the first 10 epochs. Right: measurement over the first 60 epochs. Update step adaptation (see Section \ref{subsec:features}) is applied.}
 	\label{fig:angles}
 \end{figure*}

\begin{figure}[t!]
	\newcommand\picscale{0.4}
	\newcommand\anglemin{0}
	\newcommand\anglemax{110}
	\newcommand\angles{130000}
	\newcommand\lossmin{0}
	\newcommand\lossmax{1}
	\newcommand\evalmin{0}
	\newcommand\evalmax{1}
	\newcommand\epochs{280}
	\newcommand\picheight{0.6\textwidth}
	\newcommand\legendwidth{9cm}
	\newcommand\legendheight{3cm}
	\centering

		\tikzsetnextfilename{pal_angles_continuos_1}
		\begin{tikzpicture}[scale=\picscale] 
		\begin{axis}[
		width=\textwidth, 
		height=\picheight,
		grid=major, 
		grid style={dashed,gray!30}, 
		xlabel=training step, 
		xlabel style={font=\LARGE},
		ylabel=angle between line direction and gradient at est. minimum in $^\circ$,
		ylabel style={font=\LARGE,text width=0.4\textwidth},
		xmin=0,xmax=\angles,
		ymin=\anglemin,ymax=\anglemax,
		legend pos= south east,
		legend style={font=\Large},
		x tick label style={rotate=0,anchor=near xticklabel,font=\LARGE}, 
		y tick label style={font=\LARGE},
		p1/.style={draw=blue,line width=2pt},
		p2/.style={draw=red,line width=2pt},
		p3/.style={draw=green,line width=2pt},
		p4/.style={draw=brown,line width=2pt},
		p5/.style={draw=black,line width=2pt},
		]
		\addplot [p5] table[x=Step,y=Value,col sep=comma] {plot_data/angle_data/run-mnist_models_PAL_1_PAL_45000_150000_128_1_0_0.8_3.16_1_0_0_log-tag-_data_angle_per_step.csv};  
		\addplot [p1] table[x=Step,y=Value,col sep=comma] {plot_data/angle_data/run-EFFICIENTNET_models_PAL_1_PAL_45000_150000_128_1_0_1_3.16_1_0_0_log-tag-_data_angle_per_step.csv}; 
		\addplot [p2] table[x=Step,y=Value,col sep=comma] {plot_data/angle_data/run-MOBILENETV2_models_PAL_1_PAL_45000_150000_128_1_0_0.8_3.16_1_0_0_log-tag-_data_angle_per_step.csv}; 
		\addplot [p3] table[x=Step,y=Value,col sep=comma] {plot_data/angle_data/run-RESNET32_models_PAL_1_PAL_45000_150000_128_1_0_0.8_3.16_1_0_0_log-tag-_data_angle_per_step.csv};

 		\legend{3 Layer Conv. Net on MNIST,EfficientNet on CIFAR-100,MobileNetV2 on CIFAR-100,ResNet32 on CIFAR-100}

		\end{axis}
		\end{tikzpicture}
				\tikzsetnextfilename{pal_angles_continuos_2}
				\begin{tikzpicture}[scale=\picscale] 
				\begin{axis}[
				width=\textwidth, 
				height=\picheight,
				grid=major, 
				grid style={dashed,gray!30}, 
				xlabel=training step, 
				xlabel style={font=\LARGE},
				xmin=0,xmax=\angles,
				ymin=\anglemin,ymax=\anglemax,
				legend pos= south east,
				legend style={font=\Large},
				x tick label style={rotate=0,anchor=near xticklabel,font=\LARGE}, 
				y tick label style={font=\LARGE},
				p1/.style={draw=blue,line width=2pt},
				p2/.style={draw=red,line width=2pt},
				p3/.style={draw=green,line width=2pt},
				p4/.style={draw=brown,line width=2pt},
				p5/.style={draw=black,line width=2pt},
				p6/.style={draw=violet,line width=2pt},
				]
				\addplot [p5] table[x=Step,y=Value,col sep=comma] {plot_data/angle_data/imagenet/run-RESNET50_WD_models_PAL_1_PAL_1281167_500000_100_0.0316_0_0.8_3.16_1_0_0_log-tag-_data_angle_per_step.csv};  
				\addplot [p1] table[x=Step,y=Value,col sep=comma] {plot_data/angle_data/cifar10/run-EFFICIENTNET_models_PAL_1_PAL_45000_150000_128_0.1_0_0.8_3.16_1_0_0_log-tag-_data_angle_per_step.csv}; 
				\addplot [p2] table[x=Step,y=Value,col sep=comma] {plot_data/angle_data/cifar10/run-MOBILENETV2_models_PAL_1_PAL_45000_150000_128_0.1_0_0.8_3.16_1_0_0_log-tag-_data_angle_per_step.csv}; 
				\addplot [p3] table[x=Step,y=Value,col sep=comma] {plot_data/angle_data/cifar10/run-RESNET32_models_PAL_1_PAL_45000_150000_128_0.1_0_0.8_3.16_1_0_0_log-tag-_data_angle_per_step.csv}; 
		
		 		\legend{ResNet 50 on Imagenet,EfficientNet on CIFAR-10,MobileNetV2 on CIFAR-10,ResNet32 on CIFAR-10}
				\end{axis}
				\end{tikzpicture}
		\caption{Angles between the line direction and the gradient at the estimated minimum measured on the same batch plotted over a whole training process on several networks and datasets. This figure clarifies that the parabolic observation holds also on further datasets and during the training process. It fits best for MNIST and becomes worse for ImageNet. Measuring step sizes and update step adaptations factors (see Sections \ref{sec:paremeter_update_rule},\ref{subsec:features}) were used to fit the loss along the line decently.}
		\label{fig:angles_continuous}
\end{figure}
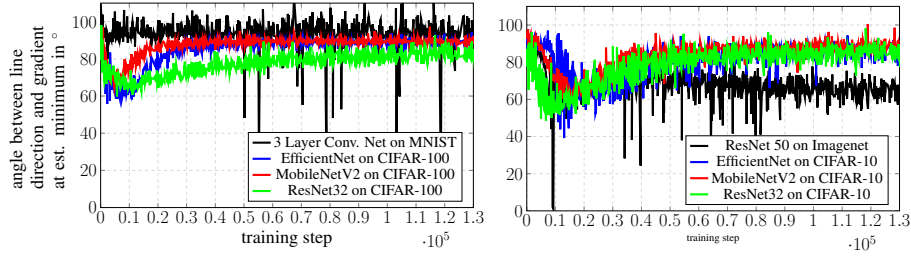

 To substantiate the later observation, we analyzed the angle between the line direction and the gradient at the -by a parabola- estimated minimum  during training.  A position is a local extremum or saddle point of the loss along a line if and only if the angle between the line direction and the gradient at the position is $90^\circ$, if measured on the same mini-batch. \footnote{This holds because if the directional derivative of the measured gradient in line direction is 0, the current position is an extremum or saddle point along the line and the angle is $90^\circ$. If the position is not an extremum or saddle point, the directional derivative is not 0 \cite{numerical_optimization}.} Measuring step sizes and update step adaptations factors (see Sections \ref{sec:paremeter_update_rule} and\ref{subsec:features}) were chosen to fit the mini-batch loss along the line decently.
As shown in Figures \ref{fig:angles} and \ref{fig:angles_continuous}, the parabolic observation holds well for several architectures trained on MNIST, CIFAR-10, CIFAR-100 and ImageNet. The observation fits best for MNIST and gets worse for more complex tasks such as ImageNet. We can ensure that the extrema found are minima since we additionally plotted the mini-batch loss along the line for each update step.\\
In addition, we analyzed the mini-batch loss along lines in conjugate-like directions and random directions. Mini-batch losses along lines in conjugate like directions also tend to have convex shapes (see Appendix \ref{fig:Parabolic property in adapted} Figure \ref{fig:angles_continuous_mom} ). However, mini-batch losses along lines in random directions rarely exhibit convex shapes. 

\label{sec:introduction}

\vfill
\pagebreak

\section{The line search algorithm}
\label{sec:algorithm}

We introduce \textbf{P}arabolic \textbf{A}pproximation \textbf{L}ine Search (\pal), which exploits the observation that parabolic approximations are suited to estimate the minimizer of the mini-batch loss along lines. This simple approach combines well-known methods from basic optimization such as parabolic approximation and line search  \cite{numerical_optimization} to perform an efficient line search. We note that the general idea of this method can be applied to any optimizer that provides an update step direction.

\subsection{Parameter update rule}
\label{sec:paremeter_update_rule}
\begin{wrapfigure}{r}{0.5\linewidth}
  	\centering
  	
  	\includegraphics[width=\linewidth]{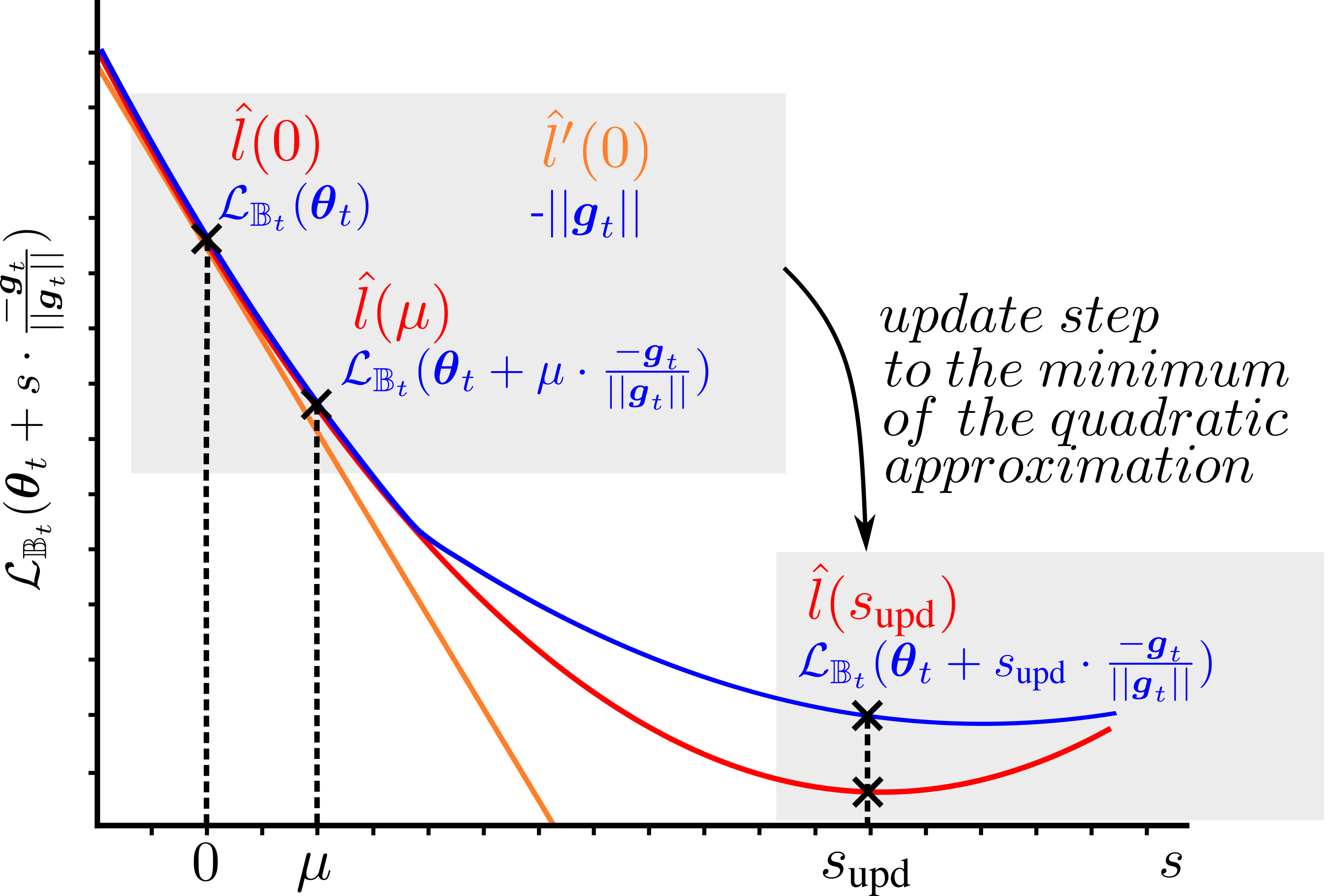}	
  	
  	\caption[sd]{Basic idea of \pal's parameter update rule. The blue curve is the mini-batch loss along the negative gradient starting at $\BL(\vec{\theta}_t)$. It is defined as  $l_t(s)=\BL(\boldsymbol{\theta}_t+s \frac{-\mathbf{g}_t}{||\mathbf{g}_t||})$ where $\vec{g}_t$ is $ \nabla_{\mathbf{\theta}_t} \BL(\boldsymbol{\theta}_t)$. The red curve is its parabolic approximation $\hat{l}(s)$. $l_t(0)$, $l_t(\mu)$ and $\mathbf{g}_t$ (orange) are the three  parameters needed to determine the update step $s_{upd}$ to the minimum of the parabolic approximation.}
  	\label{fig:update_step}
  	\vspace{-1cm}
  \end{wrapfigure}

An intuitive explanation of \pal's parameter update rule based on a parabolic approximation is given in Figure \ref{fig:update_step}.
Since $l_t$ (see Eq.\ref{eq:line}) is assumed to exhibit a convex and almost parabolic shape, we approximate it with  $\hat{l}_t(s)=as^2+bs+c$ with $a \in  \mathbb{R}^+$ and $b,c \in \mathbb{R}$. Consequently, we need three measurements of $l_t$ to define $a,b$ and $c$. Those are given by the current loss $l_t(0)$, the derivative
  in gradient direction $l'_t(0)= -||\mathbf{g}_t||$ (see Eq. \ref{eq:directional_derivative}) and an additional loss $l_t(\mu)$  with measuring distance $\mu \in \mathbb{R^+}$.
  It is simple to show that  $a=\frac{l_t(\mu)-l_t(0)-l_t'(0)\mu}{\mu^2}$, $b=l_t'(0)$, and $c=l_t(0)$. The update step $\s$ to the  minimum of the parabolic approximation $\hat{l}_t(s)$ is thus given by: 
\begin{equation}\label{eq:update_step}
	 \s[t]=-\frac{\hat{l}_t^{\prime}(0)}{\hat{l}_t^{\prime\prime}(0)}=   -\frac{b}{2a}=\frac{-l_t^{\prime}(0)}{2\frac{l_t(\mu)-l_t(0)-l_t^{\prime}(0)\mu}{\mu^2}}
\end{equation}
Note, that $\hat{l}_t^{\prime\prime}(0)$ is the second derivative of the approximated parabola and is only identical to the exact directional derivative $\frac{-\mathbf{g}_t}{||\mathbf{g}_t||}H(\BL(\vec{\theta}_t))\frac{-\mathbf{g}_t^T}{||\mathbf{g}_t||}$ if the parabolic approximation fits. 
The normalization of the gradient to unit length (Eq.\ref{eq:line}) was chosen to have the measuring distance $\mu$ independent of the gradient size and of weight scaling. 
Note that two network inferences are required to determine $l_t(0)$ and $l_t(\mu)$. Consequently, \pal\ needs two forward passes and one backward pass through a model.
 Further on, $\BL$ may include random components, but to ensure continuity during one line search, drawn random numbers have to be reused for each value determination of $\BL[t]$ at $t$ (e.g., for Dropout \cite{Dropout}).
The memory required by \pal\ is similar to \textit{SGD} with momentum, since only the last update direction has to be saved.  A basic version of \pal\ is given in Algorithm \ref{alg:PAL_basic}.

\begin{algorithm}[t]
	\caption{The basic version of our proposed line search algorithm. See Section \ref{sec:algorithm}\ for details.}  
	\vspace{-0.3cm}
	\begin{multicols}{2}
	\begin{algorithmic}[1]
	\label{alg:PAL_basic}
		\renewcommand{\algorithmicrequire}{\textbf{Input:}}
		\renewcommand\algorithmiccomment[1]{
			\hfill \eqmakebox[a][l]{\small #1} %
		}
		\REQUIRE $\mu$: measuring step size
		\REQUIRE $\BL $: mini-batch loss function
		\REQUIRE $\vec{\theta}_0$: initial parameter vector	
		\STATE $t \leftarrow 0$
		\WHILE{$\vec{\theta}_t$ not converged}
		\STATE $\mathbb{B}_t \leftarrow \text{sampleBatch}()$ 
		\STATE $l_0 \leftarrow \BL[t](\vec{\theta}_{t})$ 
		\COMMENT{\# $l_0 = l_t(0)$ see Eq. \ref{eq:line}}
		\STATE $\mathbf{g}_t \leftarrow -\nabla_{\mathbf{\theta}_t}\BL[t](\mathbf{\theta}_t)$ 
		\STATE $l_\mu \leftarrow \BL[t](\mathbf{\theta}_t+\mu \frac{\mathbf{g}_t}{||\mathbf{g}_t||})$ 
		\STATE $b \leftarrow  -||\mathbf{g}_t||$ 
		\STATE $a \leftarrow \frac{l_{\mu}-l_0-b\mu}{\mu^2}$ 
		\IF{proper curvature}
		\STATE $\s \leftarrow -\frac{b}{2a}$ 
		\ELSE  
		\STATE \COMMENT{\hspace{-2.9cm}  \# set $\s$ according to section \ref{subsec:casediscrimination}}
		\ENDIF
		\STATE	$\mathbf{\theta}_{t+1} \leftarrow \mathbf{\theta}_{t}+ \s\frac{\mathbf{g}_t}{||\mathbf{g}_t||}$ 
		\STATE $t \leftarrow t+1$
		\ENDWHILE
		\RETURN $\mathbf{\theta}_t$
	\end{algorithmic} 
	\end{multicols}
	\vspace{-0.3cm}
\end{algorithm}
\subsection{Case discrimination of parabolic approximations}
\label{subsec:casediscrimination}
Since not all parabolic approximations are suitable for parameter update steps, the following cases are considered separately. Note that $b=\hat{l}^{\prime}(0)=l^{\prime}(0)$ and  $a = 0.5\hat{l}^{\prime\prime}(0) \approx 0.5\l^{\prime\prime}(0)$..
\textbf{1:} $a>0$ and $b<0$: the parabolic approximation has a minimum in line direction, thus, the parameter update is done as described in Section \ref{sec:paremeter_update_rule}.
\textbf{2:} $a\leq0$ and $b<0$: the parabolic approximation has a maximum in negative line direction or is a line with a negative slope. In those cases, a parabolic approximation is inappropriate. $s_{upd}$ is set to $\mu$, since the second measured point has a lower loss than the first.
\textbf{3:} Since $b=-||g_t||$ cannot be greater than 0, the only case left is an extremum at the current position ($l^\prime(0)=0)$.
In this case, no weight update is performed. However, the mini-batch loss function changes with the next batch, which likely does not have an extremum at exactly the same point.
In accordance with Section \ref{sec:sample_loss_lines}, cases two and three appeared very rarely in our experiments. 
\subsection{Additions}
\label{subsec:features}
Here multiple additions for Algorithm \ref{alg:PAL_basic} to fine-tune the performance and handle degenerate cases are introduced. Our hyperparameter sensitivity analysis (Appendix \ref{sec:sens_analysis}) suggests that the influence of the introduced hyperparameters on the optimizer's performance is low. Thus, they only need to be adapted to fine-tune the results.
The full version of \pal\ including all additions is given in Appendix \ref{sec:PAL with all addition}  Algorithm \ref{alg:PAL}.

\textbf{Direction adaptation:} Instead of following the direction of the negative gradient we follow an adapted conjugate-like direction $\mathbf{d}_t$:
\begin{equation}\label{eq:9}
	\mathbf{d}_t=-\nabla_{\mathbf{\theta}_t} \BL(\mathbf{\theta}_t) + \beta \mathbf{d}_{t-1} \quad \mathbf{d}_0=-\nabla_{\mathbf{\theta}_0}\BL(\mathbf{\theta}_0) 
\end{equation}
with $\beta \in [0,1]$. Since now an adapted direction is used, $l_t^\prime(0)$ changes to: 
\begin{equation}\label{eq:directional_derivative}
	l_t^{\prime}(0) = \nabla_{\mathbf{\theta}_t}\BL(\mathbf{\theta}_t)  \frac{\mathbf{d}_t}{||\mathbf{d}_t||}
\end{equation}
This approach aims to find a more optimal search direction than the negative gradient.
We implemented and tested the formulas of Fletcher-Reeves \cite{CGFletcher}, Polak-Ribi\`ere \cite{CGRibiere}, Hestenes-Stiefel \cite{CGHestenes} and Dai-Yuan \cite{CGDai} to determine conjugate directions under the assumption that the loss function is a quadratic. However, choosing a constant $\beta$ of value $0.2$ or $0.4$ performs equally. The influence of $\beta$ and dynamic update steps on \pal's performance is discussed in Appendix \ref{sec:abldation_direction_adaptation}. In the analyzed scenario, $\beta$ can both increase and decrease the performance, whereas dynamic update steps mostly increase the performance. A combination of both is needed to achieve optimal results.

\textbf{Update step adaptation:}
Our preliminary experiments revealed a systematic error caused by constantly approximating with slightly too narrow parabolas. Therefore, $\s$ is multiplied by a parameter $\alpha\geq1$ (compare to Eq. \ref{eq:update_step}). This is useful to estimate the minimum's position along a line more exactly but has minor effects on training performance.

\textbf{Maximum step size:} To hinder the algorithm from failing due to inaccurate parabolic approximations, we use a maximum step size $s_\text{{max}}$. 
The new update step is given by $\text{min}(\s ,s_\text{{max}})$. However, most of our experiments with $s_\text{{max}}=10^{0.5}\approx 3.16$ never reached this step size and still performed well.\\

\subsection{Theoretical considerations}
\label{subsec_theory}

 Usually, convergence in deep learning is shown for convex stochastic functions with an L-Lipschitz continuous gradient. However, since our approach originates from empirical results, it is not given that a profound theoretical analysis is possible.
In order to show any convergence guarantees for parabolic approximations, we have to fall back to uncommonly strong assumptions, which lead to quadratic models. Since convergence proofs on quadratics are of minor importance for most readers, our derivations can be found in Appendix \ref{app:theoretical_considerations}.
  
\section{Evaluation}
\label{sec:evaluation}
\subsection{Experimental design}
\label{sec:exp_design}
We performed a comprehensive evaluation to analyze the performance of \pal\ on a variety of deep learning optimization tasks. Therefore, we tested \pal\ on commonly used architectures on CIFAR-10  \cite{CIFAR-10}, CIFAR-100 \cite{CIFAR-10} and ImageNet \cite{IMAGENET}. For CIFAR-10 and CIFAR-100, we evaluated on DenseNet40 \cite{denseNet}, EfficientNetB0 \cite{efficientnet}, ResNet32 \cite{resnet} and MobileNetV2 \cite{mobilenet}. On ImageNet we evaluated on DenseNet121 and ResNet50.  In addition, we considered an RNN trained on the Tolstoi war and peace text prediction task. We compare \pal\ to  \textit{SLS}\cite{backtracking_line_search_NIPS}, whose Armijo variant is state-of-the-art in the line search field for DNNs at the time of writing. In addition, we compare against the following well studied and widely used first order optimizers: \textit{SGD} with momentum \cite{grad_descent}, \textit{ADAM}\cite{adam}, and \textit{RMSProp}\cite{rmsProp} as well as against \textit{SGDHD} \cite{hypergradientdescent} and \textit{ALIG} \cite{L4_alternative}, which automatically estimate learning rates in negative gradient direction and, finally, against the coin betting approach \textit{COCOB} \cite{cocob}.
To perform a fair comparison, we compared various hyperparameter combinations of commonly used hyperparameters for each optimizer. In addition, we utilize those combinations to analyze the hyperparameter sensitivity for each optimizer. Since a grid search on Imagenet was too expensive, the best hyperparameter configuration of  CIFAR-100's evaluation was used to test hyperparameter transferability. A detailed explanation of the experiments including hyperparameters and data augmentations used, are given in Appendix \ref{sec:further_experiment_information}. All in all, we trained over 4500 networks with Tensorflow 1.15 \cite{Tensorflow} on Nvidia Geforce GTX 1080 TI graphic cards. 
Since \pal\ is a  line search approach, the predefined learning rate schedules of SGD and the generated schedules of \textit{SLS}, \textit{ALIG}, \textit{SGDHD} and \pal\ were compared. Due to normalization, \pal's learning rate is given by $s_{\text{upd}_t} / ||\mathbf{d}_t||$.

\begin{figure*}[t]
	\def \figwidth {0.32}
	\def \figheight {0.25}
	\def \figwidthb {0.25}
	\centering
	\vspace{-0.5cm}
	
	\includegraphics[width=\figwidth\linewidth,height=\figheight\linewidth]{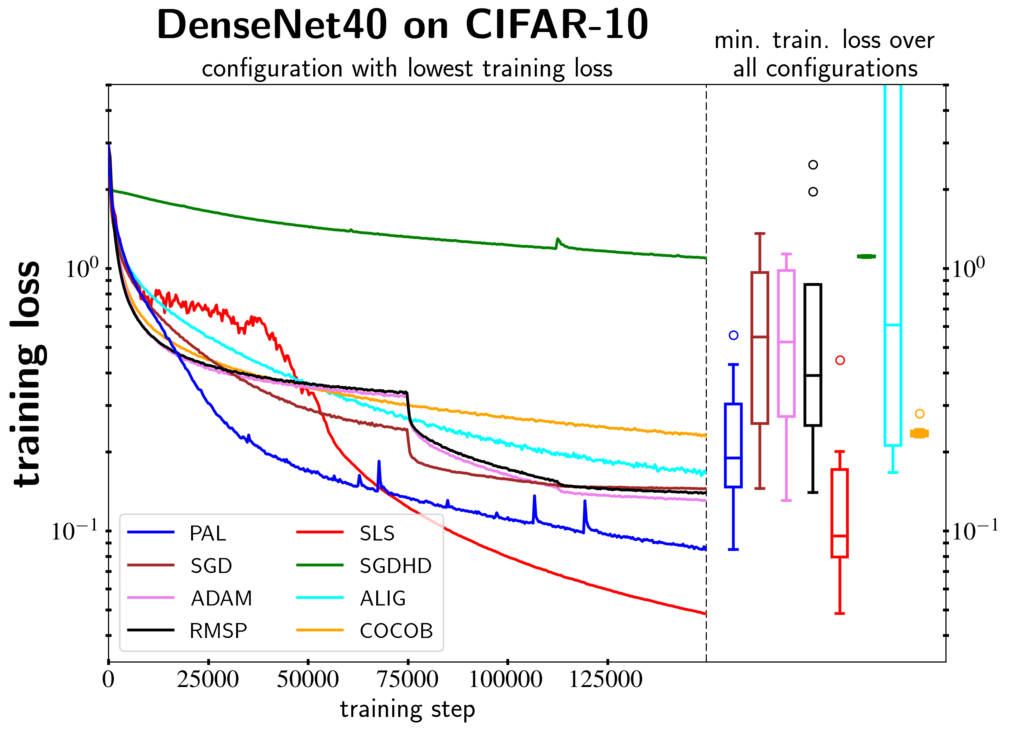}
	\includegraphics[width=\figwidth\linewidth]{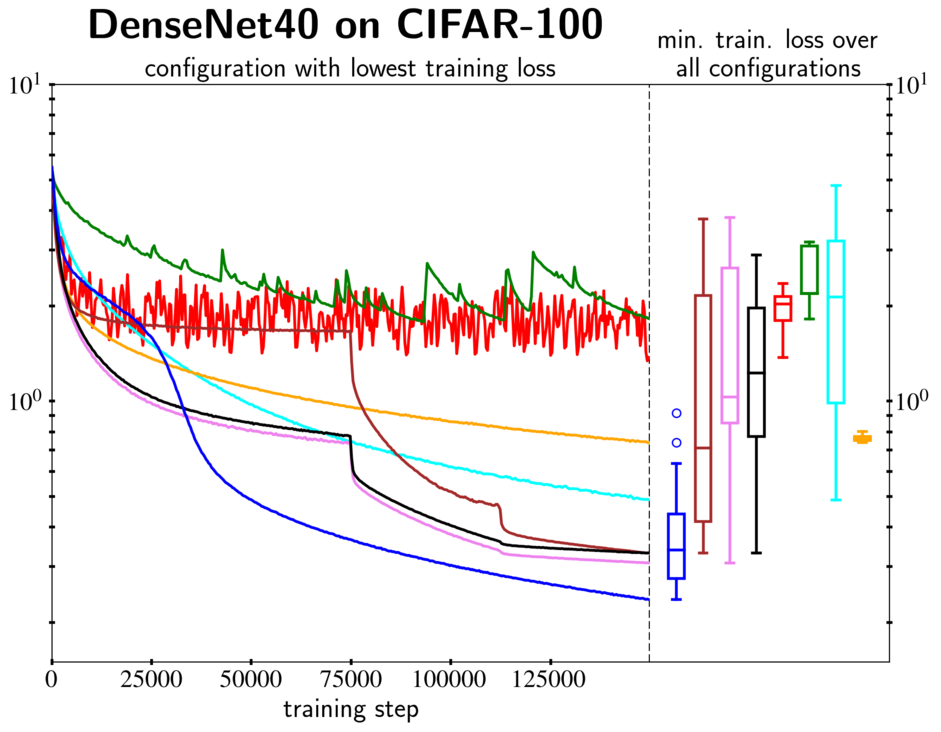}
	\includegraphics[width=\figwidth\linewidth,height=\figheight\linewidth]{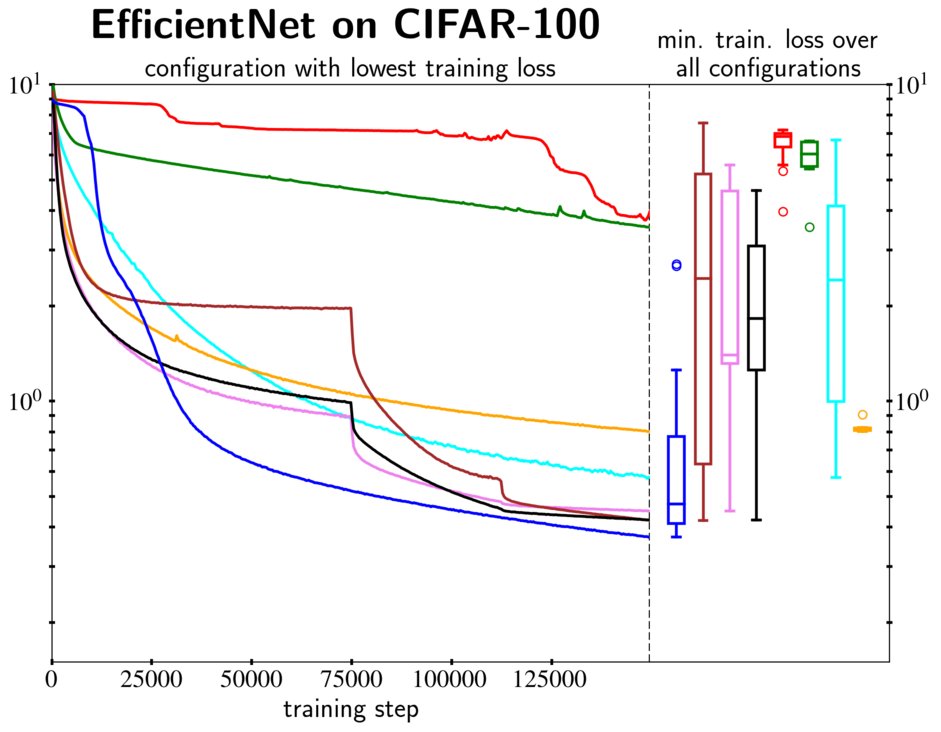}

	\includegraphics[width=\figwidth\linewidth,height=\figheight\linewidth]{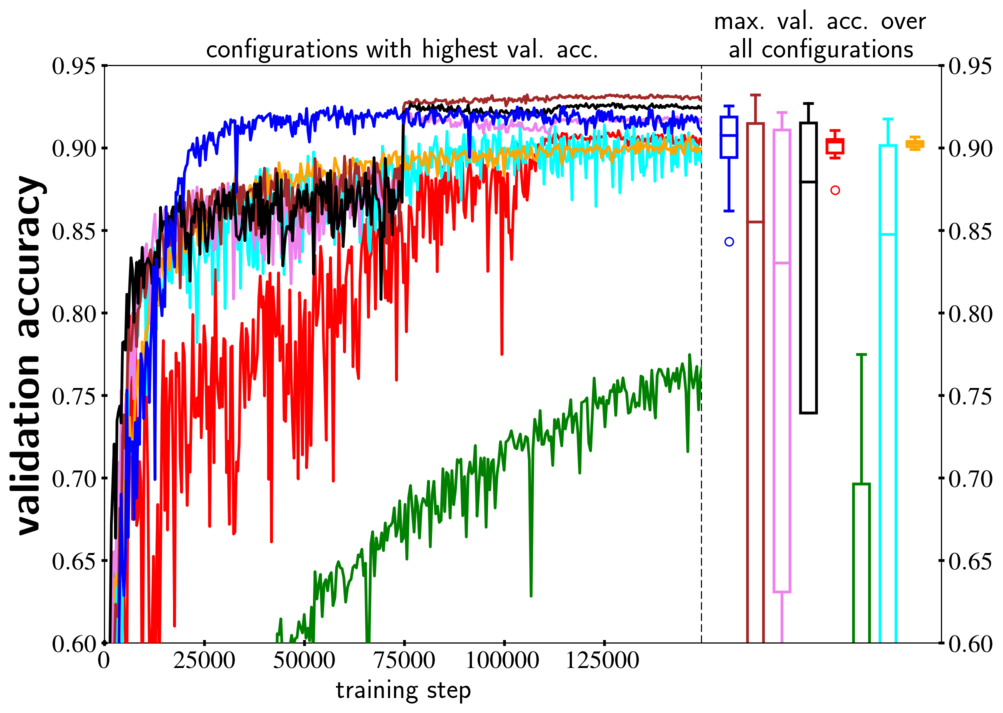}
	\includegraphics[width=\figwidth\linewidth,height=\figheight\linewidth]{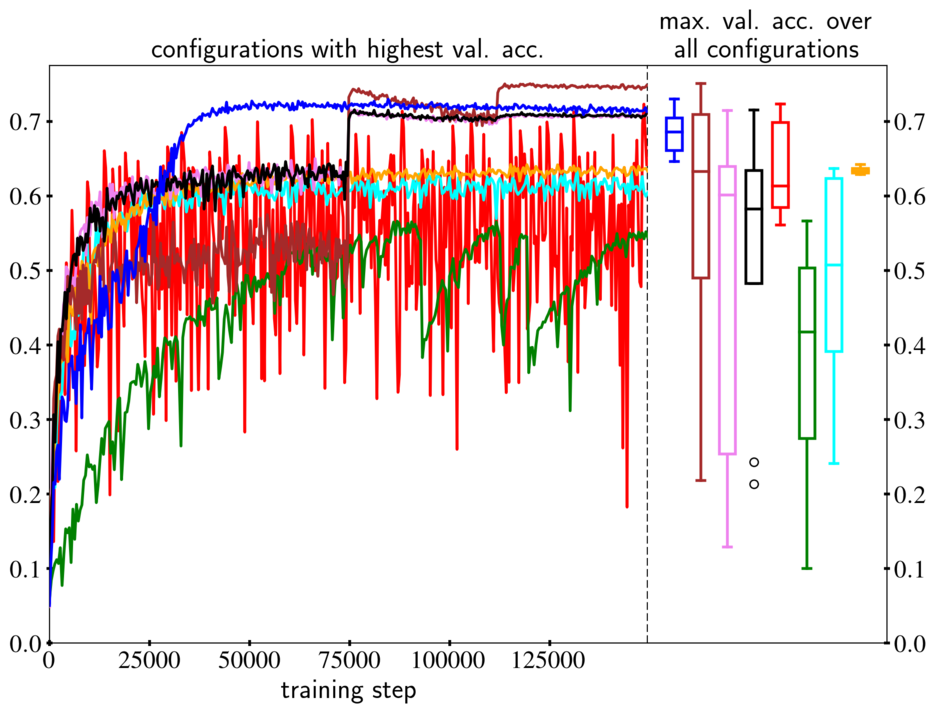}
	\includegraphics[width=\figwidth\linewidth,height=\figheight\linewidth]{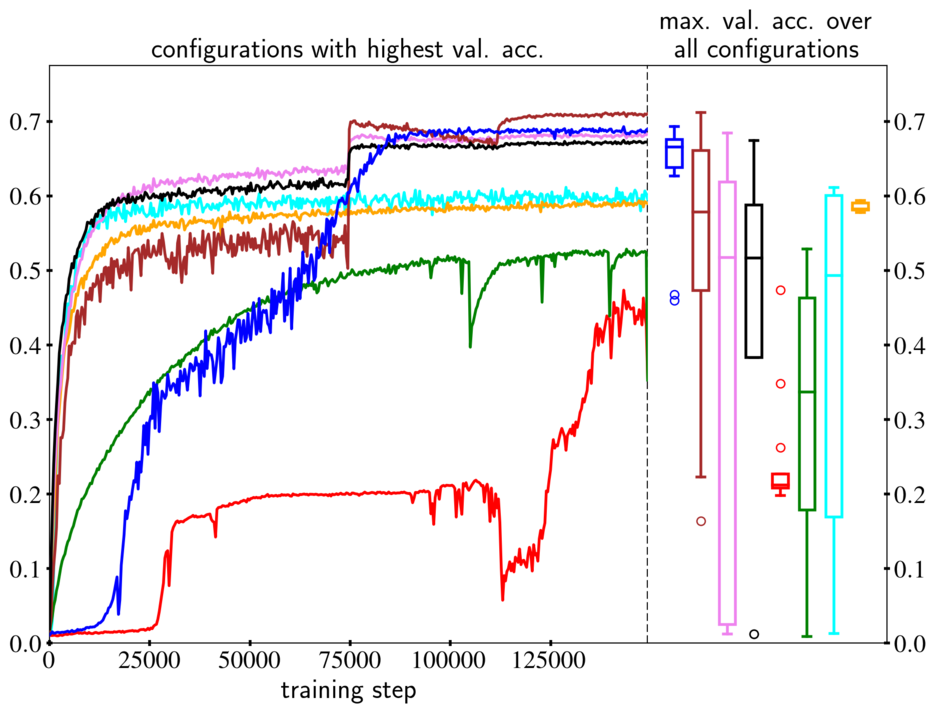}
	\\

	\includegraphics[width=\figwidth\linewidth,height=\figheight\linewidth]{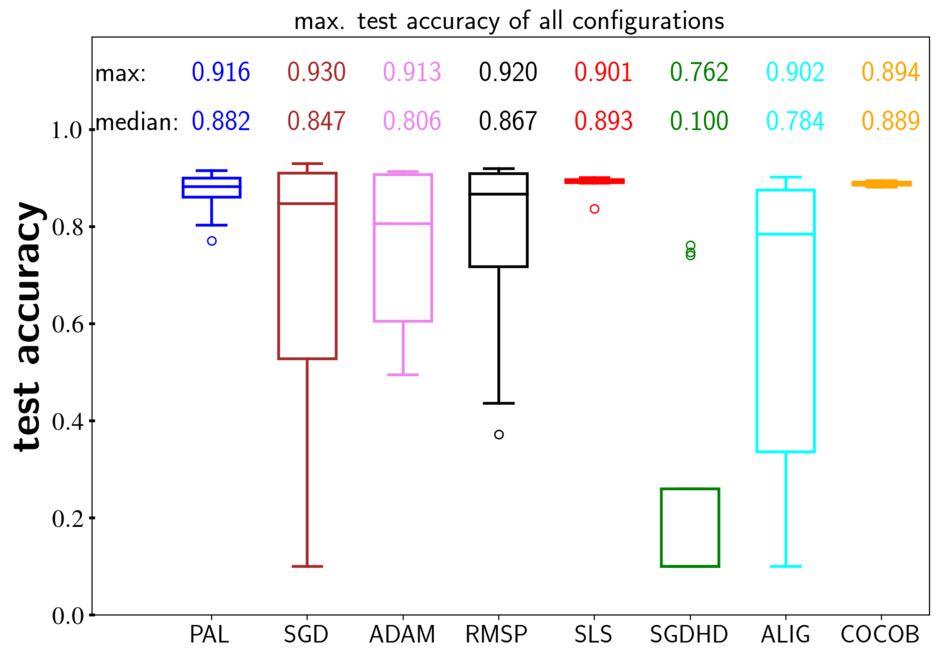}
	\includegraphics[width=\figwidth\linewidth,height=\figheight\linewidth]{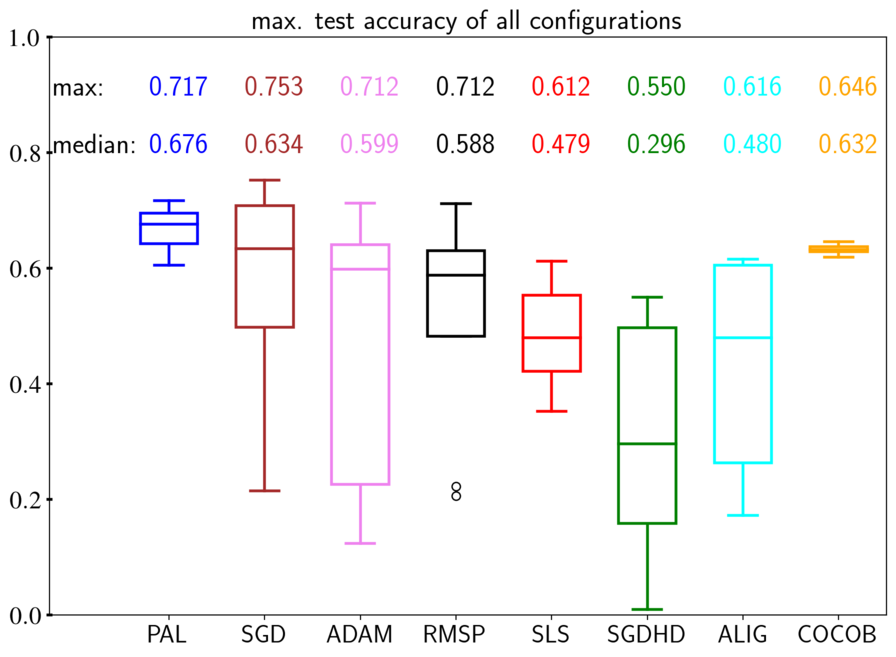}
	\includegraphics[width=\figwidth\linewidth,height=\figheight\linewidth]{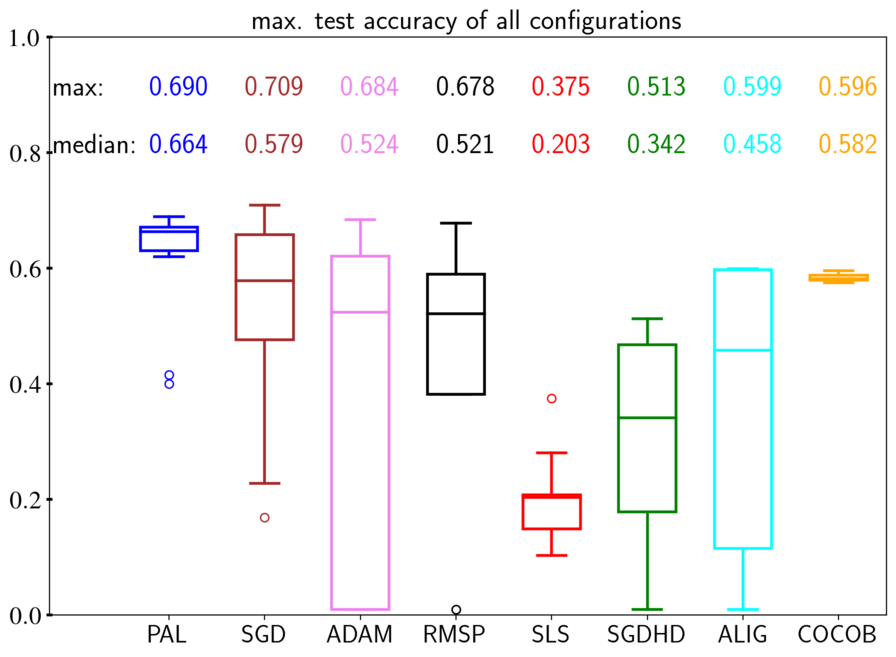}
	

	\includegraphics[width=\figwidth\linewidth,height=\figheight\linewidth]{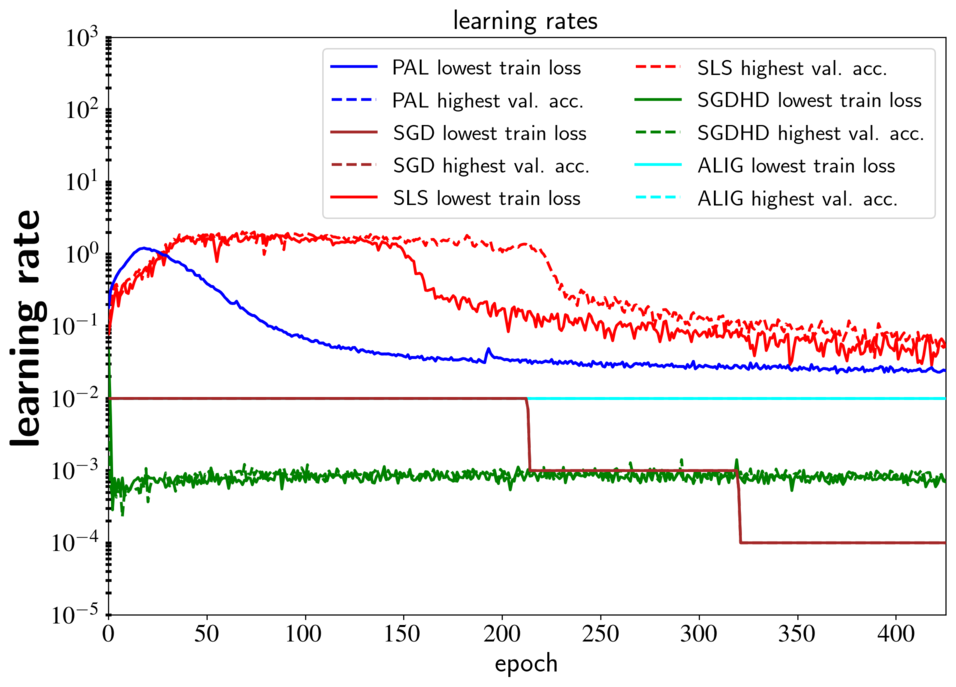}
	\includegraphics[width=\figwidth\linewidth,height=\figheight\linewidth]{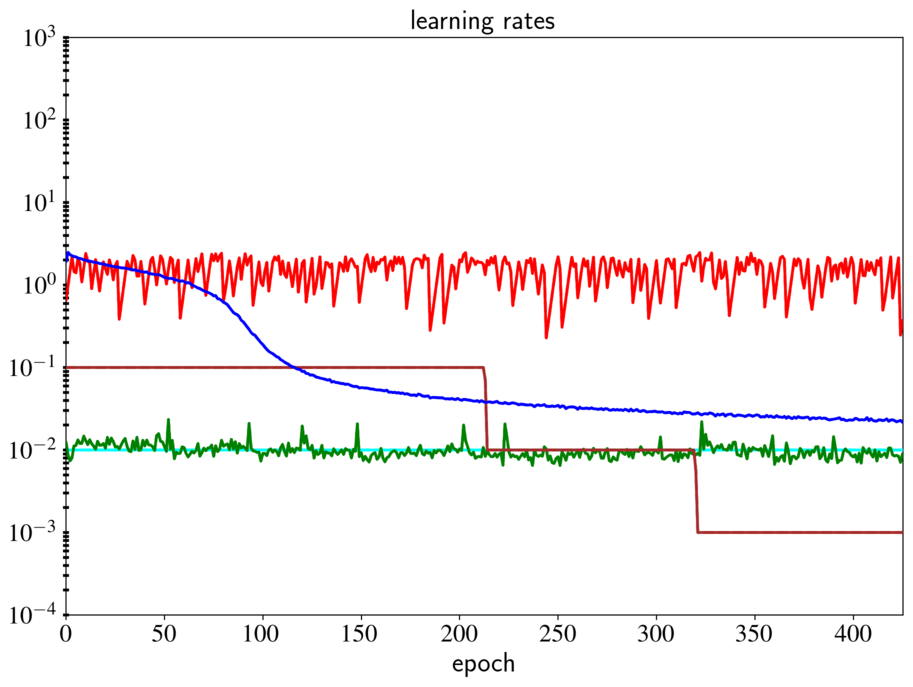}
	\includegraphics[width=\figwidth\linewidth,height=\figheight\linewidth]{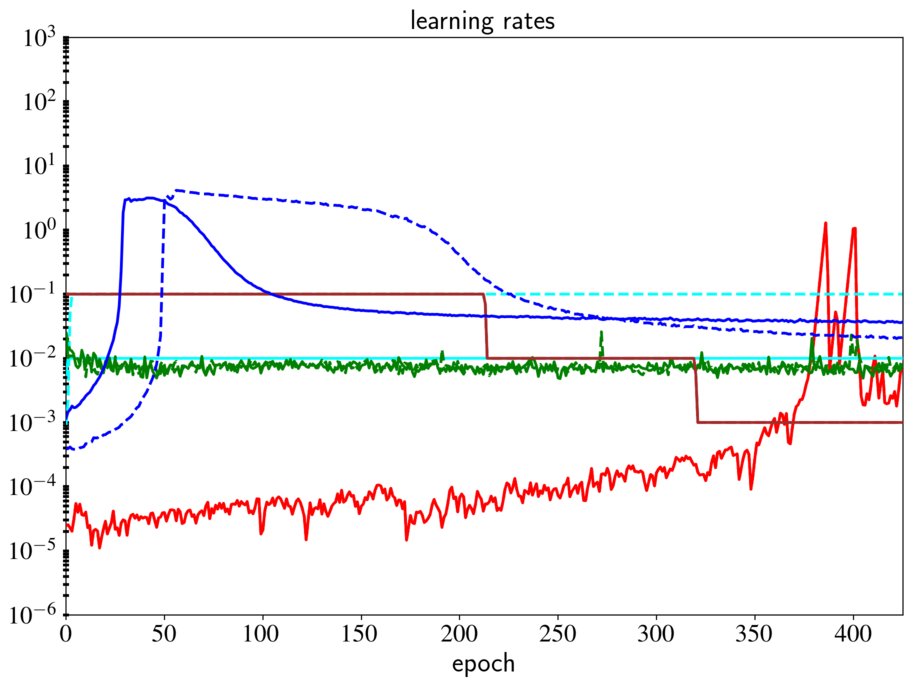}

	\caption[]{Comparison  of \pal\ against  \textit{SLS}, \textit{SGD}, \textit{ADAM}, \textit{RMSProp}, \textit{ALIG}, \textit{SGDHD} and \textit{COCOB} on train. loss (row 1), val. acc. (row 2), test. acc. (row 3) and \textit{SLS}, \textit{SGD}, \textit{ALIG}, \textit{SGDHD} and \pal\ on  learning rates (row 4). Comparison is done across several datasets and models. Further results are found in Appendix \ref{subsec:performance_comparison} Figure (\ref{fig:app_cifar10},\ref{fig.cifar100_step_plots},\ref{fig.IMAGENETplots}). Results are averaged over 3 runs. Box plots result from comprehensive hyperparameter grid searches in plausible intervals. Learning rates are averaged over epochs.   \pal\ surpasses, \textit{ALIG}, \textit{SGDHD}, and \textit{COCOB} and competes against all other optimizers except against SGD.}
	\label{fig.cifar10_step_plots}
	\vspace{-0.5cm}
\end{figure*}

\subsection{Results}
\label{sec:results}
A selection of our results is  given in Figure \ref{fig.cifar10_step_plots}. The results of other architectures trained on CIFAR-10, CIFAR-100, Imagenet and Tolstoi are found in Appendix \ref{sec:further_results} Figures \ref{fig:app_cifar10},\ref{fig.cifar100_step_plots},\ref{fig.IMAGENETplots}.
A table with exact numerical results of all experiments is provided in Appendix \ref{sec:exp_table}.

In most cases \pal\ decreases the training loss faster and to a lower value than the other optimizers (row 1 of Figures \ref{fig.cifar10_step_plots},\ref{fig:app_cifar10},\ref{fig.cifar100_step_plots},\ref{fig.IMAGENETplots}). 
Considering validation and test accuracy, \pal\ surpasses \textit{ALIG}, \textit{SGDHD} and \textit{COCOB}, competes with \textit{RMSProp} and \textit{ADAM} but gets surpassed by \textit{SGD} (rows 2,3 of Figures \ref{fig.cifar10_step_plots},\ref{fig:app_cifar10},\ref{fig.cifar100_step_plots},\ref{fig.IMAGENETplots}). However, \textit{RMSProp}, \textit{ADAM} and \textit{SGD} were tuned with a step size schedule.
If we compare \pal\ to their basic implementations without a schedule, which roughly corresponds to the first plateau reached in row 2 of Figures \ref{fig.cifar10_step_plots},\ref{fig:app_cifar10},\ref{fig.cifar100_step_plots},\ref{fig.IMAGENETplots}, \pal\ would surpass the other optimizers and shows that it can find a well performing step size schedule. This is especially interesting for problems for which default schedules might not work.

\textit{SLS} decreases the training loss further than the other optimizers on a few problems but shows weak performance and poor generalization on most. This contrasts to the results of \cite{backtracking_line_search_NIPS}, where  \textit{SLS} behaves robustly and excels. To exclude the possibility of errors on our side, we reimplemented \textit{SLS} experiment on ResNet34 and could reproduce a similar well performance as in \cite{backtracking_line_search_NIPS} (Appendix \ref{subsec:slsre-impl}). Our results suggest that the interpolation assumption on which \textit{SLS} is based is not always valid for the considered tasks.

Considering the box plots of Figures \ref{fig.cifar10_step_plots} and \ref{fig.cifar100_step_plots}, which represent the sensitivity to hyperparameter combinations, one would likely try on a new unknown objective, we can see that \pal\ has a strong tendency to exhibit low sensitivity in combination with good performance. To emphasize this statement, a sensitivity analysis of \pal's hyperparameters (Appendix Figure \ref{fig:sensitivity analysis}) shows that \pal\ performs well on a wide range for each hyperparameter on a ResNet32.

On wall-clock-time \pal\ performs as fast as \textit{SLS} but slower than the other optimizers, which achieve similar speeds (Appendix \ref{subsec:time_per_epoch}). However, an automatic, well-performing learning rate schedule might compensate for the slower speed depending on the scenario.

Considering the learning rate schedules of \pal\ (row 4 of Figures \ref{fig.cifar10_step_plots},\ref{fig:app_cifar10},\ref{fig.cifar100_step_plots},\ref{fig.IMAGENETplots}) we achieved unexpected results. \pal, which estimates the learning rate directly from approximated local shape information, does not follow a schedule that is similar to the one of \textit{SLS},  \textit{ALIG}, \textit{SGDHD} or any of the commonly used handcrafted schedules such as piece-wise constant or cosine decay. However, it achieves similar results. An interesting side result is that \textit{ALIG} and \textit{SGDHD} tend to perform best if hyperparameters are chosen in a way that the learning rate is only changed slightly and, therefore, virtually an SGD training with a fixed learning rate is performed. 

\vfil
\pagebreak
\section{On the exactness of line searches on mini-batch losses}

\label{sec:optimality}
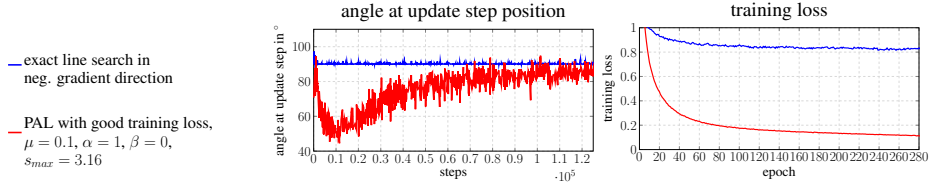
\begin{figure*}[t!]
	\newcommand\picscale{0.3}
	\newcommand\anglemin{40}
	\newcommand\anglemax{110}
	\newcommand\angles{125000}
	\newcommand\lossmin{0}
	\newcommand\lossmax{1}
	\newcommand\evalmin{0}
	\newcommand\evalmax{1}
	\newcommand\epochs{280}
	\newcommand\picheight{0.5\linewidth}
	\newcommand\legendwidth{9cm}
	\newcommand\legendheight{3cm}
	\begin{center}
		
		\tikzsetnextfilename{pal_exact_line_search_1}
		\begin{tikzpicture}[scale=\picscale] 
		\begin{axis}[
		width=\linewidth, 
		height=\picheight,
		grid=major, 
		grid style={dashed,gray!30}, 
		xlabel=steps, 
		xlabel style={font=\LARGE},
		ylabel=angle at update step in $^\circ$,
		ylabel style={font=\LARGE},
		xmin=0,xmax=\angles,
		ymin=\anglemin,ymax=\anglemax,
		legend style={at={(-0.7,1)},anchor=north,font=\huge,minimum height= \legendheight,draw=none}, 
		x tick label style={rotate=0,anchor=near xticklabel,font=\LARGE}, 
		y tick label style={font=\LARGE},
		p1/.style={draw=blue,line width=2pt},
		p2/.style={draw=red,line width=2pt},
		title= angle at update step position,
		title style={font=\Huge},
		]
		\addplot [p1] table[x=Step,y=Value,col sep=comma] {plot_data/optimal_batch_loss_line_search/ol_angle.csv}; 
		\addplot [p2] table[x=Step,y=Value,col sep=comma] {plot_data/optimal_batch_loss_line_search/pal_angle.csv}; 

		\legend{\parbox{\legendwidth}{exact line search in \\ neg. gradient direction}, \parbox{\legendwidth}{PAL with good training loss,\\$\mu=0.1$, $\alpha=1$, $\beta=0$, \\$s_{max}=3.16$}}
		\end{axis}
		\end{tikzpicture}
		\tikzsetnextfilename{pal_exact_line_search_2}
		\begin{tikzpicture}[scale=\picscale] 
		\begin{axis}[
		width=\linewidth, 
		height=\picheight,
		grid=major, 
		grid style={dashed,gray!30}, 
		xlabel style={font=\LARGE},
		ylabel style={font=\LARGE},
		xlabel=epoch, 
		ylabel=training loss,
		xmin=0,xmax=\epochs,
		ymin=\lossmin,ymax=\lossmax,
		legend style={at={(-0.5,0.5)},anchor=north}, 
		x tick label style={rotate=0,anchor=near xticklabel,font=\LARGE}, 
		y tick label style={font=\LARGE},
		p1/.style={draw=blue,line width=1.5pt},
		p2/.style={draw=red,line width=1.5pt},
		title=training loss,
		title style={font=\Huge},
		]		\addplot [p1] table[x=Step,y=Value,col sep=comma] {plot_data/optimal_batch_loss_line_search/ol_train_loss.csv}; 
		\addplot [p2] table[x=Step,y=Value,col sep=comma] {plot_data/optimal_batch_loss_line_search/pal_train_loss.csv}; 
		\end{axis}
		\end{tikzpicture}
	\end{center}
	\caption{Comparison of \pal\ against an exact line search. The first plot shows the angle between the direction and gradient vector at the update step position. A ResNet32 was trained on CIFAR-10. One can observe that an exact line search exhibits poor performance.
	}
\label{fig:exactlinesearch}
\end{figure*}
\begin{figure*}[b!]
	\centering
	\newcommand\disfigwidth{0.25}
	\vspace{-0.3cm}
	\includegraphics[width=\disfigwidth\linewidth]{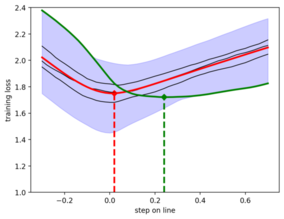}
	\includegraphics[width=\disfigwidth\linewidth]{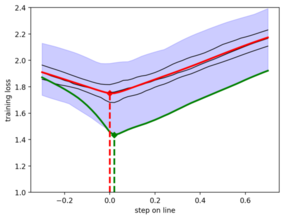}
	\includegraphics[width=\disfigwidth\linewidth]{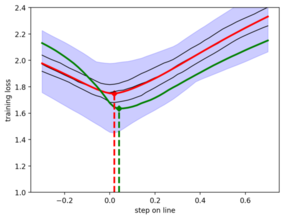}\\
	\vspace{-0.1cm}
	\includegraphics[width=\disfigwidth\linewidth]{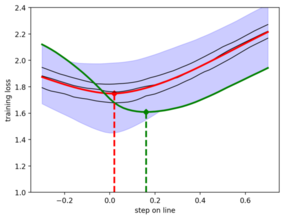}
	\includegraphics[width=\disfigwidth\linewidth]{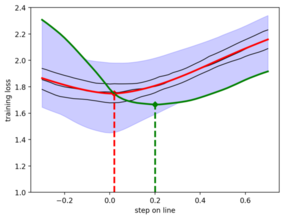}
	\includegraphics[width=\disfigwidth\linewidth]{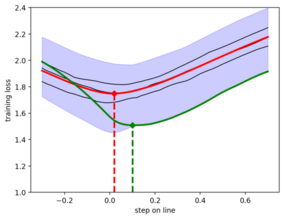}\\
\caption{Distributions (\textcolor{BlueViolet}{blue}) overall mini-batch losses along representative lines during a training of a ResNet32 on CIFAR-10. The full-batch loss, which is the mean value of the distribution, is given in \textcolor{red}{red}.  Quartiles are given in black. The mini-batch loss, whose negative gradient defines the search direction, is given in \textcolor{ForestGreen}{green}. It can be observed that the minimum of the green mini-batch loss is not always an adequate estimator of the minimum of the full-batch loss along the line.}
\label{fig:distributionplots}
\end{figure*}

In this section, we investigate the general question of whether line searches that estimate the position of a minimum of mini-batch losses exactly are beneficial. In Figure \ref{fig:angles} we showed that \pal\ can perform an almost exact line search on batch losses if we use a fixed update step adaptation factor (Section \ref{subsec:features}). However, \pal's best hyperparameter configuration does not perform an exact line search (see Figure \ref{fig:exactlinesearch}).
Consequently, we analyzed how an exact line search, which exactly estimates a minimum along the line, behaves. We implemented an inefficient binary line search (see Appendix \ref{sec:binary_line_search}), which measured up to 20 values on each line to estimate the position of a minimum. The results, given in Figure \ref{fig:exactlinesearch}, show that an optimal line search does not optimize well. Thus, the reason why \pal\ performs well is not the exactness of its update steps. In fact, slightly inexact update steps seem to be beneficial.\\
These results query Assumption \ref{ass:line_min},
which assumes that the position of a minimum on a line in negative gradient direction of $\BL$  is a suitable estimator for the minimum of the full-batch loss $\mathcal{L}$ on this line to perform a successful optimization process. To investigate this further, we measured $\mathcal{L}$ and the distribution of mini-batch losses for multiple SGD update directions on a ResNet32. Our results suggest, as exemplarily shown in Figure \ref{fig:distributionplots} that the position of $\BL[i]$'s minimum along a line in negative gradient direction is not always a good estimator for the position of $\mathcal{L}$'s corresponding minimum. This explains why exact line searches on the batch loss perform weakly.

Corollaries are that the empirical loss on the investigated lines also tends to be locally convex and that the optimal step size tends to be smaller than the step size given by the batch loss on such lines. This is a possible explanation why the slightly too narrow parabolic approximations of \pal\ without update step adaptation perform well.

\section{PAL and Interpolation}
\label{sec:interpolation_analysis}
This section analyzes whether the reason why \pal\ performs well is related to the interpolation condition. Formally, interpolation requires that the gradient with respect to each sample converges to zero at the optimum. We repeated the experiments of the \textit{SLS} paper (see \cite{backtracking_line_search_NIPS} Section 7.2 and 7.3), which analyze the performance on problems for which interpolation hold or does not hold. 

\begin{figure}[h]
\includegraphics[width=0.24\linewidth]{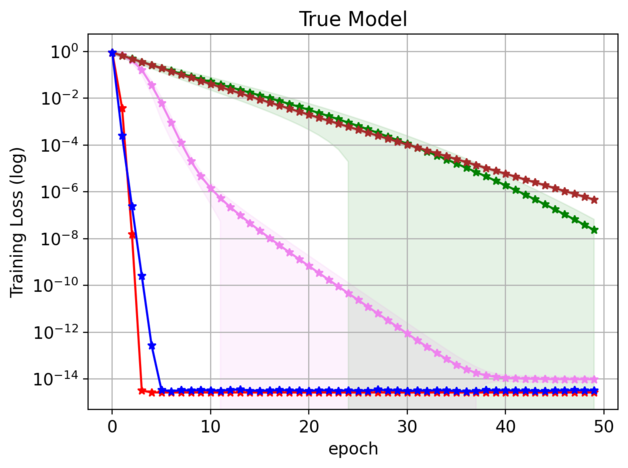}
\includegraphics[width=0.24\linewidth]{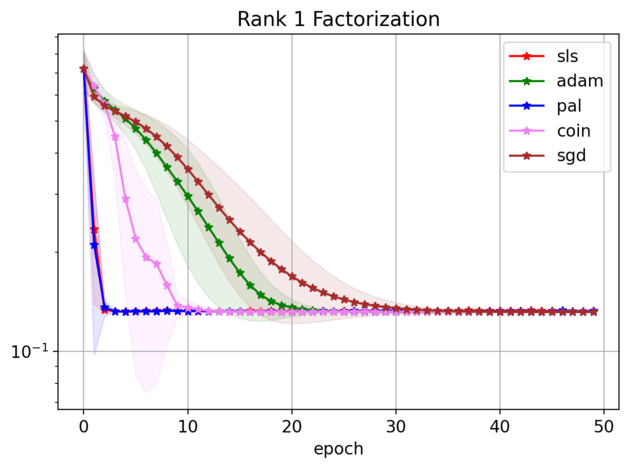}
\includegraphics[width=0.24\linewidth]{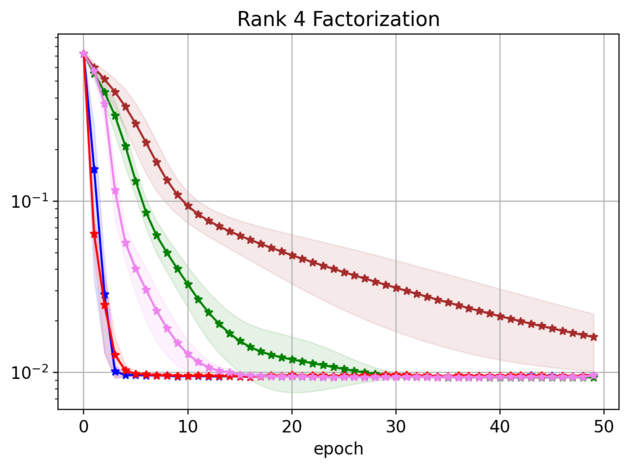}
\includegraphics[width=0.24\linewidth]{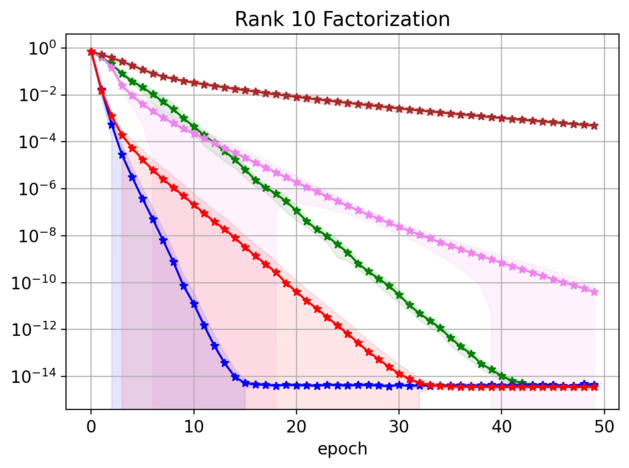}
\caption{The matrix factorization problem of \cite[\S 7.2]{backtracking_line_search_NIPS} . For $k=1$ and $k=4$ interpolation does not hold. Rank 1 factorization is under-parameterized, whereas rank 4 and rank 10 factorizations are over-parameterized.}
\label{fig:matrix_fac}
\end{figure}
\begin{figure}[h]
\includegraphics[width=0.24\linewidth]{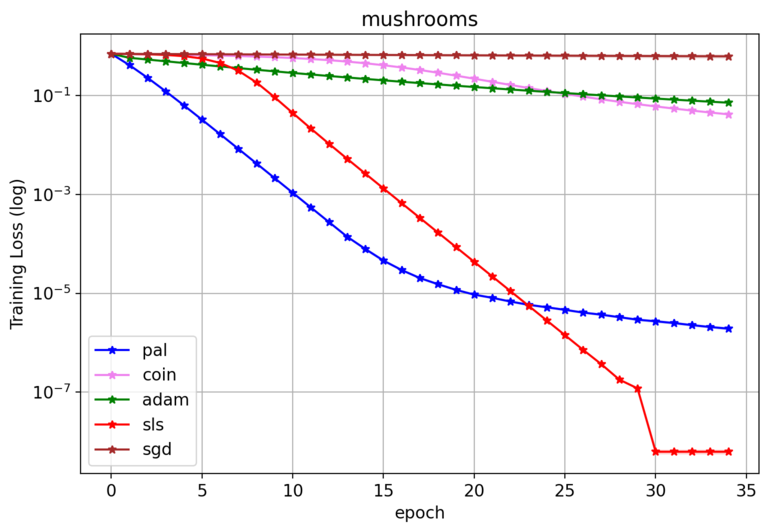}
\includegraphics[width=0.24\linewidth]{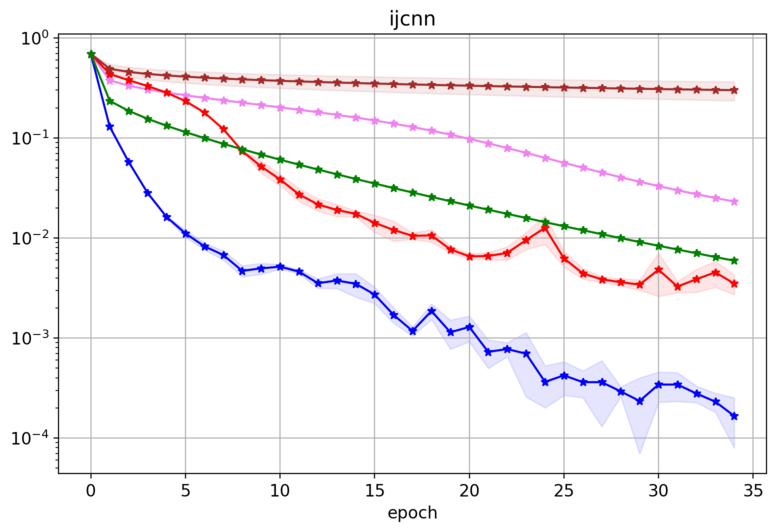}
\includegraphics[width=0.24\linewidth]{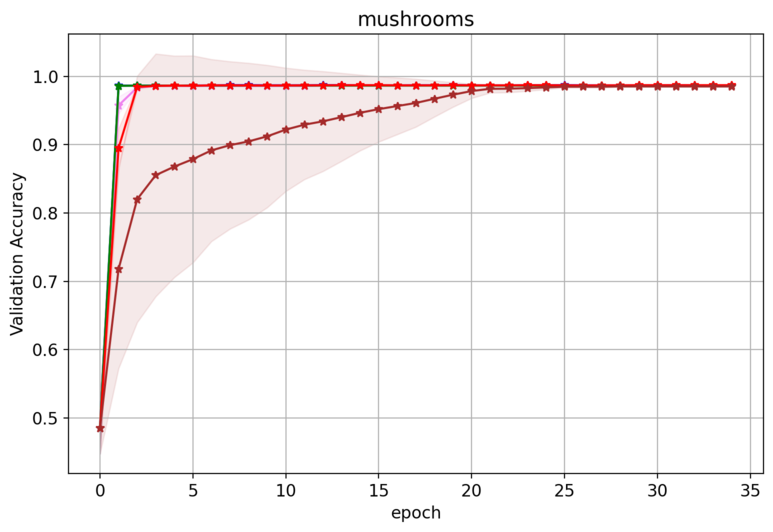}
\includegraphics[width=0.24\linewidth]{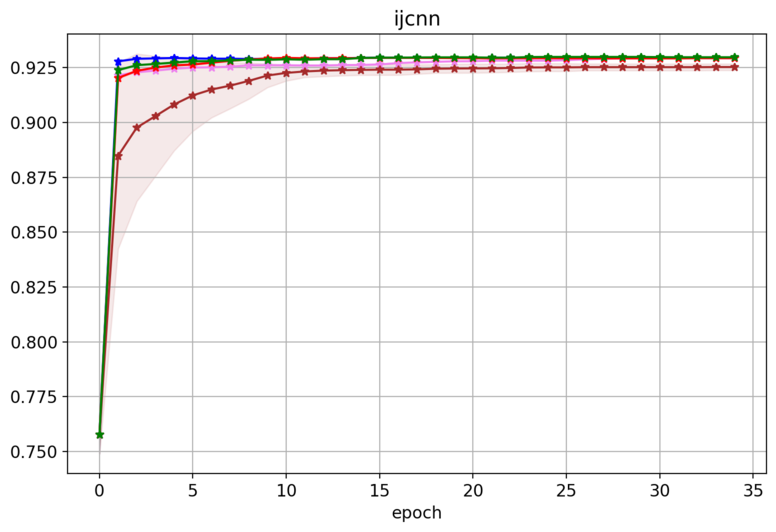}

\caption{Binary classification task of \cite[\S 7.3]{backtracking_line_search_NIPS} using a softmax loss and RBF kernels for mushrooms and ijcnn datasets. With RBF kernels, the mushrooms dataset is linear separable in kernel-space with the selected kernel bandwidths, while the ijcnn dataset is not.}
\label{fig:mushroom}
\end{figure}

Figure \ref{fig:matrix_fac} shows that \pal\ such as \textit{SLS} converge faster to an artificial optimization floor on non-over-parameterized models ($k=4$) of the matrix factorization problem  of \cite[\S 7.2]{backtracking_line_search_NIPS}. In the interpolation case  \pal\ and \textit{SLS} converge linearly to machine precision. 
On the binary classification problem of \cite{backtracking_line_search_NIPS} Section 7.3, which uses a softmax loss and RBF kernels on the mushrooms and ijcnn datasets, we observe that \pal\ and \textit{SLS} converge fast on the mushrooms task, for which the interpolation condition holds (Figure \ref{fig:mushroom}). However, \pal\ converges faster on the ijcnn task, for which the interpolation condition does not hold.

The results indicate that the interpolation condition is beneficial for \pal, but, \pal\ performs also robust when it is likely not satisfied (see Figure  \ref{fig.cifar10_step_plots},\ref{fig:app_cifar10},\ref{fig.cifar100_step_plots},\ref{fig.IMAGENETplots}. In those experiments \pal\ mostly performs competitive but \textit{SLS} does not. However, the relation of the parabolic observation to interpolation needs to be investigated more closely in future.

\section{Conclusions}
This work tackles a major challenge in current optimization research for deep learning: to automatically find optimal step sizes for each update step. In detail, we focus on line search approaches to deal with this challenge. We introduced a robust and straightforward line search approach based on one-dimensional parabolic approximations of mini-batch losses. The introduced algorithm is an alternative to \textit{SGD} for objectives where default decays are unknown or do not work.

Loss functions of DNNs are commonly perceived as being highly non-convex. Our analysis suggests that this intuition does not hold locally since lines of loss landscapes across models and datasets can be approximated parabolically to high accuracy. This new knowledge might further help to explain why update steps of specific optimizers perform well.

To gain deeper insights into line searches in general, we analyzed how an expensive but exact line search on batch losses behaves. Intriguingly, its performance is weak, which lets us conclude that the minor inaccuracies of the parabolic approximations are beneficial for training.\\

\section*{Retrospective from 10.2021}
After the publication of this paper, we conducted further research in this field and observed the following: \pal's performance is highly dependent on the weight initialization and batch size.
In general, it does not work well with Pytorch's default weight initialization. Note that the experiments here were done with Tensorflow 1.5. In addition, it is hard to find hyperparameters that compete with SGD. Our further works exploring the empirics-based line search direction are \cite{line_analysis,labpal}.\cite{line_analysis} performs a comprehensive empirical study of the shape of the full-batch loss in line direction and the performance of several optimizers along such lines. \cite{labpal} introduces a line search approach that fits the full-batch loss with a parabola and can also handle smaller batch sizes.

\section*{Ethics Statement}
 Since we understand our work as basic research, it is highly error-prone to estimate its \textit{specific} ethical aspects and future positive or negative social consequences. As optimization research influences the whole field of deep learning, we refer to the following works, which discuss the ethical aspects and social consequences of AI and Deep Learning in a comprehensive and general way:\cite{yudkowsky2008artificial,muehlhauser2012singularity,bostrom2014ethics}.
\section*{Acknowledgments}
Maximus Mutschler heartly thanks Lydia Federmann, Kevin Laube, Jonas Tebbe, Mario Laux, Valentin Bolz, Hauke Neitzel, Leon Varga, Benjamin Kiefer, Timon H\"ofer, Martin Me{\ss}mer, Cornelia Schulz, Hamd Riaz, Nuri Benbarka, Samuel Scherer, Frank Schneider, Robert Geirhos and Frank Hirschmann  for their comprehensive support.

\section*{Funding}
This research was supported by the German Federal Ministry of Education and Research (BMBF) project 'Training Center Machine Learning, T\"ubingen'  with grant number 01|S17054.

{\small
	\nocite{*}
	\bibliographystyle{ieee}
	\bibliography{neurips_2020}
}

\vfill
\pagebreak
\onecolumn
\appendix

\section{Further line plots}
\label{sec:further_line_plots}
\begin{figure*}[h!]
	\centering
		\vspace{-0.2cm}
	\includegraphics[width=0.19\linewidth]{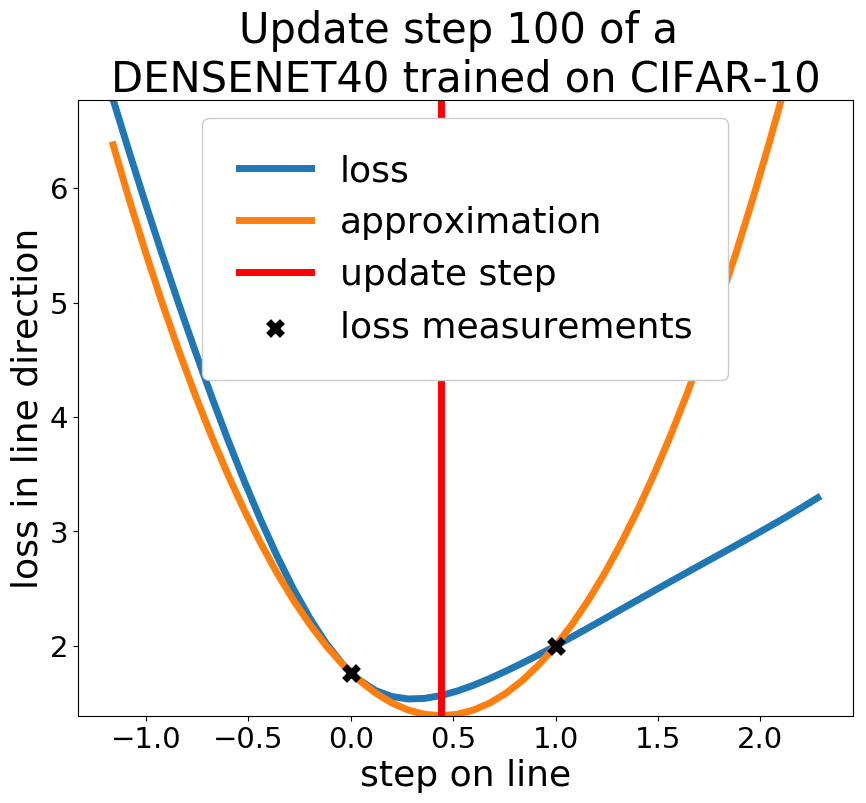}
	\includegraphics[width=0.19\linewidth]{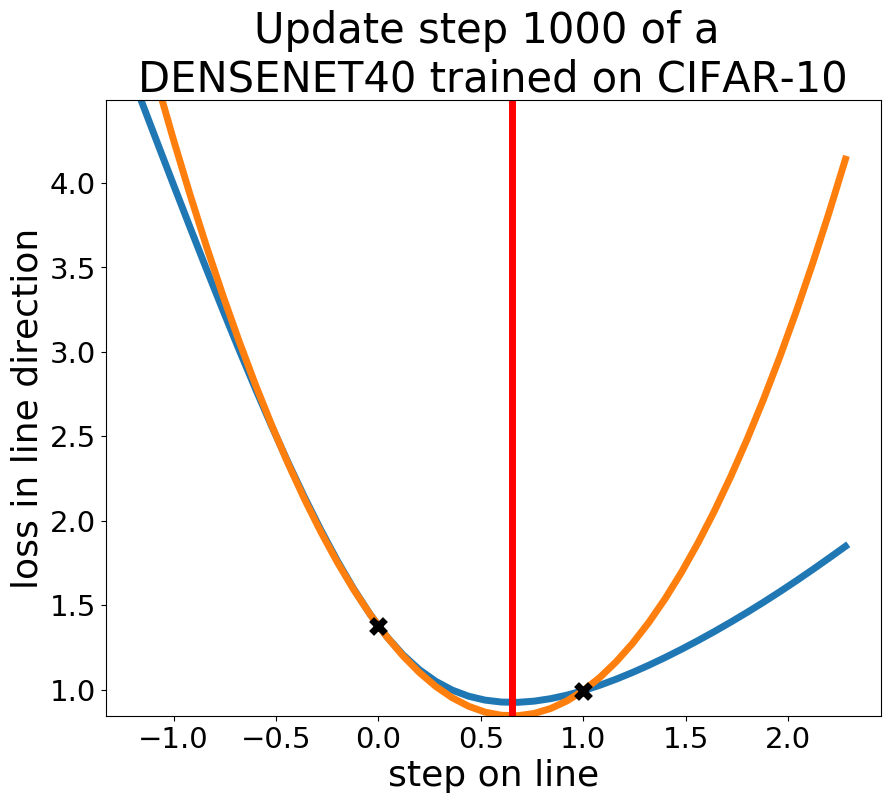}
	\includegraphics[width=0.19\linewidth]{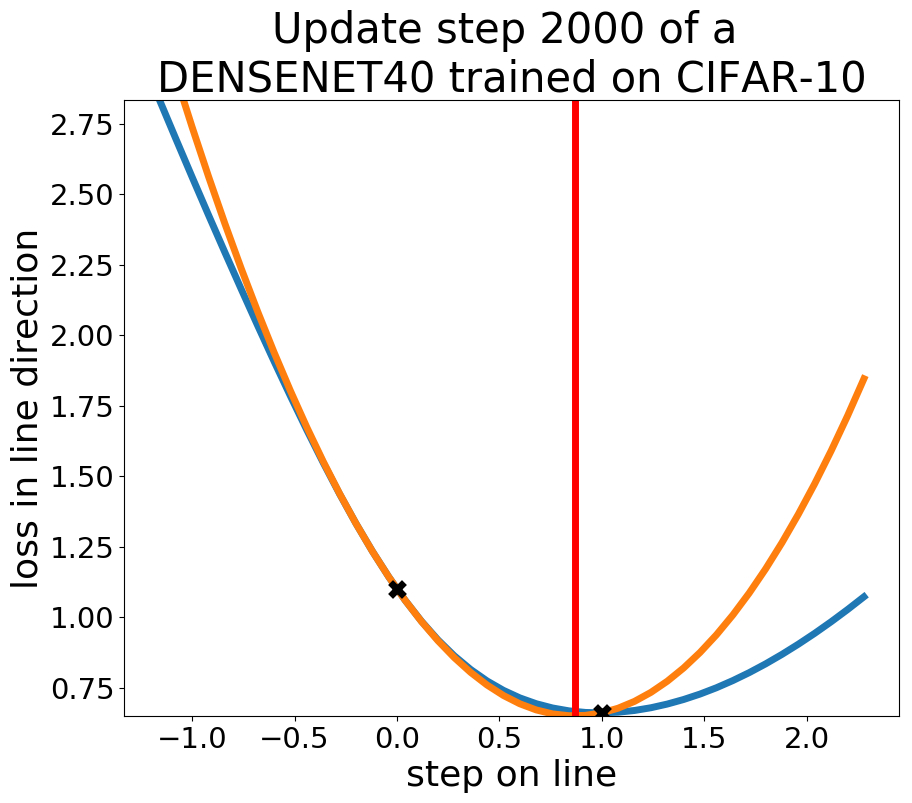}
	\includegraphics[width=0.19\linewidth]{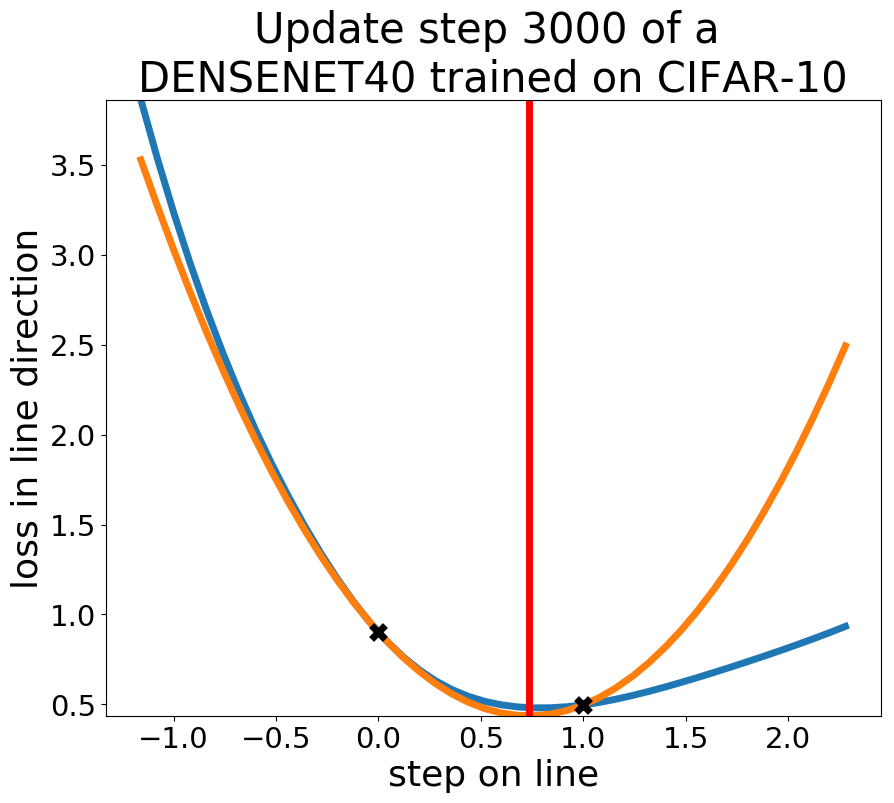}
	\includegraphics[width=0.19\linewidth]{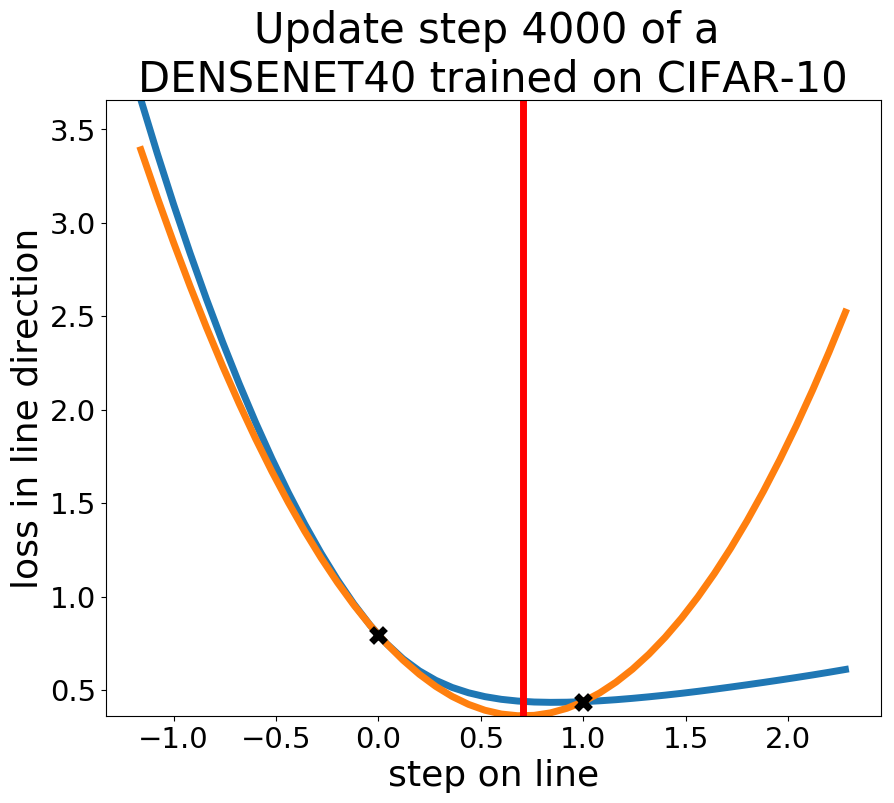}\\
	\includegraphics[width=0.19\linewidth]{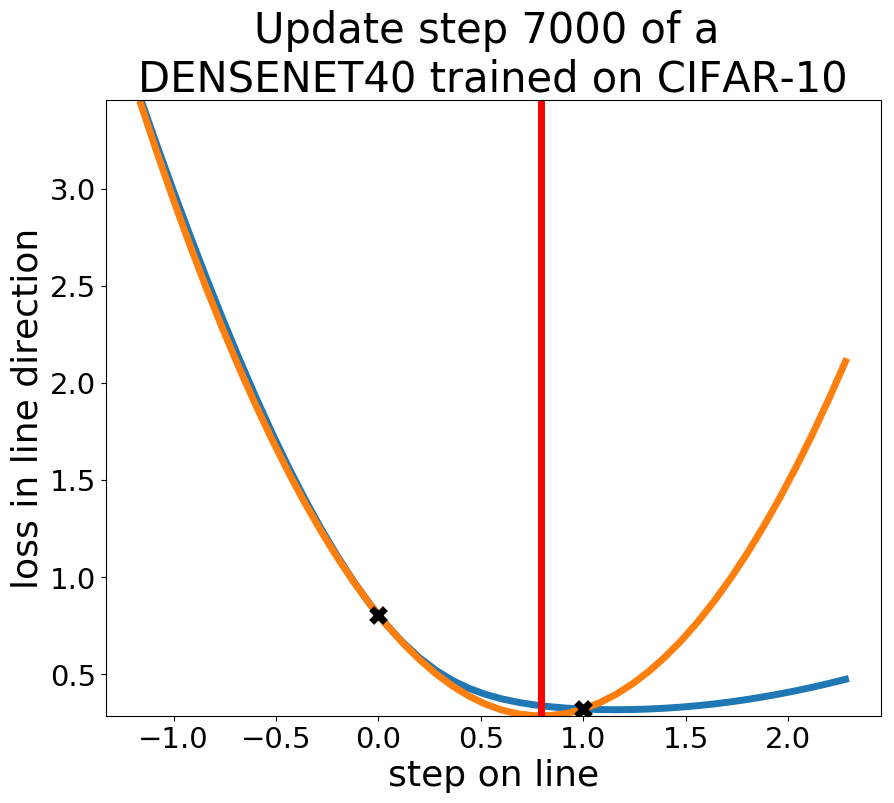}
	\includegraphics[width=0.19\linewidth]{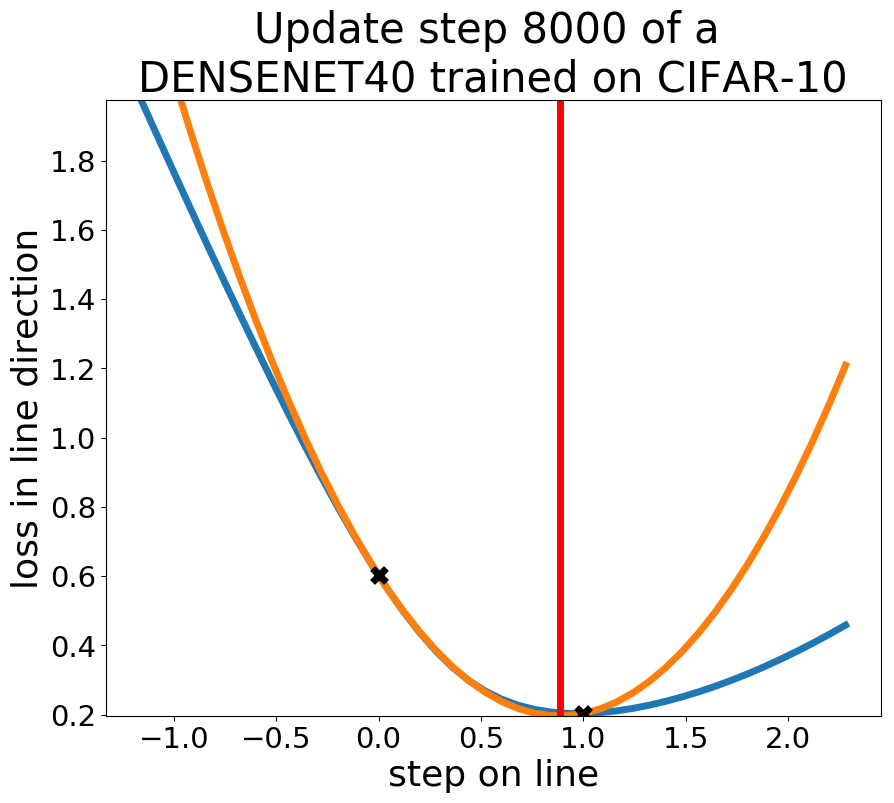}
	\includegraphics[width=0.19\linewidth]{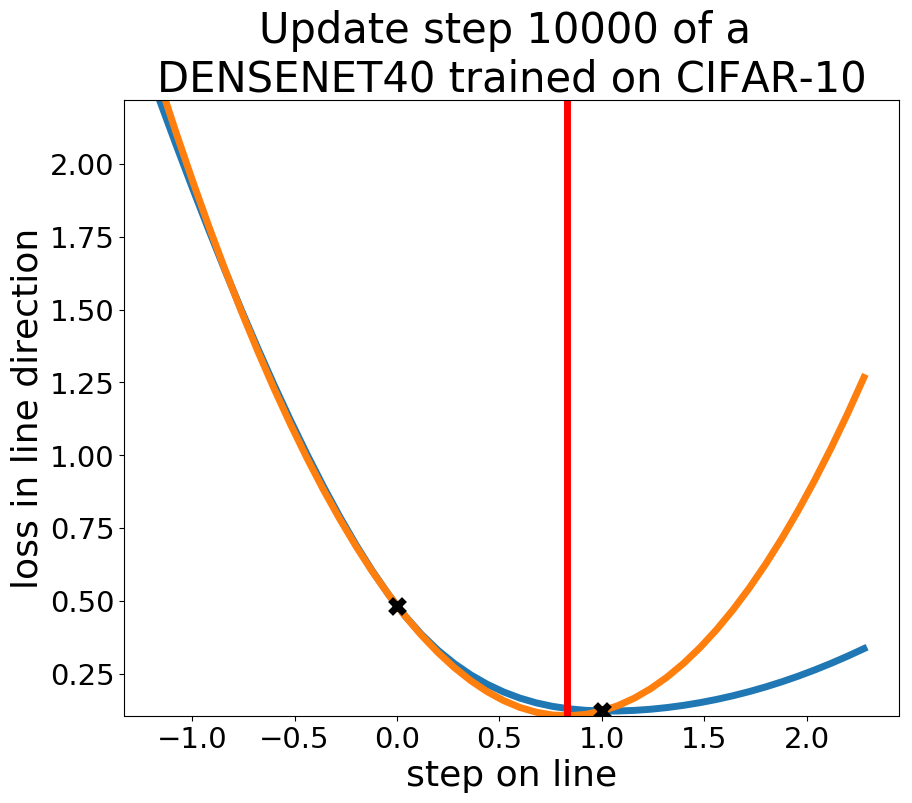}
	\includegraphics[width=0.19\linewidth]{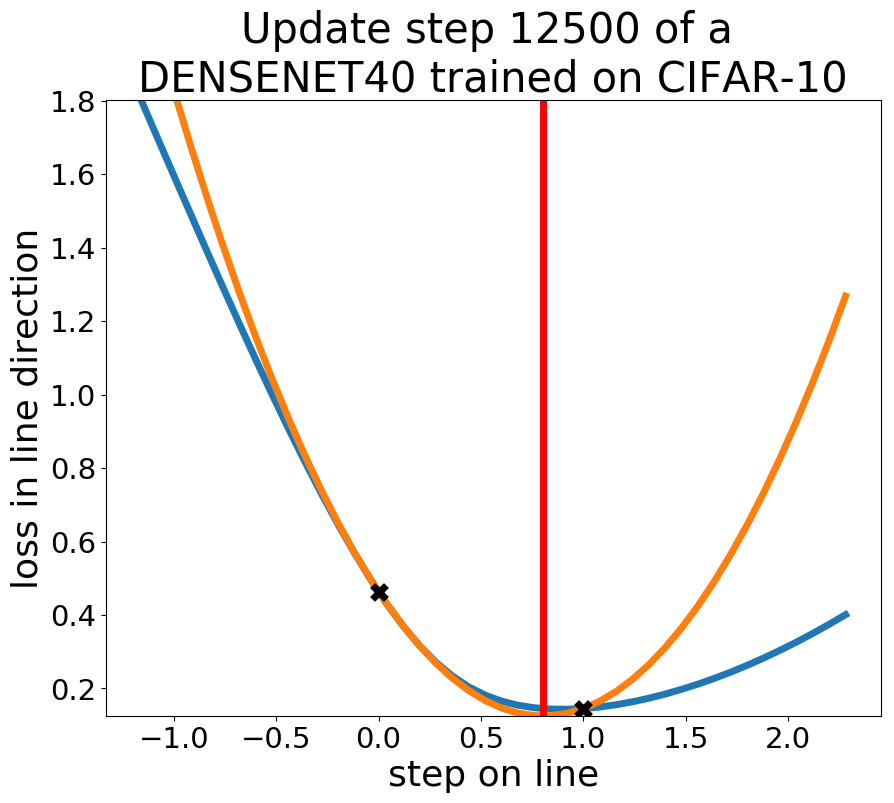}
	\includegraphics[width=0.19\linewidth]{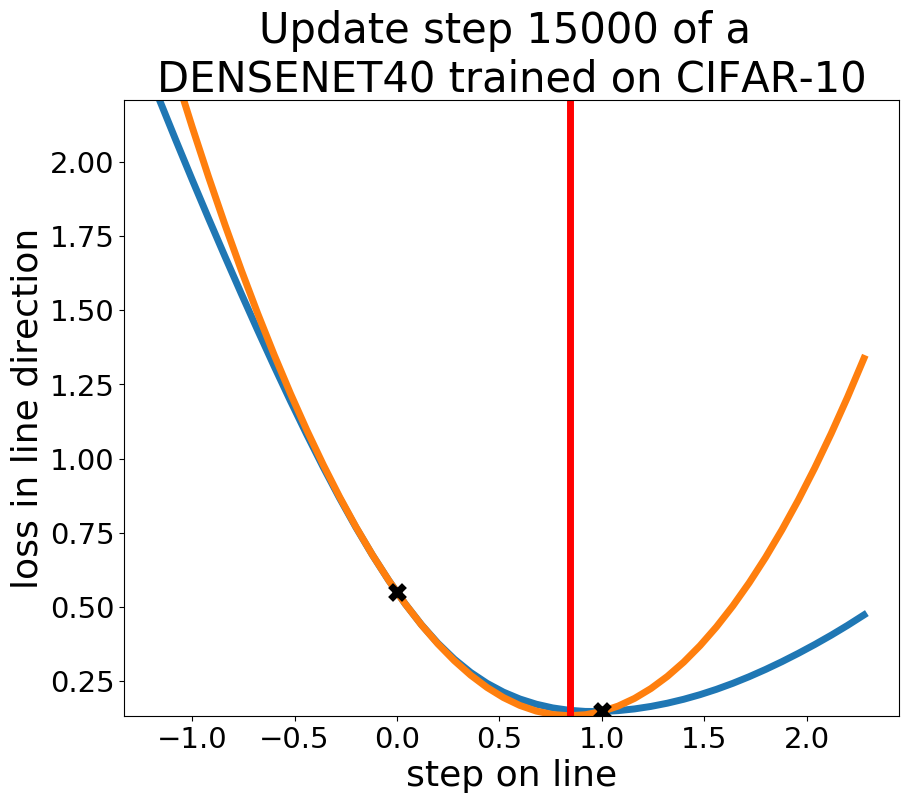}
	\caption[]{\textbf{DenseNet40} mini-batch losses along in negative gradient direction (\textcolor{blue}{blue}) combined with our parabolic approximation (\textcolor{orange}{orange}) and the position of the minimum (\textcolor{red}{red}). The unit of the horizontal axis is the change of $\theta$.}
	\vspace{-0.5cm}
	\label{fig_loss_lines_dense_net_40}
\end{figure*}

\begin{figure*}[h!]
	\centering
	\includegraphics[width=0.19\linewidth]{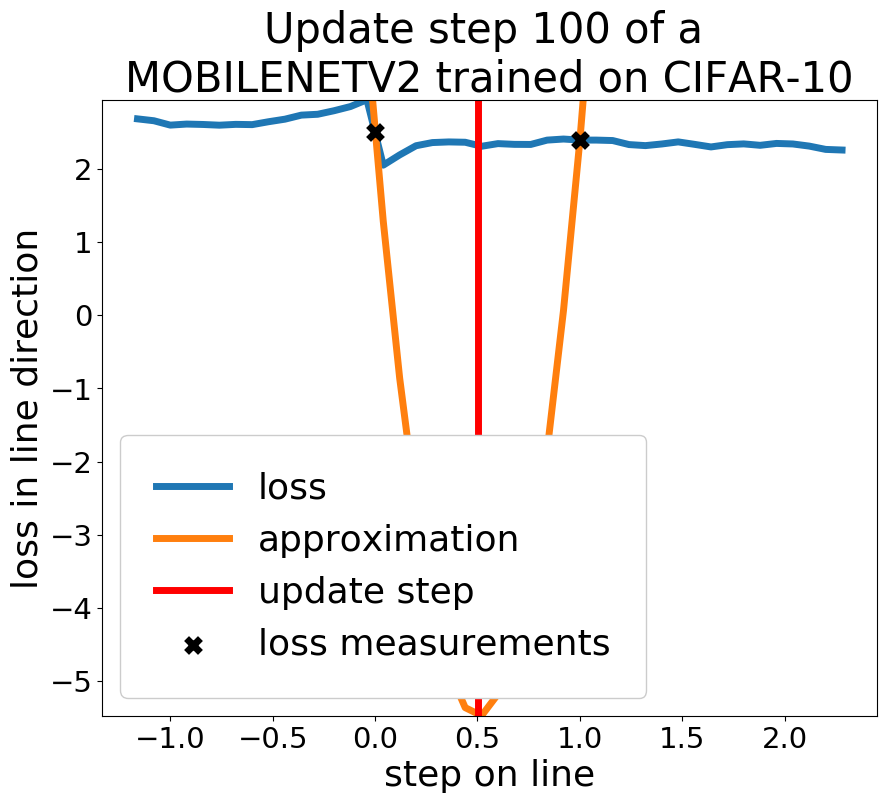}
	\includegraphics[width=0.19\linewidth]{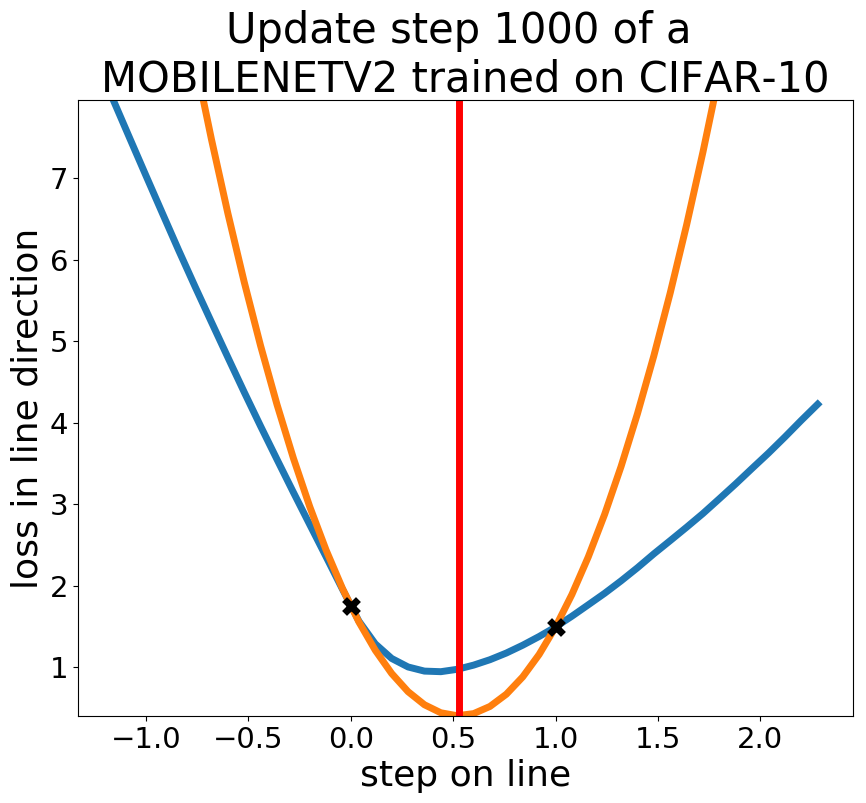}
	\includegraphics[width=0.19\linewidth]{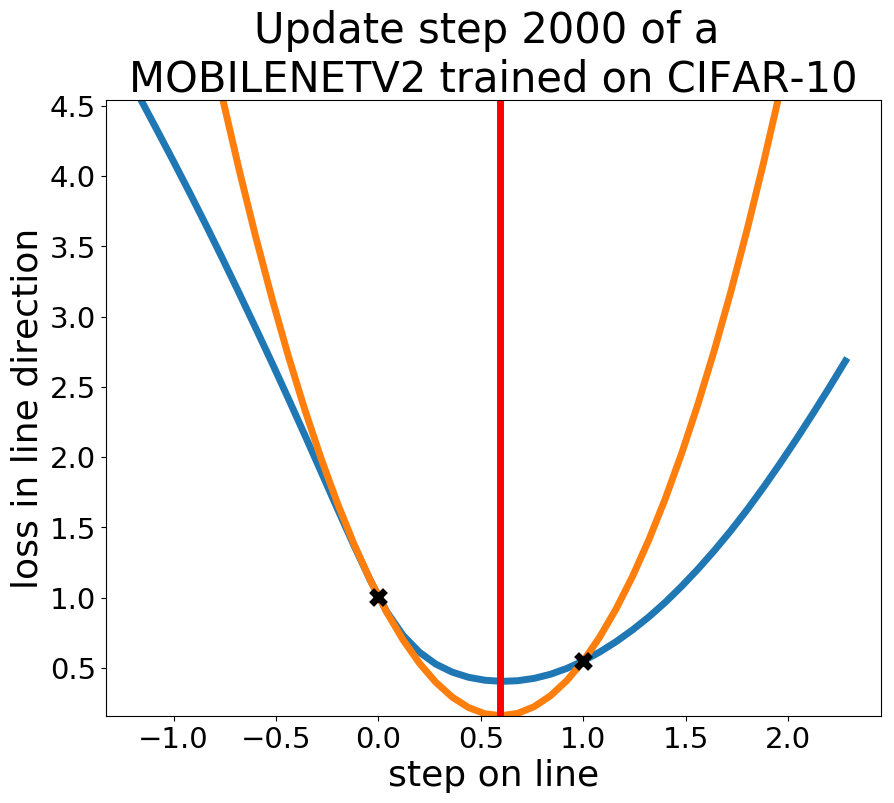}
	\includegraphics[width=0.19\linewidth]{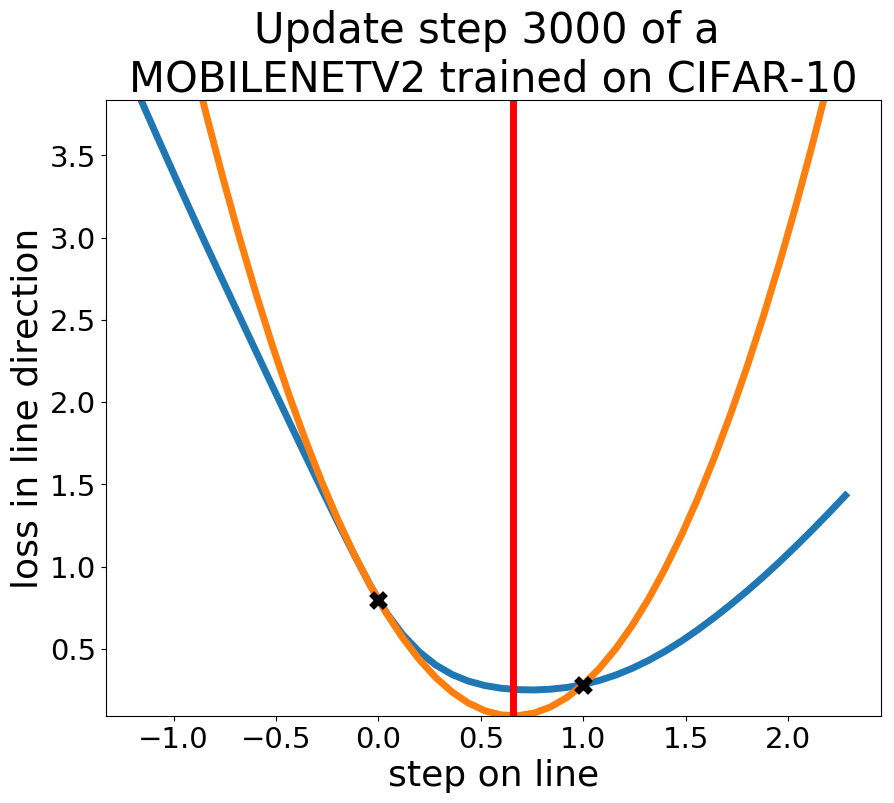}
	\includegraphics[width=0.19\linewidth]{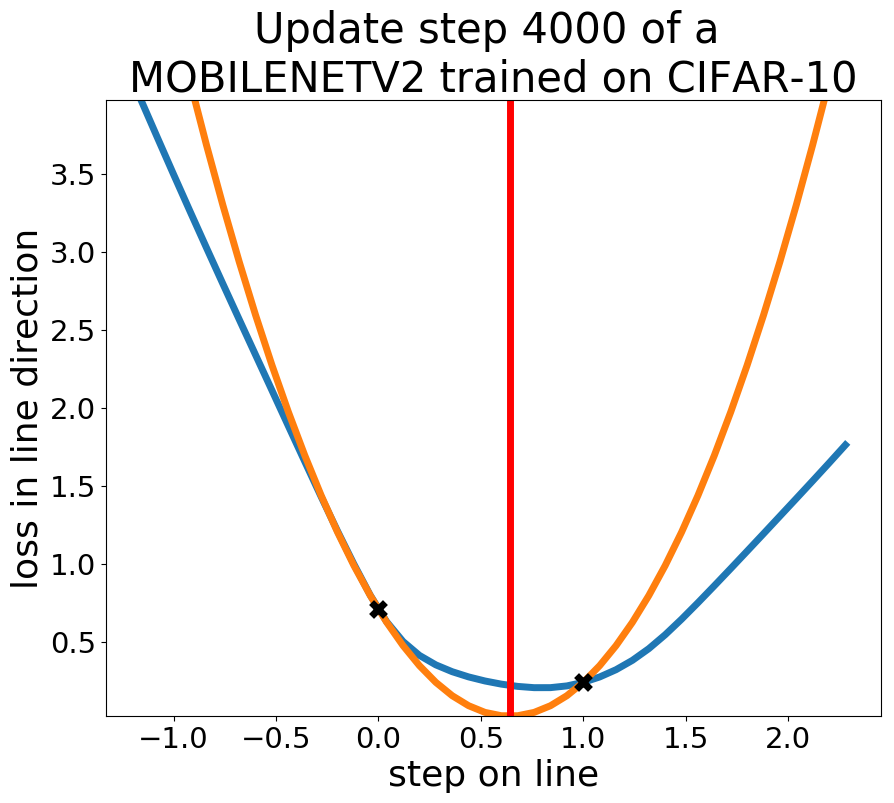}\\
	\includegraphics[width=0.19\linewidth]{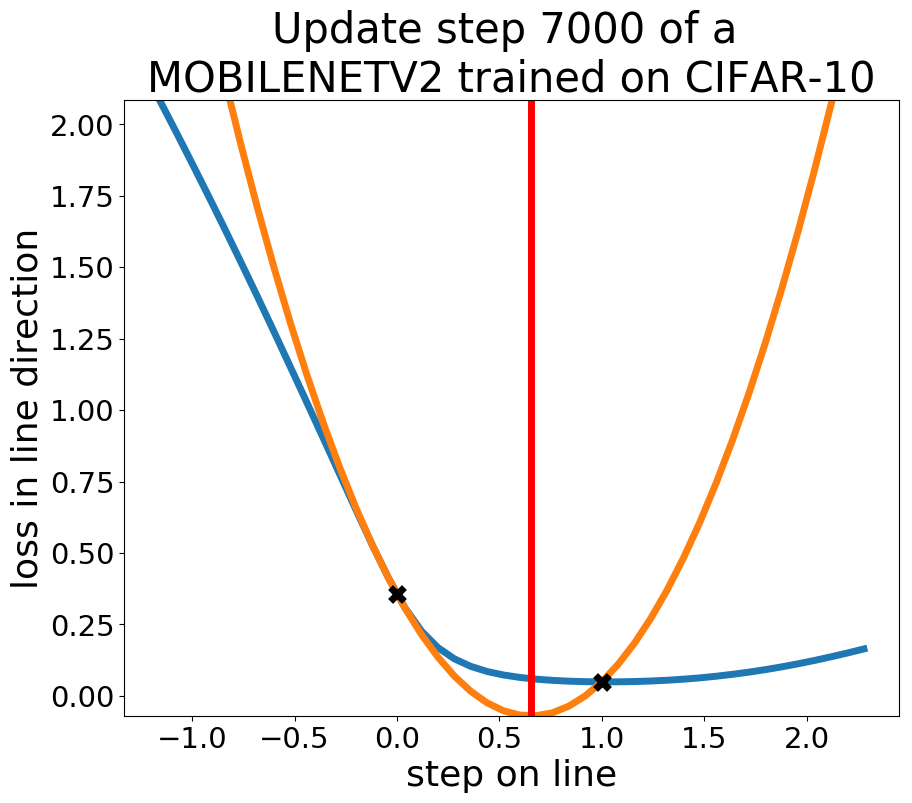}
	\includegraphics[width=0.19\linewidth]{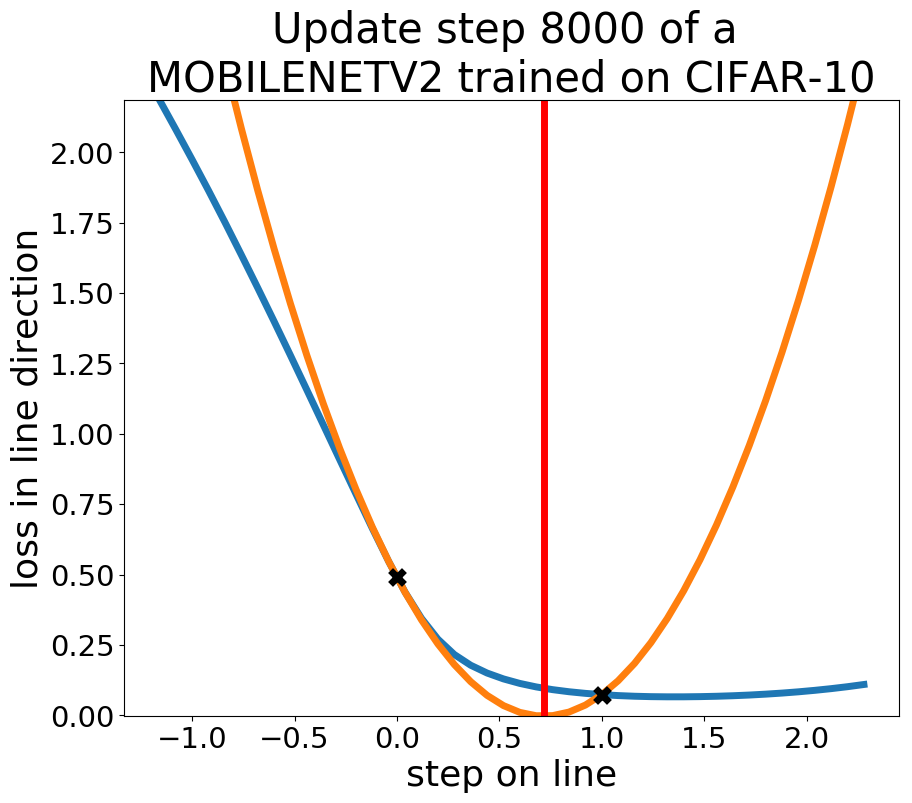}
	\includegraphics[width=0.19\linewidth]{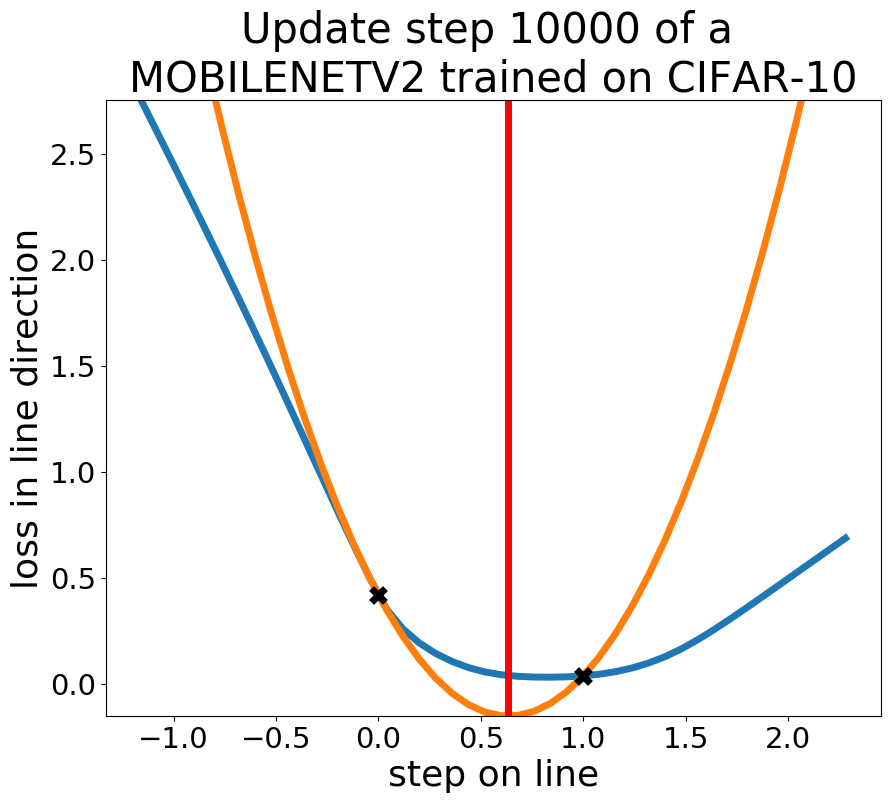}
	\includegraphics[width=0.19\linewidth]{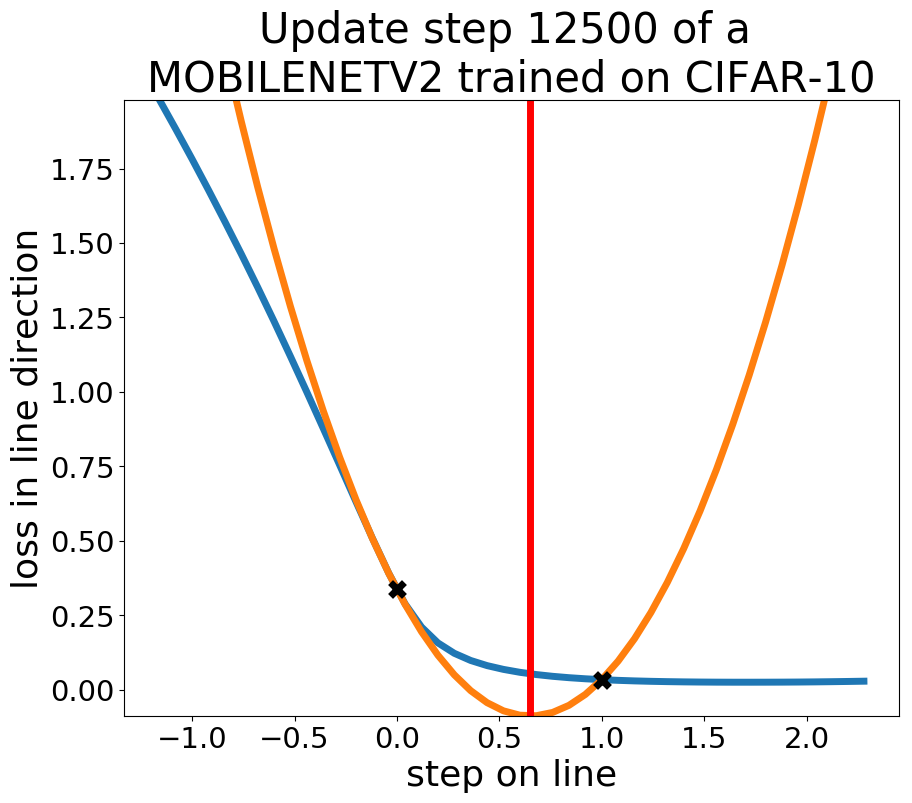}
	\includegraphics[width=0.19\linewidth]{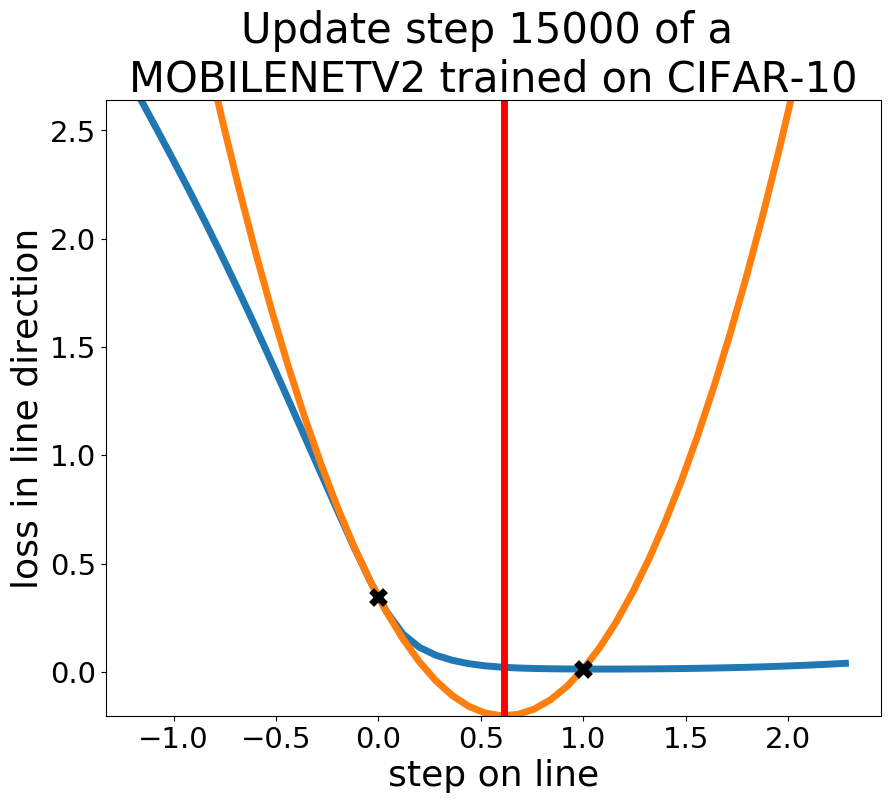}
	\caption[]{ Mini-batch losses along lines of \textbf{MobileNetV2}. For explanations see Figure \ref{fig_loss_lines_dense_net_40}. During training the parabolic approximation fits less accurately on the right hand side and during the first 150 steps it does not fit at all, however, the minimum of the parabola is still a good estimator for a low loss value on the line.}
\end{figure*}

\begin{figure*}[h!]
	\centering
	\includegraphics[width=0.19\linewidth]{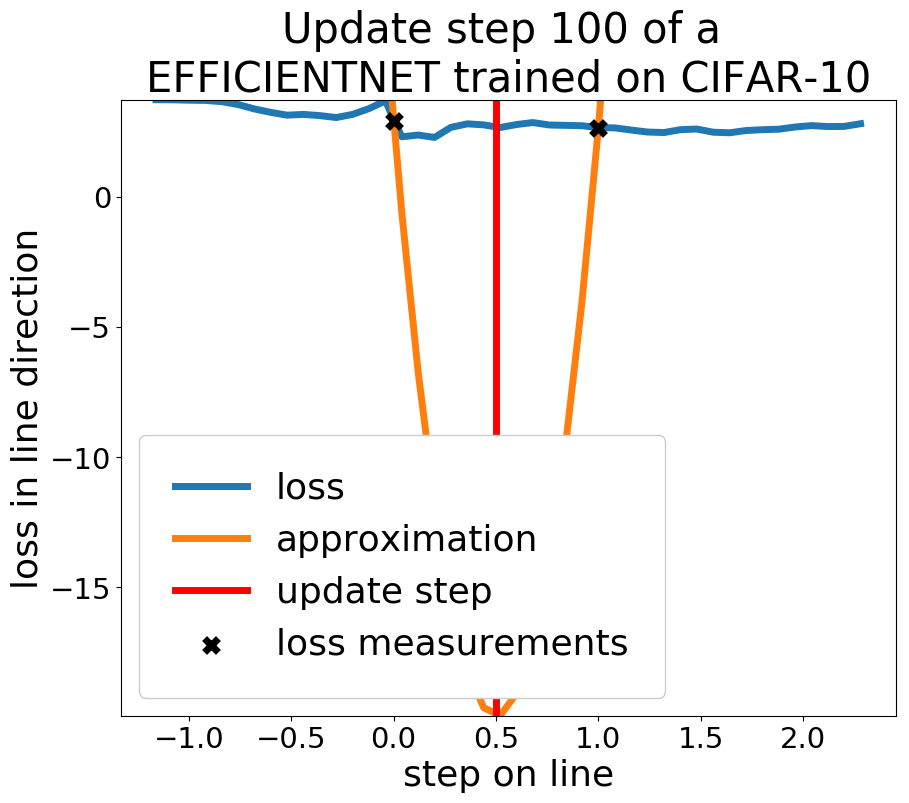}
	\includegraphics[width=0.19\linewidth]{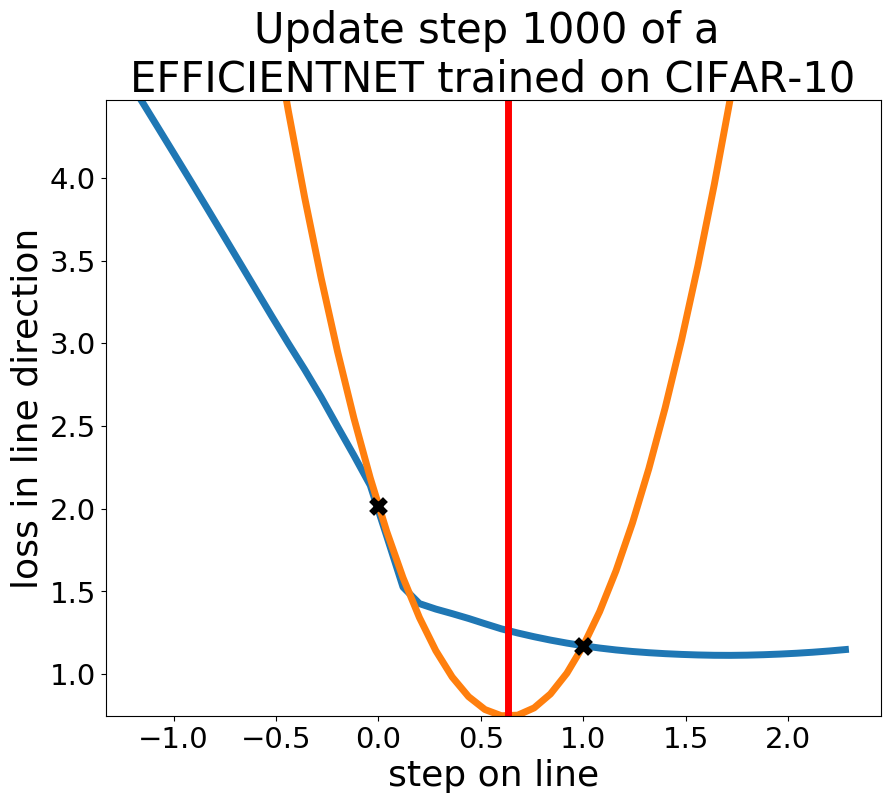}
	\includegraphics[width=0.19\linewidth]{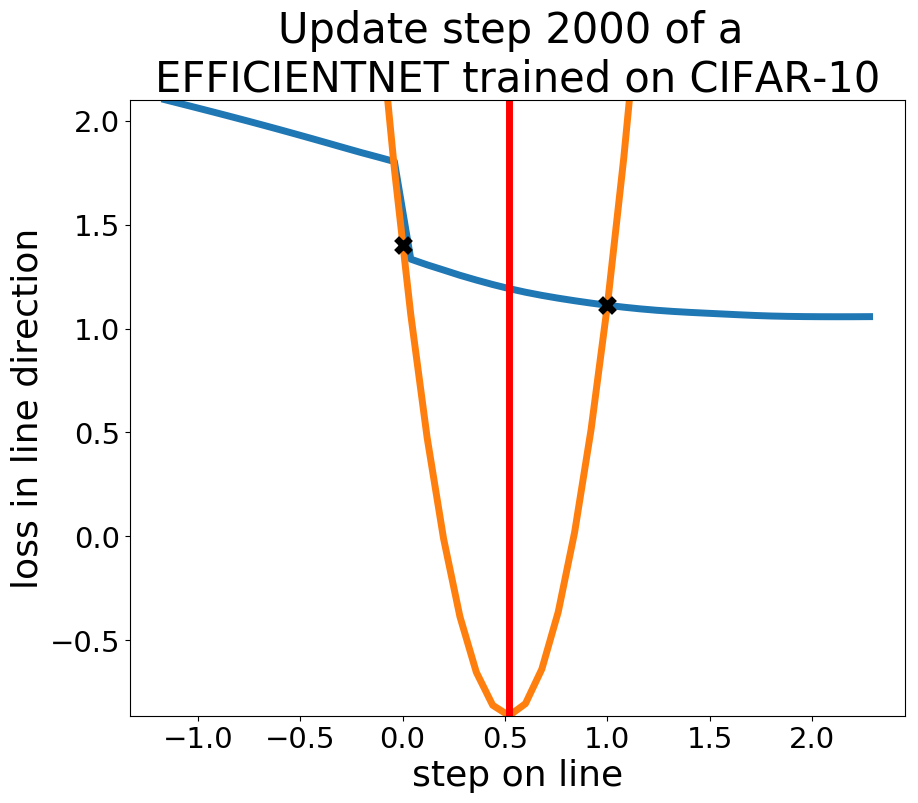}
	\includegraphics[width=0.19\linewidth]{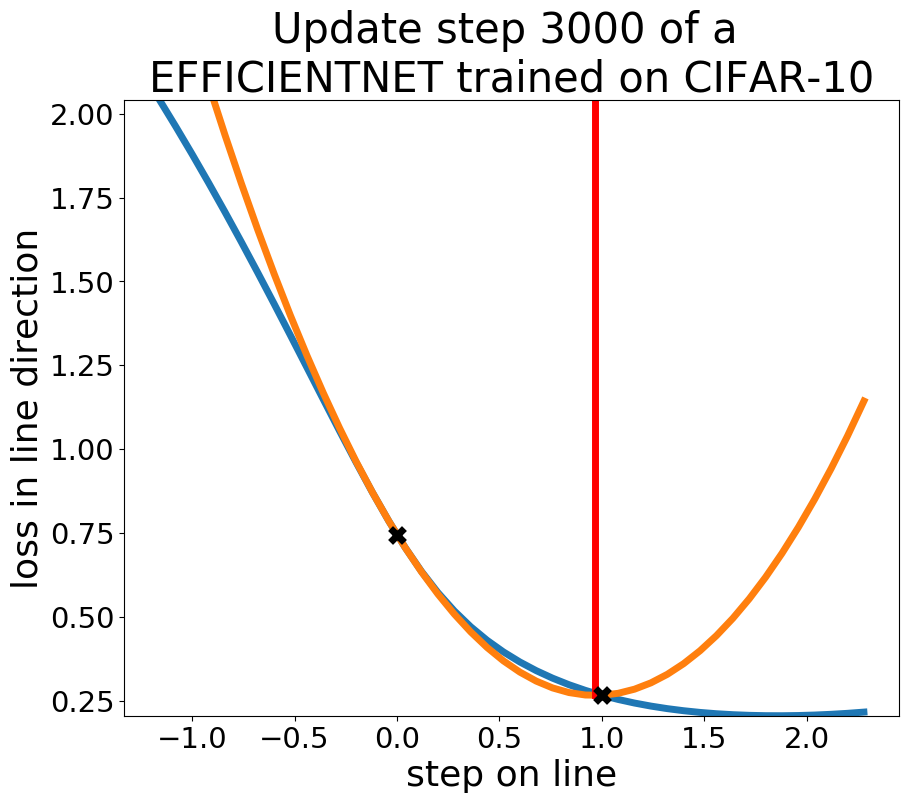}
	\includegraphics[width=0.19\linewidth]{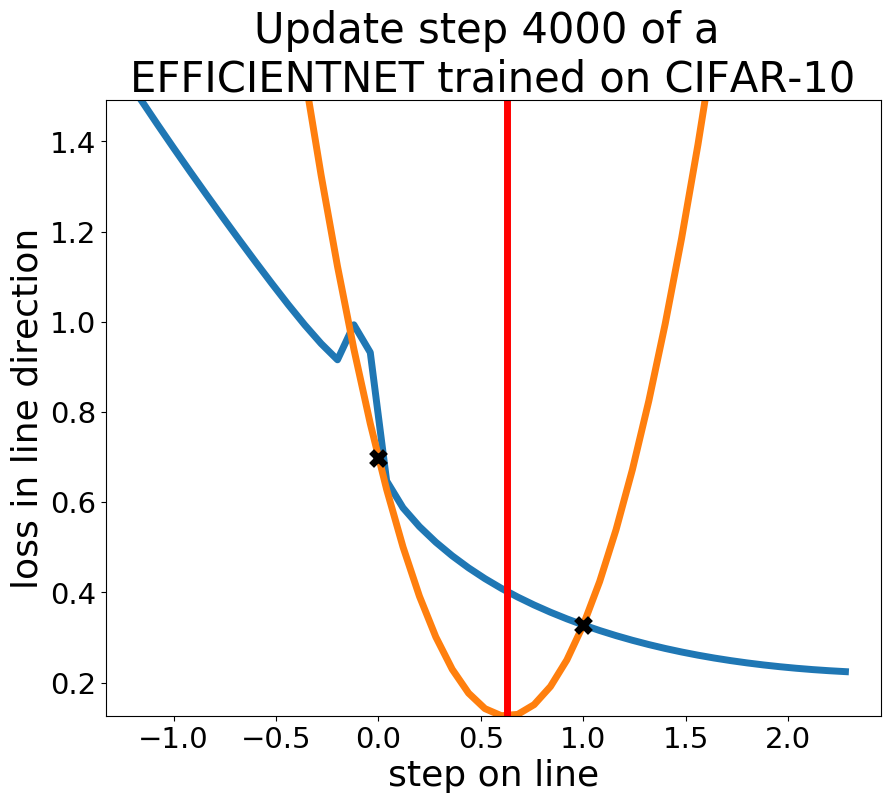}
	\includegraphics[width=0.19\linewidth]{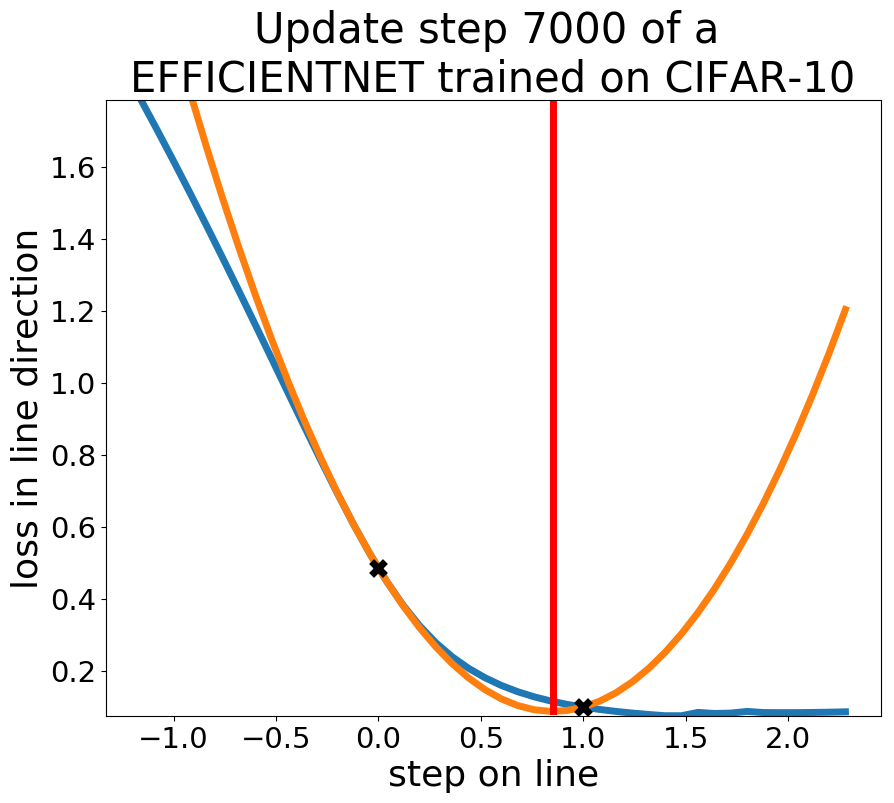}
	\includegraphics[width=0.19\linewidth]{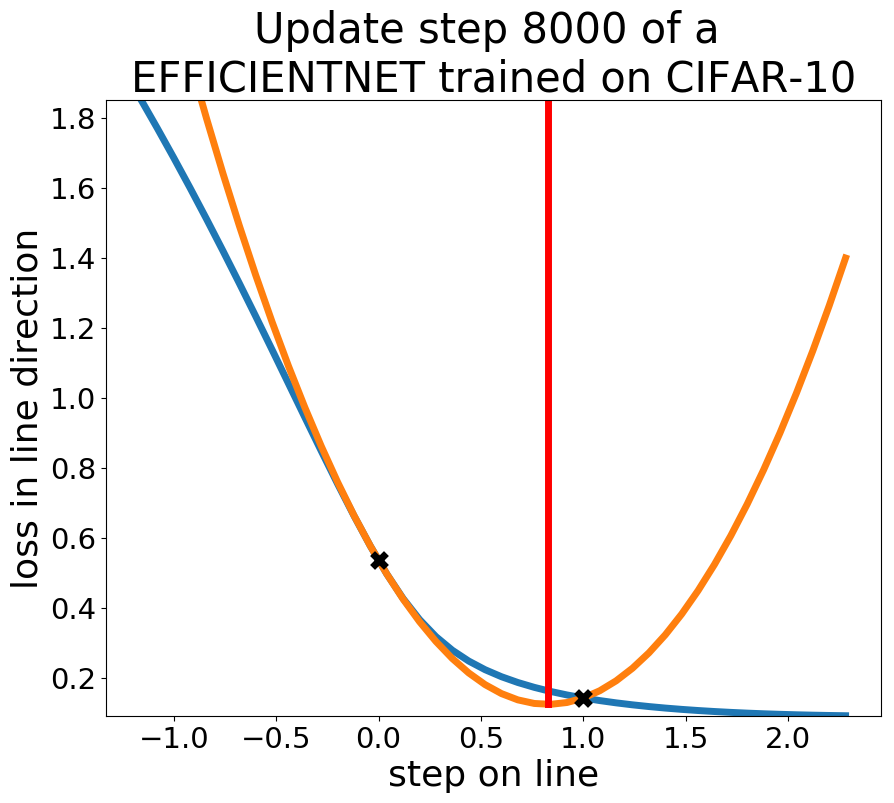}
	\includegraphics[width=0.19\linewidth]{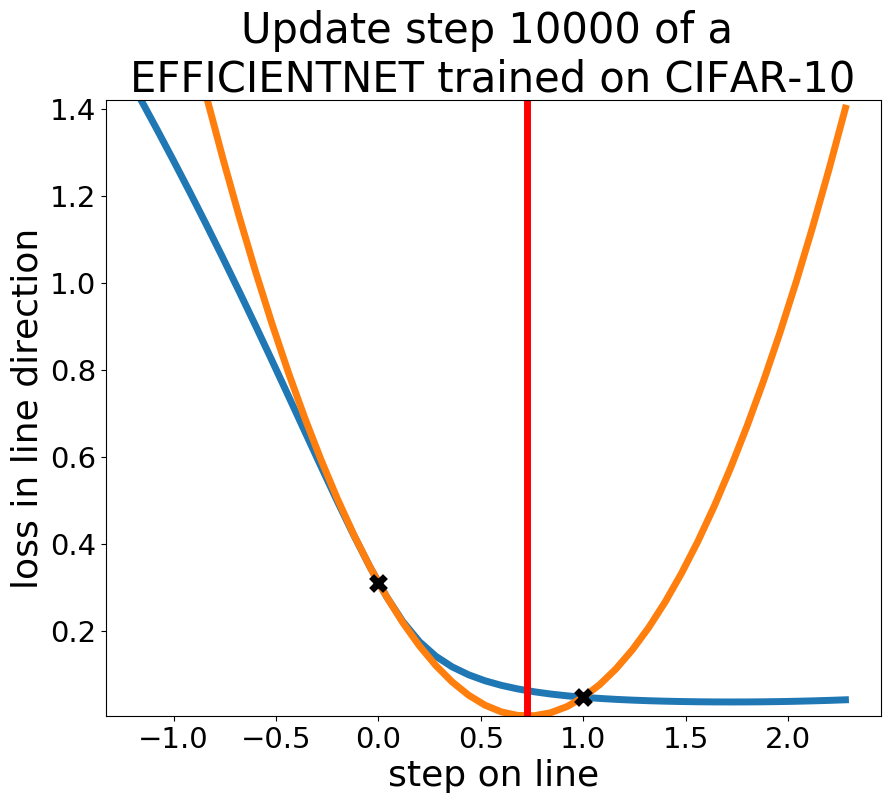}
	\includegraphics[width=0.19\linewidth]{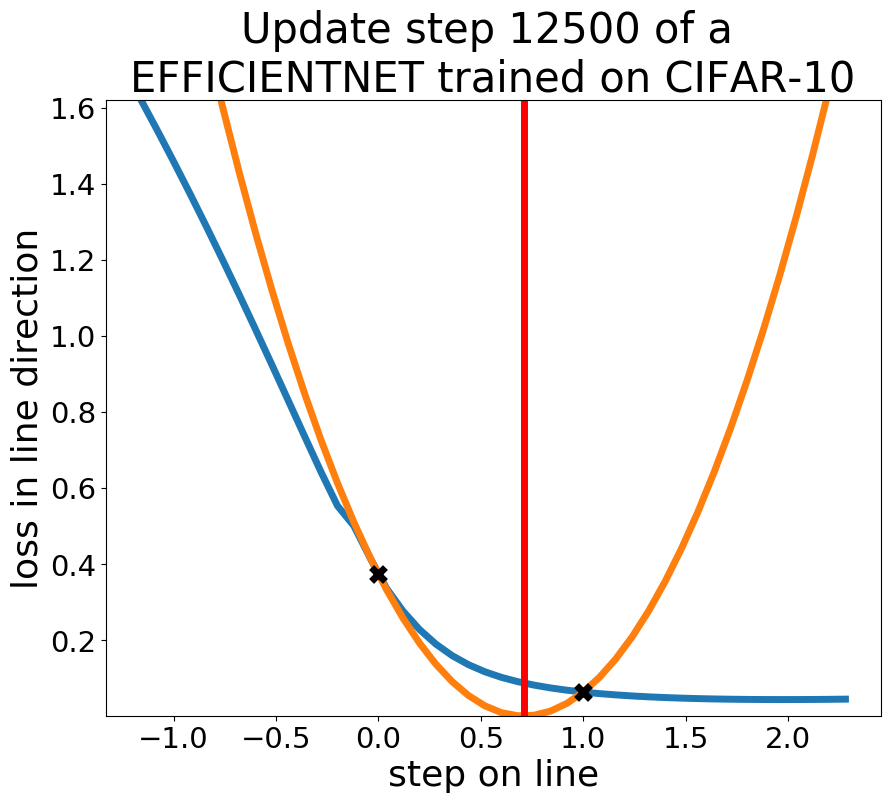}
	\includegraphics[width=0.19\linewidth]{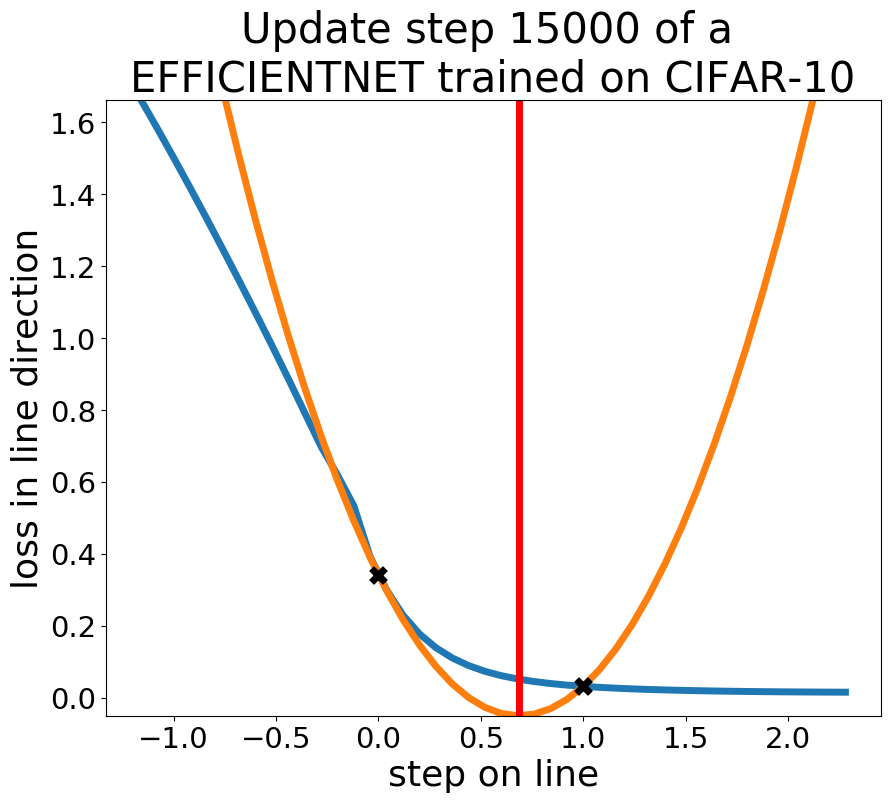}
	\caption[]{Mini-batch losses along lines of \textbf{EfficientNet}. For explanations see Figure \ref{fig_loss_lines_dense_net_40}. The parabolic approximation fits only on the left hand side and during the first 150 steps it does not fit at all, however, the minimum of the parabola is still a good estimator for a low loss value on the line.}
	\vspace{-0.5cm}
\end{figure*}
\vfil
\pagebreak
\section{PAL with all additions:}
\label{sec:PAL with all addition}
Algorithm \ref{alg:PAL} provides the full description of \pal\ including the additions described in Section \ref{subsec:features}.
After analyzing the best hyperparameter combinations of \pal\ over all experiments, we suggest to use values from the following parameter intervals: $\mu=[0.1,1]$, $\alpha=[1.0,1.6]$, $\beta=[0,0.4]$, $s_{\text{max}}=3$. Where  $\alpha,\beta $ and $s_{\text{max}}$ usually have a low sensitivity. Thus, the basic implementation of \pal\ (Algorithm \ref{alg:PAL_basic}) performs already well and is always found in the upper quartile considering the performance of all analyzed hyperparameter configurations in our experiments.
PyTorch and Tensorflow 1.5 implementations are provided at \url{https://github.com/cogsys-tuebingen/PAL}.
\begin{algorithm}[h]
	\caption{\pal, our proposed line search algorithm for DNNs. See Section \ref{sec:algorithm}\ for details.}  
	\begin{multicols}{2}
	\begin{algorithmic}[1]
		\label{alg:PAL}
		\renewcommand{\algorithmicrequire}{\textbf{Input:}}
		\renewcommand\algorithmiccomment[1]{
			\hfill \eqmakebox[a][l]{\small #1} %
		}
		\REQUIRE Hyperparameters: $\mu$: measuring step size, $\alpha$: update step adaptation, $\beta$: direction adaptation factor, $s_{\text{max}}$: maximum step size.
		\REQUIRE $\BL $: mini-batch loss function
		\REQUIRE $\vec{\theta}_0$: initial parameter vector	
		\STATE $t \leftarrow 0$
		\STATE $\mathbf{d}_t \leftarrow \vec{0}$ 
		\WHILE{$\mathbf{\theta}_t$ not converged} 
		\STATE $\mathbb{B}_t \leftarrow \text{sampleBatch}()$ 		
		\STATE $l_0 \leftarrow \BL[t](\mathbf{\theta}_{t})$ 
		\STATE $\mathbf{d}_t \leftarrow -\nabla_{\mathbf{\theta}_t}\BL[t](\mathbf{\theta}_t)+ \beta \mathbf{d}_{t-1} $ 
		\STATE $l_\mu \leftarrow \BL[t](\mathbf{\theta}_t+\mu \frac{\mathbf{d}_t}{||\mathbf{d}_t||})$ 
		\STATE $b \leftarrow  \nabla_{\mathbf{\theta}_t}\BL[t](\mathbf{\theta}_t) \frac{\mathbf{d}_t}{||\mathbf{d}_t||}$ 
		\STATE $a \leftarrow \frac{l_{\mu}-l_0-b\mu}{\mu^2}$ 
		\IF{$a>0$ \AND  $b<0$}
		\STATE $\s \leftarrow -\alpha\frac{b}{2a}$ 
		\ELSIF{$a\leq 0$ \AND  $b<0$}
		\STATE $\s \leftarrow \mu$
		\ELSE
		\STATE $\s \leftarrow0$ 
		\ENDIF
		\IF{$\s > s_{\text{max}}$}
		\STATE $\s \leftarrow s_{\text{max}}$
		\ENDIF
		
		\STATE	$\mathbf{\theta}_{t+1} \leftarrow \mathbf{\theta}_{t}+ \s \frac{\mathbf{d}_t}{||\mathbf{d}_t||}$ 
		\STATE $t \leftarrow t+1$
		\ENDWHILE
		\RETURN $\mathbf{\theta}_t$
	\end{algorithmic} 
	\end{multicols}
\end{algorithm}

\section{Further Theoretical Considerations}
\label{app:theoretical_considerations}
We have to note that the following derivation is based upon \textbf{strong assumptions} and if they are valid at all, they are likely more valid locally than globally.
We assume that each slice of the loss function is a one-dimensional parabolic function: 
\begin{assumption}
	\label{ass:convergence}
	Let $n \in \mathbb{N}$ be the number of parameters and let $\mathbf{l},\mathbf{d} \in \mathbb{R}^n$ be vectors. Then for all $\mathbf{l},\mathbf{d}$ there exists $a,b,c \in \mathbb{R}$ with $a>0$, such that $\BL[](\mathbf{l}+\mathbf{d}s)= as^2+bs+c$ for all   $s \in \mathbb{R}.$\ 
\end{assumption}
This \textbf{strong} assumption is a simplified adaptation to our empirical results that lines in negative gradient direction behave locally almost parabolic (see Section \ref{sec:introduction}). 
For the following derivations we assume a basic \pal\ without the additions introduced in Section \ref{subsec:features}. Proofs are provided in Appendix \ref{sec:Proofs}.
At first we show that $\BL[](\mathbf{\theta})$ is a n-dimensional parabolic function:
\newtheorem{lemma}{Lemma}
\newtheorem{proposition}{Proposition}\vspace{-0.07cm}
\begin{lemma}
	Let $f: \mathbb{R}^n \rightarrow \mathbb{R}$ be a k-times continuously differentiable function. Furthermore, assume there exists $a,b,c \in \mathbb{R}$ with $a>0$, such that $f(\mathbf{l}+\mathbf{d}s)= as^2+bs+c$ for all $s \in \mathbb{R}$. Then there exist $z \in \mathbb{R}, \mathbf{r} \in \mathbb{R}^n$ and a positive definite Matrix $\mathbf{Q}\in \mathbb{R}^{n\times n}$ such that $f(\mathbf{x})= c+\mathbf{r}^T\mathbf{x}+\mathbf{x}^T\mathbf{Q}\mathbf{x}$ for all $\mathbf{x} \in \mathbb{R}^n$.
\end{lemma}
\noindent Now we show that \pal\ converges on $\BL[](\mathbf{\theta})$:
\begin{proposition}
	PAL converges on  $f : \mathbb{R}^n \rightarrow \mathbb{R}, \mathbf{x} \mapsto c+\mathbf{r}^T\mathbf{x}+\mathbf{x}^T\mathbf{Q}\mathbf{x}$ with $\mathbf{Q}\in \mathbb{R}^{n\times n}$ hermitian and positive definite.
\end{proposition}
We have to note that in this scenario, PAL is identical to the method of steepest descent for which the convergence, including convergence rates on quadratics, is already proven in \cite[Page 235]{nonlinear_programming}. Nevertheless, we have attached our proof better adapted to this scenario.

For a noisy scenario where each batch defines a quadratic, \pal\ has no convergence guarantee. Given two shifted one-dimensional parabolas, $ax^2+bx+c$ and $a(x+d)^2+b(x+d)+c$, which are presented to \pal\ alternately, \pal\ will always perform an update step to the minimum position of one of these but never to the minimum position of the average of both. By slightly changing the training procedure and assuming that each $\BL[](\mathbf{\theta})$ has the same $\mathbf{Q}$ this can be fixed:
\begin{proposition}
	If $ \mathcal{L}(\mathbf{\theta}) : \mathbb{R}^n \rightarrow \mathbb{R}$ $\mathbf{\theta} \mapsto  \mathcal{L}(\mathbf{\theta})=\frac{1}{m}\sum_{i=1}^m c_i+\mathbf{r}_i^T\mathbf{\theta}+\mathbf{\theta}^T\mathbf{Q}_i\mathbf{\theta}$ and $ c_i+\mathbf{r}_i^T\mathbf{\theta}+\mathbf{\theta}^T\mathbf{Q}_i\mathbf{\theta}= \BL[i](\mathbf{\theta})$ with $m$ being number the of batches $\mathbb{B}_i$. (Each batch defines a parabola. The empirical loss $\mathcal{L}(\mathbf{\theta})$ is the mean of these parabolas). And for all  $i,j \in \mathbb{N}$ it holds that $ \mathbf{Q}_i=\mathbf{Q}_j$ and that $\mathbf{Q_i}$ is positive definite. Then $\argmin{\theta} \mathcal{L}(\mathbf{\theta}) =\frac{1}{m}\sum_{i=1}^m\argmin{\theta} \BL[i](\mathbf{\theta})$ holds.
\end{proposition}
\noindent This implies that under Assumption \ref{ass:convergence} and a fixed $\mathbf{Q}$ the position of the minimum of the empirical loss is given by the average of the positions of the mini-batch losses' minima. The minimum position of full-batch loss is found by \pal, by slightly adapting \pal\, to search on one batch until it finds the minimum's position and then averaging the minimum of each batch. As a result, \pal\ converges in this noisy scenario. However, we have to emphasize at this point that our assumptions about $\mathbf{l}$ and $\mathbf{Q}$  are likely not valid for general deep learning scenarios. Nevertheless, if it is locally valid, this direction might be a further explanation, in addition to those of \cite{assymetric_valley,largesclelandscape}, why   averaging the weights of each update step in the last epoch (stochastic weight averaging) \cite{swa} performs well.

\subsection{Proofs}
\newtheorem{lemma_ap}{Lemma}
\newtheorem{proposition_ap}{Proposition}
\label{sec:Proofs}
\begin{lemma_ap}
	Let $f: \mathbb{R}^n \rightarrow \mathbb{R}$ be a k-times continuously differentiable function. Furthermore, assume there exists $a,b,c \in \mathbb{R}$ with $a>0$, such that $f(\mathbf{l}+\mathbf{d}s)= as^2+bs+c$ for all $s \in \mathbb{R}$. Then there exist $z \in \mathbb{R}, \mathbf{r} \in \mathbb{R}^n$ and a positive definite Matrix $\mathbf{Q}\in \mathbb{R}^{n\times n}$ such that $f(\mathbf{x})= c+\mathbf{r}^T\mathbf{x}+\mathbf{x}^T\mathbf{Q}\mathbf{x}$ for all $\mathbf{x} \in \mathbb{R}^n$.
\end{lemma_ap}
\begin{proof}
\begin{equation}
\begin{aligned}
 g(\mathbf{x})= u+\mathbf{v}^T\mathbf{x}+\mathbf{x}^T\mathbf{W}\mathbf{x} \text{ for some } u\in \mathbb{R}, \mathbf{v}\in \mathbb{R}^n \text{ and } \mathbf{W} \in \mathbb{R}^{n\times n} \\ \Leftrightarrow \forall \mathbf{l},\mathbf{d} \in \mathbb{R}^n \land ||\mathbf{d}||=1 : \sum\limits_{j=1}^{n}\sum\limits_{k=1}^{n}\sum\limits_{l=1}^{n} \frac{\partial^3 g(\mathbf{l}) }{\partial x_j,\partial x_k,\partial x_l}d_{j}d_{k}d_{l}=0
 \end{aligned}
\end{equation}
$\Rightarrow$ holds since we have a polynomial of degree 2 and its third derivative is always a $\mathbf{0}$ tensor.\\
$\Leftarrow$ holds since the reminder of the quadratic Taylor expansion is always 0.

\noindent
In our case the right part is 0 since:
	\begin{equation}
\sum\limits_{j=1}^{n}\sum\limits_{k=1}^{n}\sum\limits_{l=1}^{n} \frac{\partial^3 f(\mathbf{l}) }{\partial x_j,\partial x_k,\partial x_l}d_{j}d_{k}d_{l}= \frac{\partial}{\partial s^3}f(\mathbf{l}+\mathbf{d}s)=0
	\end{equation}
\\	
 In words: $f(\mathbf{x})$ is a parabolic function if and only if for each location $\mathbf{l}$ the third directional derivative of $f(\mathbf{\mathbf{l}})$ in each direction $\mathbf{d}$ is $0$. Which is the case, since the third derivative of each intersection is 0.
\\
$\mathbf{W}$ is positive definite since: \\
\begin{equation}
\forall \mathbf{d},\mathbf{l} \in \mathbb{R}^n \land ||\mathbf{d}||=1: \mathbf{d}^T \mathbf{W} \mathbf{d} =\frac{1}{2}\mathbf{d}^T \mathbf{H}(\mathbf{l}) \mathbf{d}= \frac{1}{2} \frac{\partial}{\partial s^2}f(\mathbf{l}+\mathbf{d}s) = a > 0
\end{equation}
where $\mathbf{H}$ is the Hessian.
\end{proof}
\vspace{0.5cm}\pagebreak

\begin{proposition_ap}
	PAL converges on  $f : \mathbb{R}^n \rightarrow \mathbb{R}, \mathbf{x} \mapsto c+\mathbf{r}^T\mathbf{x}+\mathbf{x}^T\mathbf{Q}\mathbf{x}$ with $\mathbf{Q}\in \mathbb{R}^{n\times n}$ hermitian and positive definite.
\end{proposition_ap}
\begin{proof}
	\quad
\\For this prove we consider a basic \pal\ without the features introduced in Section \ref{subsec:features}.
Note, that during the proof we will see, that $a>0$ and $b<0$. Thus, only the update step for this case has to be considered (see Section \ref{subsec:casediscrimination}.
	
 \quad \\ 
 $f(\mathbf{x})$ is convex since $\mathbf{Q}$ is positive definite. Thus it has one minimum.\\
 Without loss of generality we set $c=0,\mathbf{r}=\mathbf{0},\mathbf{x}_n \neq 0$ 
 \begin{equation}
 f(\mathbf{x})=\mathbf{x}^T\mathbf{Q}\mathbf{x} \text{ and }\nabla_{\mathbf{x}} f(\mathbf{x})= f'(\mathbf{x})=2\mathbf{Q}\mathbf{x}
   \end{equation}
The values of $f(x)$ along a line through $\mathbf{x}$ in the direction of $-f'(\mathbf{x})$ are given by:
 \begin{equation}
 f(-f'(\mathbf{x})\hat{s}+\mathbf{x})
   \end{equation}   
Now we expand the line function:
\begin{equation}
\begin{aligned}
f(-f'(\mathbf{x})\hat{s}+\mathbf{x})&=
f(-2\mathbf{Q}\mathbf{x}\hat{s}+\mathbf{x})\\&=(-2\mathbf{Q}\mathbf{x}\hat{s}+\mathbf{x})^T\mathbf{Q}(-2\mathbf{Q}\mathbf{x}\hat{s}+\mathbf{x})
\\&=\underbrace{4\mathbf{x}^T\mathbf{Q}^3\mathbf{x}}_{=:a}\hat{s}^2+\underbrace{-4\mathbf{x}^T\mathbf{Q}^2\mathbf{x}}_{=:b}\hat{s}+\underbrace{\mathbf{x}^T\mathbf{Q}\mathbf{x}}_{=:c}
\end{aligned}
\end{equation}
 \\ Here we see that $f(\hat{s})$ is indeed a parabolic function with $a>0$, $b<0$ and $c>0$ since $\mathbf{Q}^3$, $\mathbf{Q}^2$ and $\mathbf{Q}$ are positive definite.\\
The location of the minimum $s_{min}$ of $f(\hat{s})$ is given by:
\begin{equation}
	 \hat{s}_{min}=\argmin{\hat{s}} f(-f'(\mathbf{x})\hat{s}+\mathbf{x})=-\frac{b}{2a}
\end{equation}
 \pal\ determines $\hat{s}_{min}$ exactly with $\hat{s}_{min}=\frac{s_{upd}}{||f'(x)||}$ (see equation 1 and 2). $||f'(x)||>0$  since otherwise we are already in the minimum.\\

\noindent The value at the minimum is given by: 
\begin{equation}
f(\hat{s}_{min})\\= a(\frac{-b}{2a})^2+b(\frac{-b}{2a})+c\\=-\frac{b^2}{4a}+c\\
=-\underbrace{\frac{(-\mathbf{x}^T\mathbf{Q}^2\mathbf{x})^2}{\mathbf{x}^T\mathbf{Q}^3\mathbf{x}}}_{=:g(\mathbf{x})}+\mathbf{x}^T\mathbf{Q}\mathbf{x}
\\=-g(\mathbf{x})+f(\mathbf{x}) \\
\end{equation}
Since $\mathbf{Q}^2$ and $\mathbf{Q}^3$ are positive definite and $\mathbf{x}\neq 0$: \begin{equation}g(\mathbf{x})>0\end{equation}
Now we consider the sequence $f(\mathbf{x}_n)$, with $\mathbf{x}_n$ defined by \textit{PAL} (see Equation 1): \\
\begin{equation}
\mathbf{x}_{n+1}=-\frac{f'(\mathbf{x}_n)}{||f'(x_n)||}\hat{s}_{upd}+\mathbf{x}_n= -f'(\mathbf{x}_n)\hat{s}_{min}+\mathbf{x}_n
\end{equation}
 It is easily seen by induction that:
 \begin{equation}
  0<f(\mathbf{x}_{n+1})<f(\mathbf{x}_{n})=\sum\limits_{i=0}^{n-1} -g(\mathbf{x}_i) +f(\mathbf{x}_0)< f(\mathbf{x}_0). 
\end{equation}
  $g(\mathbf{x}_{n})$ converges to 0. Since $\forall n:g(\mathbf{x_n})>0$ and $\sum\limits_{i=0}^{n-1} -g(\mathbf{x}_i)$ is bounded. \\
  \\ Now we have to show that $\mathbf{x}_n$ converges to 0.
 \\ We have:
  \begin{equation}
  g(\mathbf{x}_n)=\frac{(\mathbf{x}_n^T\mathbf{Q}^2\mathbf{x}_n)^2}{\mathbf{x}_n^T\mathbf{Q}^3\mathbf{x}_n}=\frac{\langle\mathbf{x}_n,\mathbf{Q}^2\mathbf{x}_n\rangle^2}{\langle\mathbf{x}_n,\mathbf{Q}^3\mathbf{x}_n\rangle}
  \end{equation}
 Now we use the theorem of Courant-Fischer:
   \begin{equation}
   \begin{aligned}
   \langle x,x\rangle\min\{\lambda_1,\dots,\lambda_n\}\leq \langle x,Ax \rangle\leq \langle x,x \rangle \max\{\lambda_1,\dots,\lambda_n\}\\ \text{ for any symmetric }  A \in \mathbb{R}^{n\times n} \text{ with } \lambda_1,\dots,\lambda_n
      \end{aligned}
   \end{equation}
    And get: 
\begin{equation}
g(\mathbf{x}_n) \geq \frac{\lambda^2_{\mathbf{Q}^2\min} \langle \mathbf{x}_n,\mathbf{x}_n \rangle^2}{\lambda_{\mathbf{Q}^3\max} \langle \mathbf{x}_n,\mathbf{x}_n \rangle}=C \frac{||\mathbf{x}_n||^4}{||\mathbf{x}_n||^2}=C ||\mathbf{x}_n||^2
\end{equation} 
with \begin{equation}C=\frac{\lambda^2_{\mathbf{Q}^2\min}}{\lambda_{\mathbf{Q}^3\max}}>0 \text{ since all } \lambda \text{ of the positive definite } \mathbf{Q} \text{ are positive} \end{equation}
 Thus, we have:
\begin{equation}
 g(\mathbf{x}_n) \geq C ||\mathbf{x}_n||^2 \geq 0 \end{equation} 
 Since $g(\mathbf{x}_n)$ converges to 0,  $C ||\mathbf{x}_n||^2 $ converges to 0. \\This means, $\mathbf{x}_n$ converges to $\mathbf{0} $, which is the location of the minimum. 
\end{proof}
\singlespacing
\normalsize


\vspace{0.5cm}
\begin{proposition}
	If $ \mathcal{L}(\mathbf{\theta}) : \mathbb{R}^n \rightarrow \mathbb{R}$ $\mathbf{\theta} \mapsto  \mathcal{L}(\mathbf{\theta})=\frac{1}{m}\sum_{i=1}^m c_i+\mathbf{r}_i^T\mathbf{\theta}+\mathbf{\theta}^T\mathbf{Q}_i\mathbf{\theta}$ and $ c_i+\mathbf{r}_i^T\mathbf{\theta}+\mathbf{\theta}^T\mathbf{Q}_i\mathbf{\theta}= \BL[i](\mathbf{\theta})$ with $m$ being number the of batches $\mathbb{B}_i$. (Each batch defines a parabola. The empirical loss $\mathcal{L}(\mathbf{\theta})$ is the mean of these parabolas). And for all  $i,j \in \mathbb{N}$ it holds that $ \mathbf{Q}_i=\mathbf{Q}_j$ and that $\mathbf{Q_i}$ is positive definite. Then $\argmin{\theta} \mathcal{L}(\mathbf{\theta}) =\frac{1}{m}\sum_{i=1}^m\argmin{\theta} \BL[i](\mathbf{\theta})$ holds.
\end{proposition}

\begin{proof}
 \quad\\ 
 Since $\mathcal{L}(\mathbf{\theta})$ is a sum of convex functions, it is also convex and has one minimum.\\
 At first we determine the derivative of $\mathcal{L}(\mathbf{\theta})$ with respect to $\mathbf{\theta}$:
  \begin{equation}
 \frac{\partial}{\partial \mathbf{\theta}}\mathcal{L}(\mathbf{\theta})= \frac{1}{m}\sum\limits_{i=1}^m(\mathbf{r}_i+2\mathbf{Q}_i\mathbf{\theta})= 2\mathbf{Q}\mathbf{\theta}+\frac{1}{m}\sum\limits_{i=1}^m\mathbf{r}_i
 \end{equation}
Then we determine the minima:
   \begin{equation}
  \argmin{\mathbf{\theta}} \mathcal{L}(\mathbf{\theta}) \Leftrightarrow
 \frac{\partial}{\partial \mathbf{\theta}}\mathcal{L}(\mathbf{\theta}){=}\mathbf{0} \Leftrightarrow \mathbf{\theta}=-\frac{1}{2}( \sum\limits_{i=1}^m \mathbf{Q}_i)^{-1} \sum\limits_{i=1}^m \mathbf{r_i}= -\frac{1}{2m} \mathbf{Q}^{-1} \sum\limits_{i=1}^m \mathbf{r_i}
   \end{equation}
   \begin{equation}
\argmin{\mathbf{\theta}} \BL[i](\mathbf{\theta})=-\frac{1}{2}\mathbf{Q}^{-1}\mathbf{r}_i
 \end{equation}
 Thus, we get:
 \begin{equation}
  \argmin{\mathbf{\theta}} \mathcal{L}(\mathbf{\theta})= -\frac{1}{2m}\mathbf{Q}^{-1}\sum\limits_{i=1}^m \mathbf{r}_i=\frac{1}{m}\sum\limits_{i=1}^m -\frac{1}{2}\mathbf{Q}^{-1}\mathbf{r}_i =\frac{1}{m}\sum\limits_{i=1}^m\argmin{\mathbf{\theta}} \BL[i](\mathbf{\theta})
  \end{equation}
\end{proof}
\vfill
\vspace{4cm}
\pagebreak
\section{Further experimental results}
\label{sec:further_results}
\subsection{Performance Comparison on ImageNet, CIFAR-10, CIFAR-100 and Tolstoi}
\label{subsec:performance_comparison}
\hspace{1cm}
\begin{figure*}[h!]
	\def \figwidth {0.32}
	\def \figheight {0.25}
	\def \figwidthb {0.25}
	\centering
	\vspace{-0.5cm}
	
	\includegraphics[width=\figwidth\linewidth,height=\figheight\linewidth]{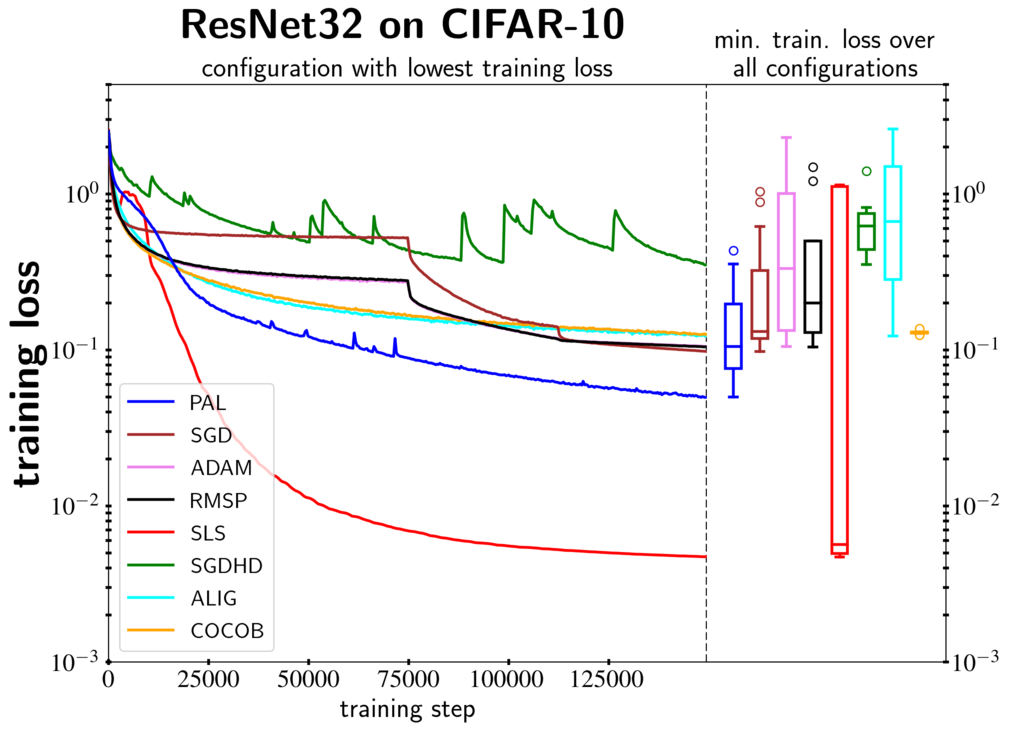}	
	\includegraphics[width=\figwidth\linewidth,height=\figheight\linewidth]{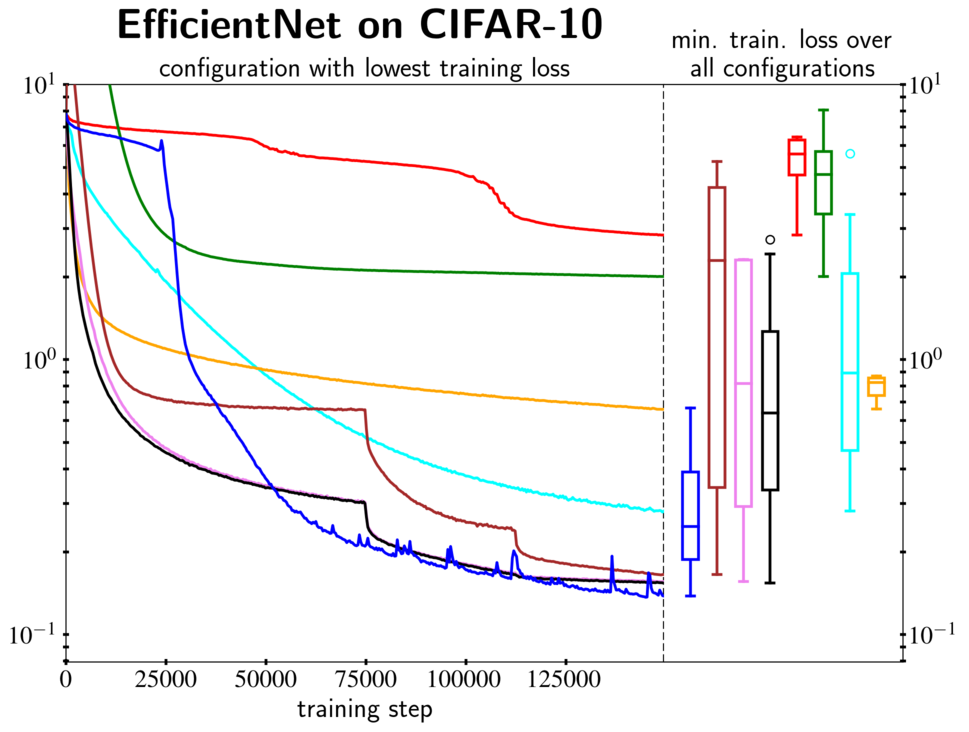}
	\includegraphics[width=\figwidth\linewidth,height=\figheight\linewidth]{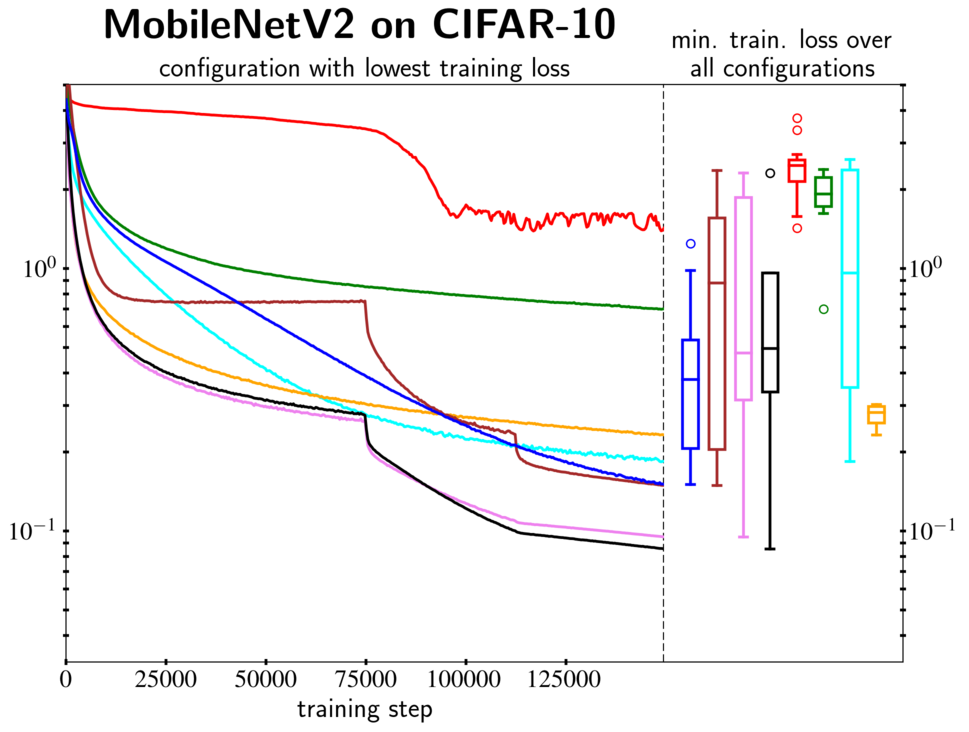}

	\includegraphics[width=\figwidth\linewidth,height=\figheight\linewidth]{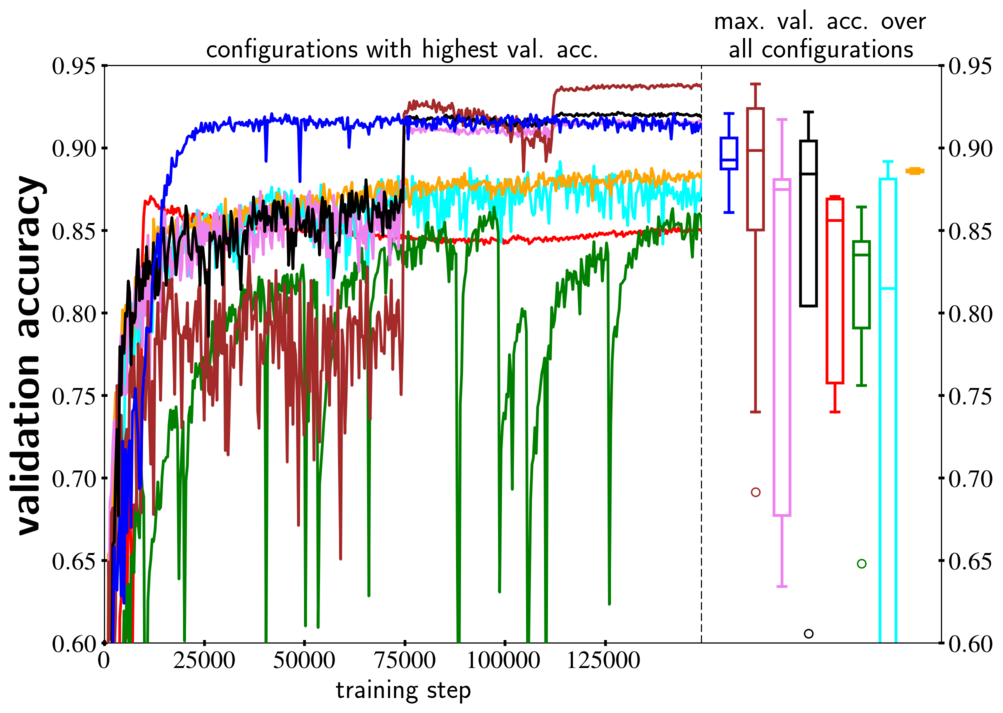}
	\includegraphics[width=\figwidth\linewidth,height=\figheight\linewidth]{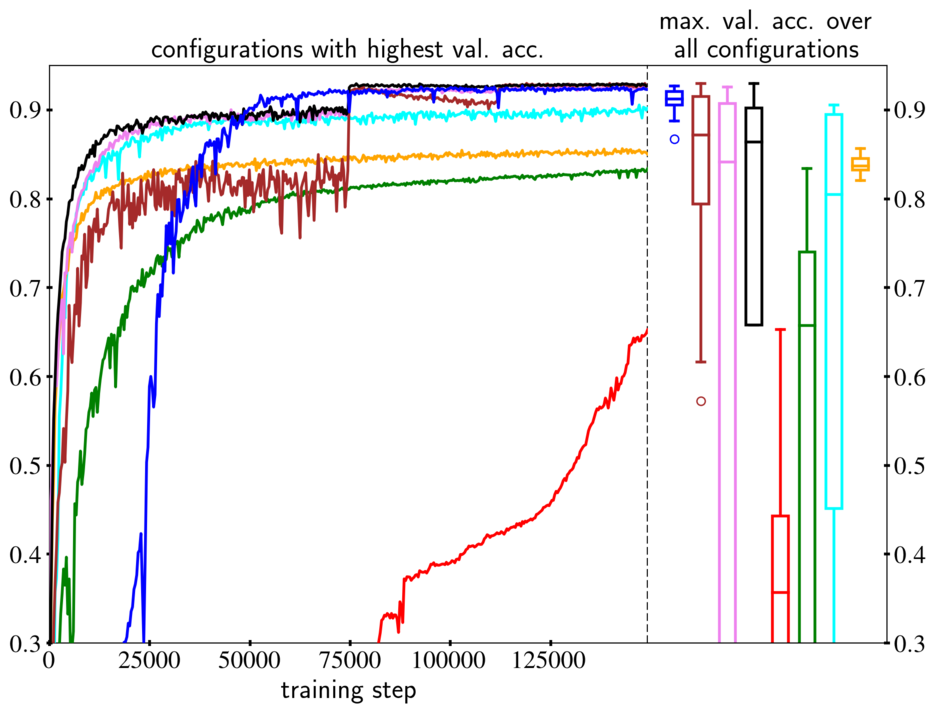}
	\includegraphics[width=\figwidth\linewidth,height=\figheight\linewidth]{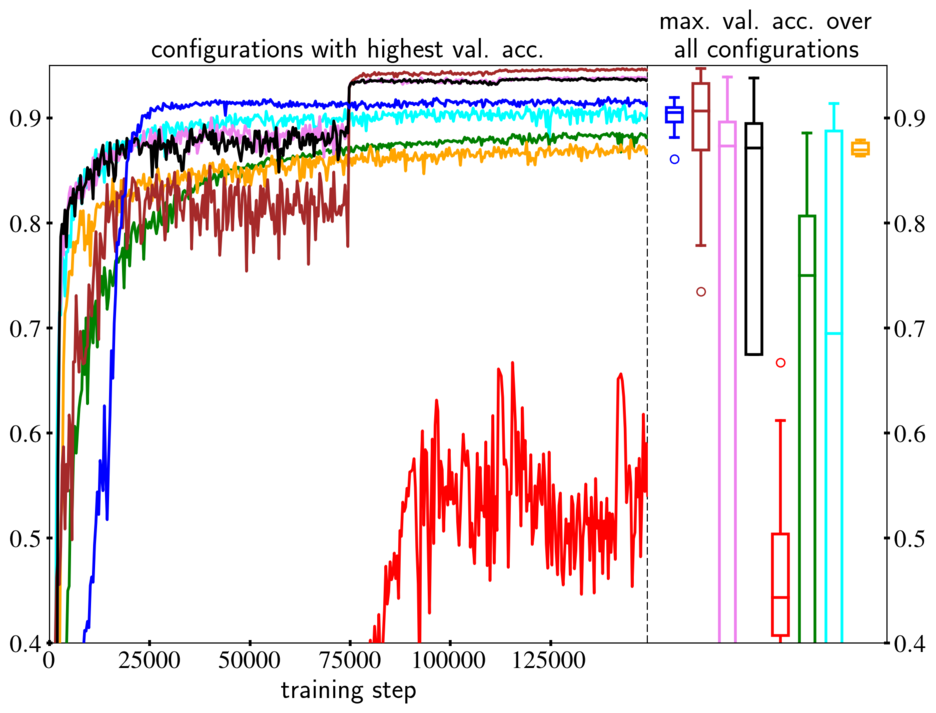}
	
	\includegraphics[width=\figwidth\linewidth,height=\figheight\linewidth]{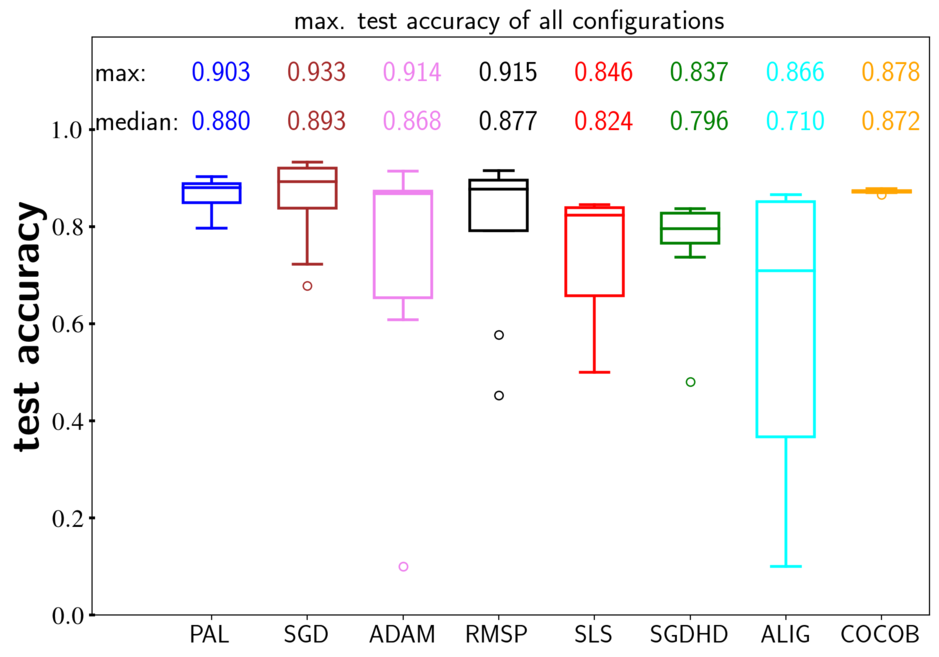}
	\includegraphics[width=\figwidth\linewidth,height=\figheight\linewidth]{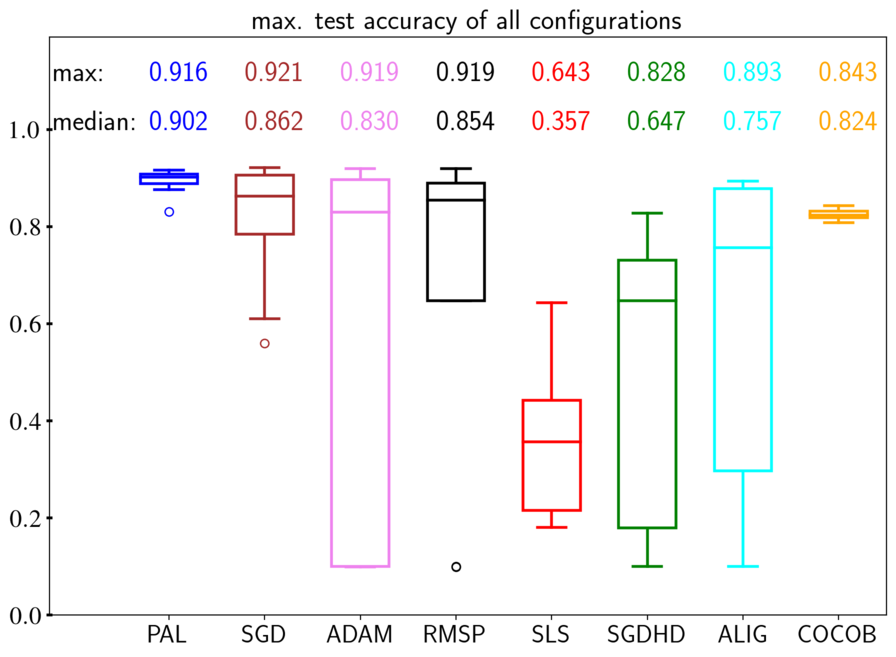}
	\includegraphics[width=\figwidth\linewidth,height=\figheight\linewidth]{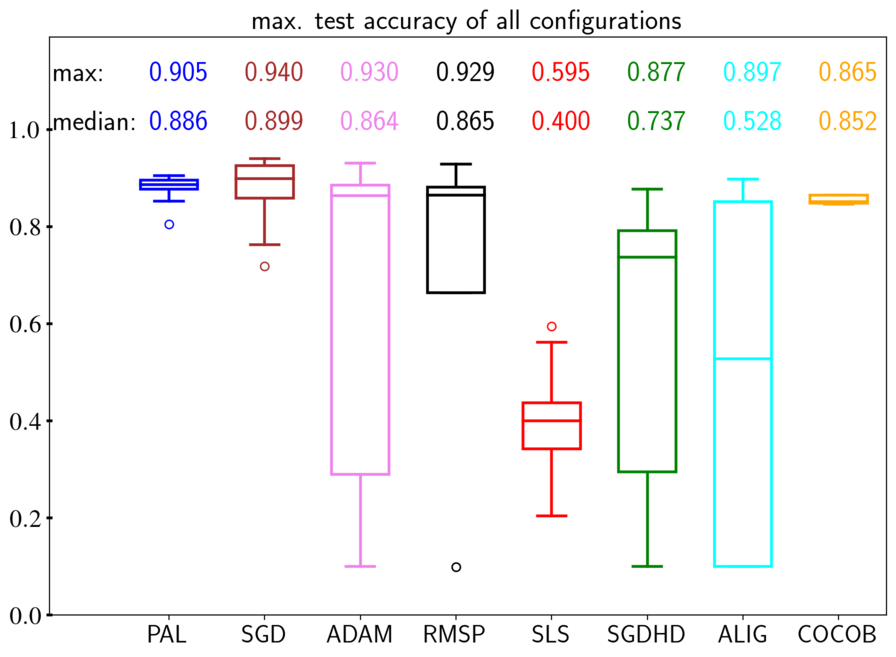}

	\includegraphics[width=\figwidth\linewidth,height=\figheight\linewidth]{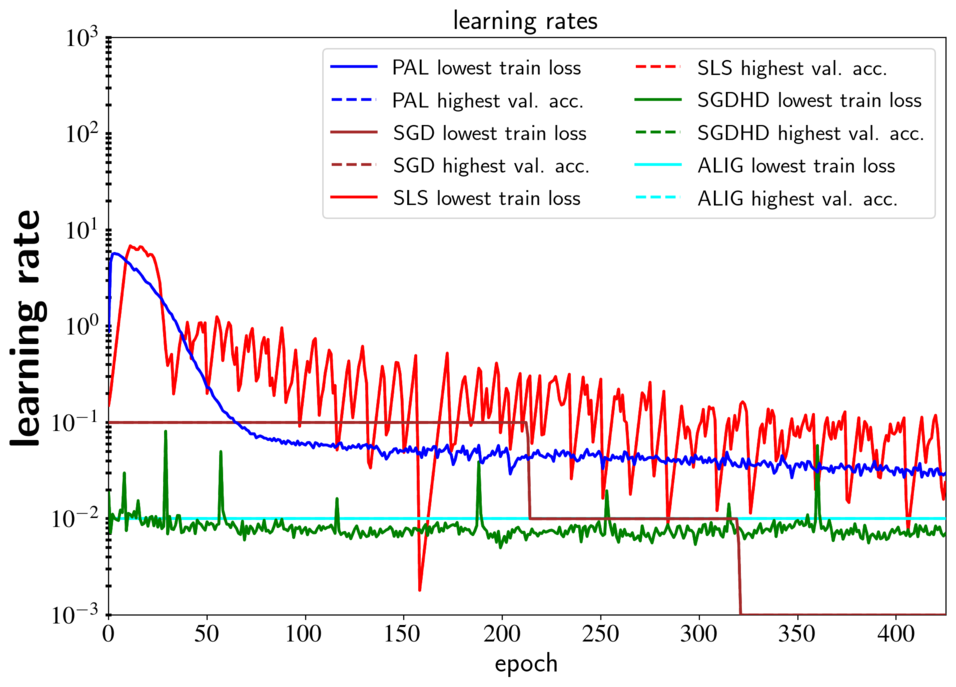}
	\includegraphics[width=\figwidth\linewidth,height=\figheight\linewidth]{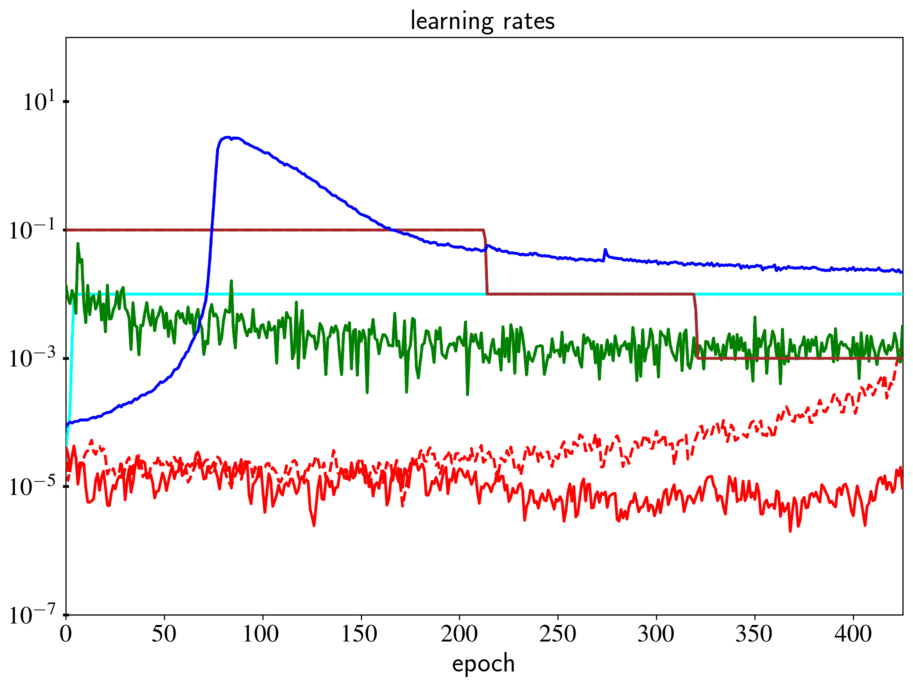}
	\includegraphics[width=\figwidth\linewidth,height=\figheight\linewidth]{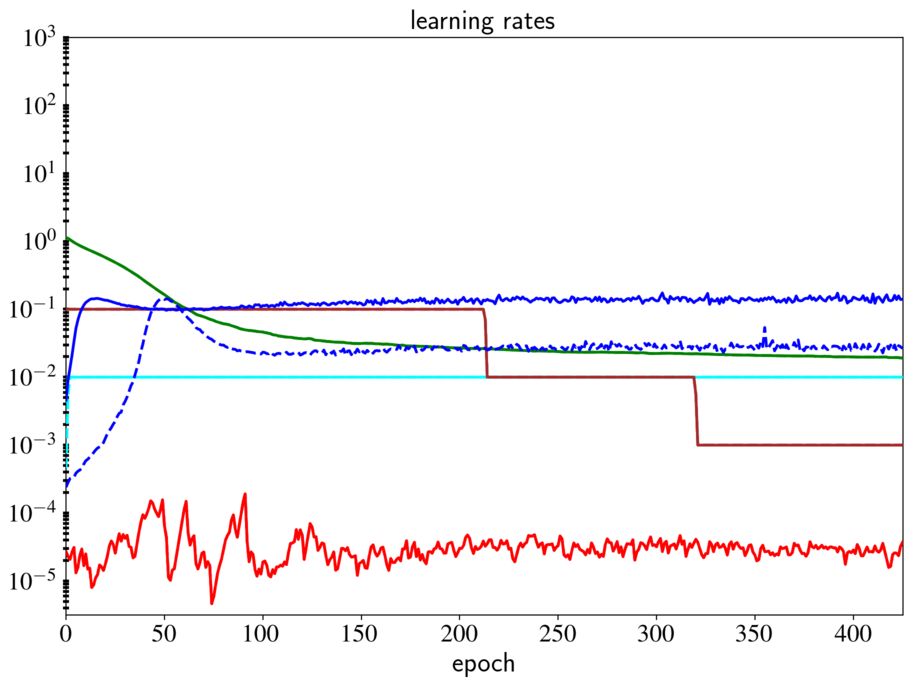}
	\caption[]{Comparison  of \pal\ on \textbf{CIFAR-10} against  \textit{SLS}, \textit{SGD}, \textit{ADAM}, \textit{RMSProp}, \textit{ALIG}, \textit{SGDHD} and \textit{COCOB} on train. loss (row 1), val. acc. (row 2), test. acc. (row 3) and \textit{SLS}, \textit{SGD}, \textit{ALIG}, \textit{SGDHD} and \pal\ on  learning rates (row 4). Results are averaged over 3 runs. Box plots result from comprehensive hyperparameter grid searches in plausible intervals. Learning rates are averaged over epochs.  \pal\ surpasses \textit{SLS}, \textit{ALIG}, \textit{SGDHD} and competes against all other optimizers except against SGD. The learning rate schedule comparison shows that \pal\ performs competitive although elaborating significantly different schedules.}
	\label{fig:app_cifar10}
\end{figure*}

\begin{figure*}[h!]
	\def \figwidth {0.31}
	\def \figheight {0.25}
	\def \figwidthb {0.25}
	\centering
	\vspace{-0.5cm}
	
	\includegraphics[width=\figwidth\linewidth,height=\figheight\linewidth]{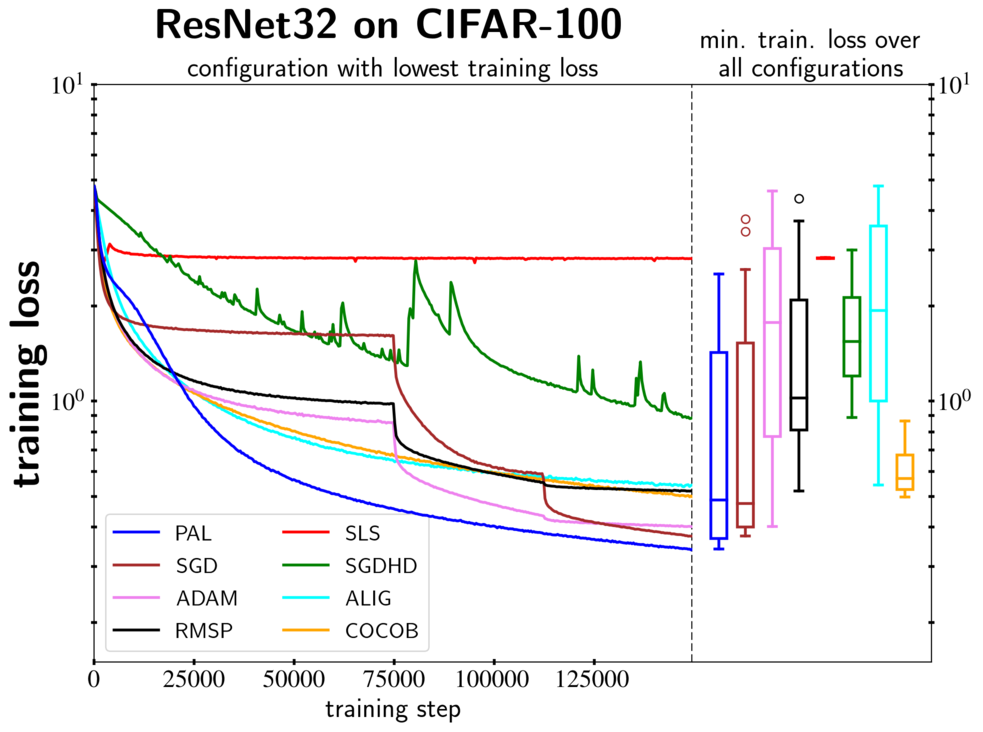}	
	\includegraphics[width=\figwidth\linewidth,height=\figheight\linewidth]{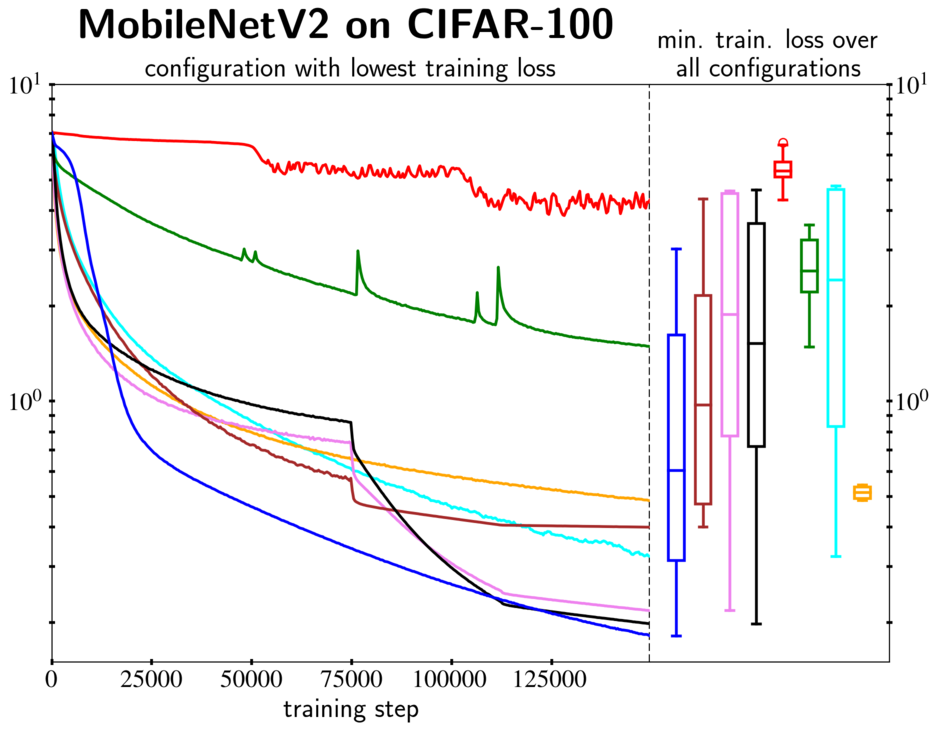}\\

	\includegraphics[width=\figwidth\linewidth,height=\figheight\linewidth]{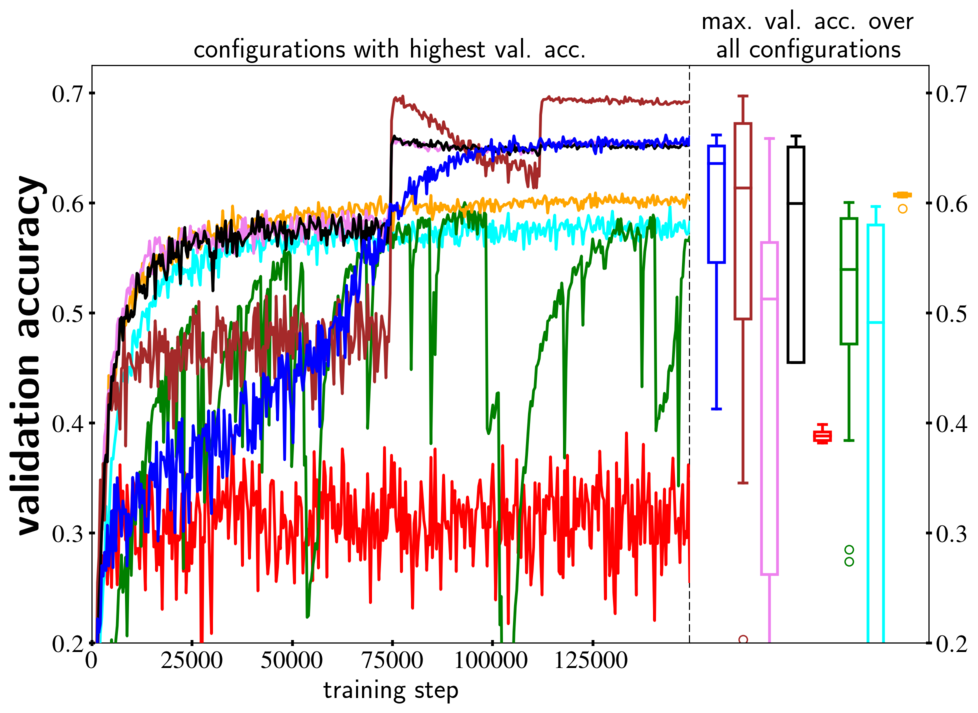}
	\includegraphics[width=\figwidth\linewidth,height=\figheight\linewidth]{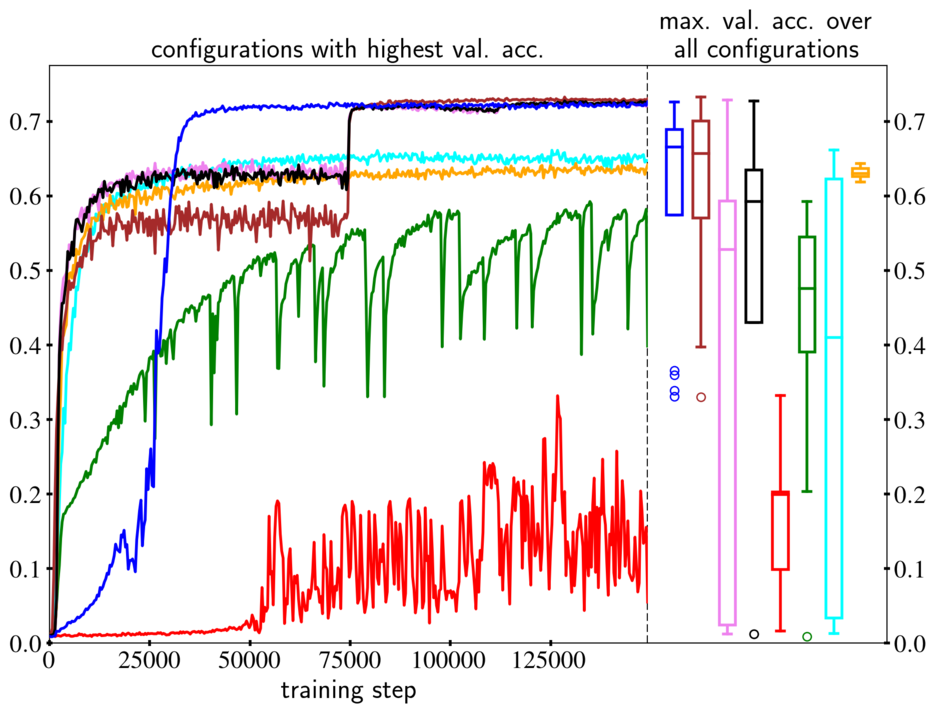}\\
	
	\includegraphics[width=\figwidth\linewidth,height=\figheight\linewidth]{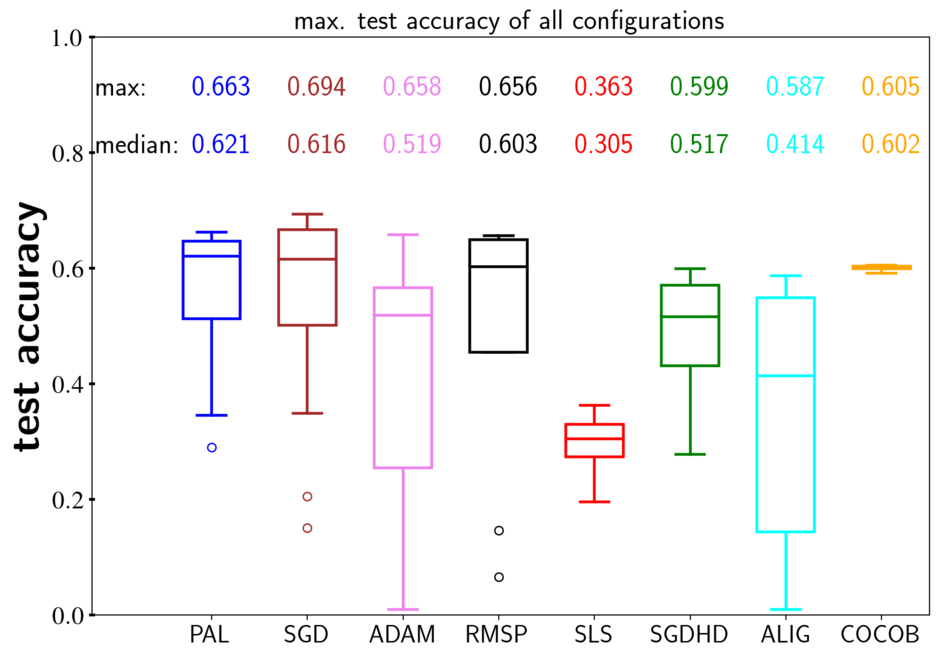}
	\includegraphics[width=\figwidth\linewidth,height=\figheight\linewidth]{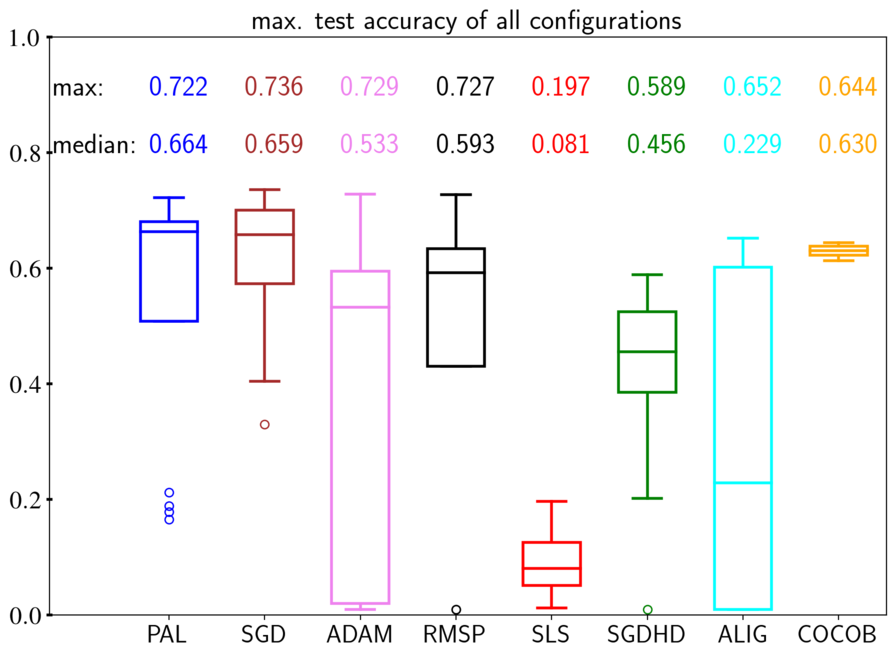}\\

	\includegraphics[width=\figwidth\linewidth,height=\figheight\linewidth]{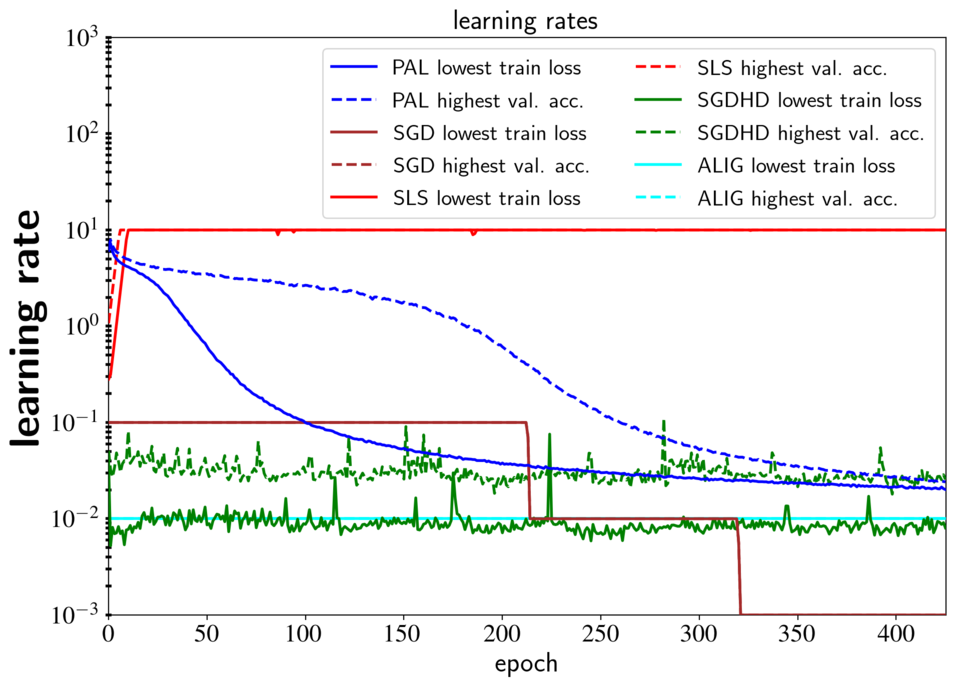}
	\includegraphics[width=\figwidth\linewidth,height=\figheight\linewidth]{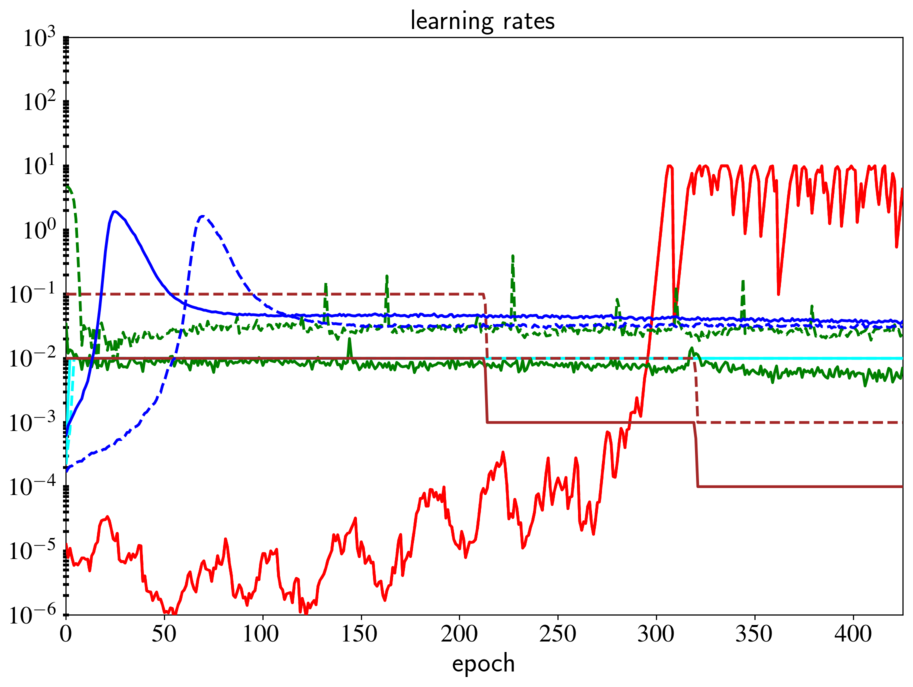}

	\caption[]{Comparison of \pal\ on \textbf{CIFAR-100}  against  \textit{SLS}, \textit{SGD}, \textit{ADAM}, \textit{RMSProp}, \textit{ALIG}, \textit{SGDHD} and \textit{COCOB} on train. loss (row 1), val. acc. (row 2), test. acc. (row 3) and \textit{SLS}, \textit{SGD}, \textit{ALIG}, \textit{SGDHD} and \pal\ on  learning rates (row 4)). Results are averaged over 3 runs. Box plots result from comprehensive hyperparameter grid searches in plausible intervals. Learning rates are averaged over epochs. \pal\ surpasses \textit{SLS}, \textit{ALIG}, \textit{SGDHD} and competes against all other optimizers except against SGD. The learning rate schedule comparison shows that \pal\ performs competitive although elaborating significantly different schedules.}
	\label{fig.cifar100_step_plots}
\end{figure*}

\begin{figure*}[h!]
	\def \figwidth {0.31}
	\def \figwidthb {0.22}
	\def \figwidthc {0.31}
	\def \figwh {0.25}
	\centering

	\includegraphics[width=\figwidth\linewidth,height=\figwh\linewidth]{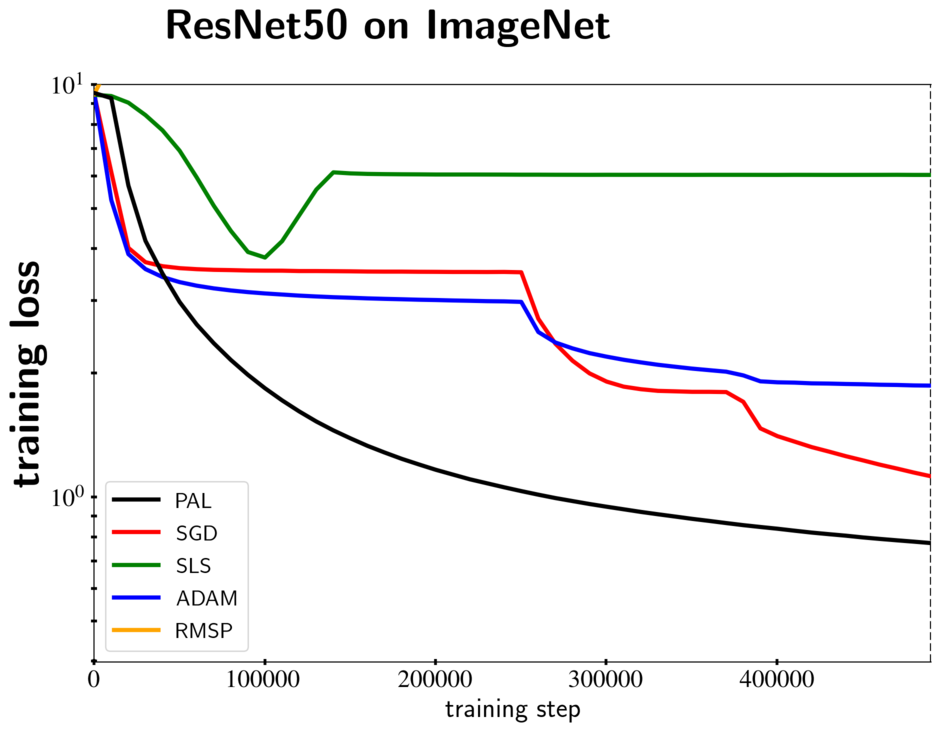}\quad
	\includegraphics[width=\figwidth\linewidth,height=\figwh\linewidth]{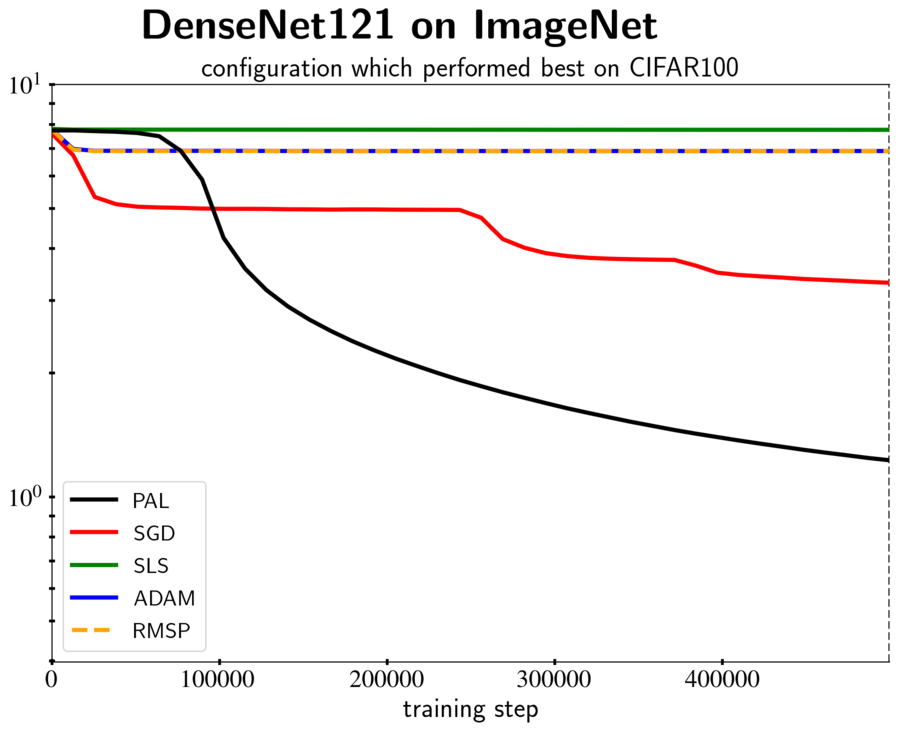} \quad
	\includegraphics[width=\figwidthc\linewidth,height=\figwh\linewidth]{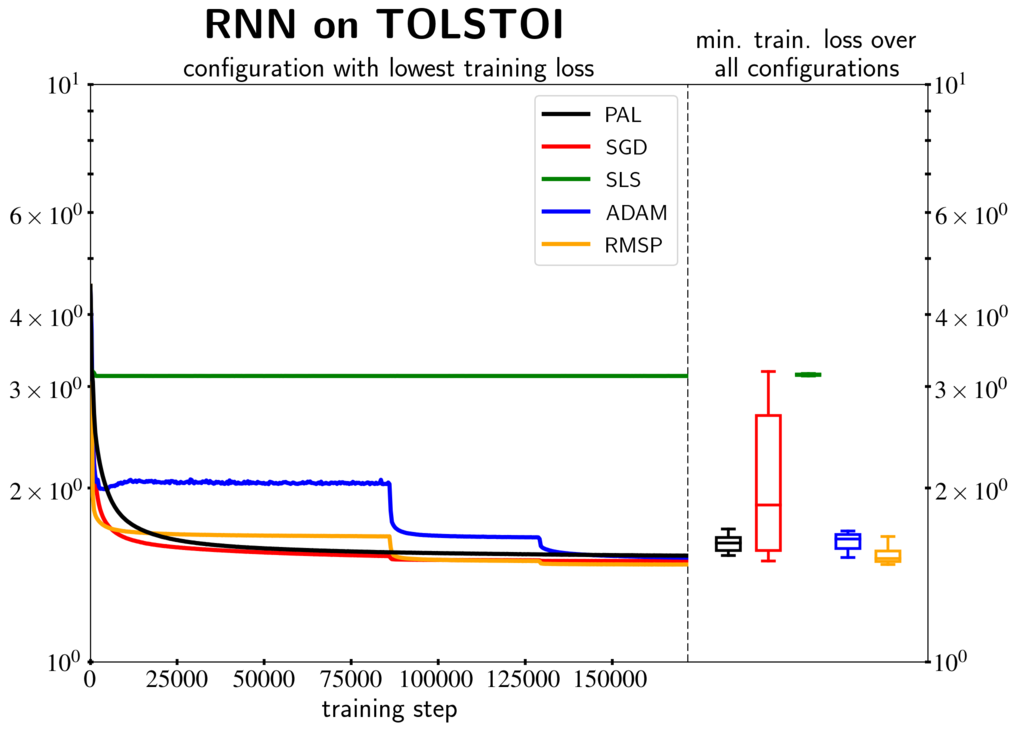}

	\includegraphics[width=\figwidth\linewidth,height=\figwh\linewidth]{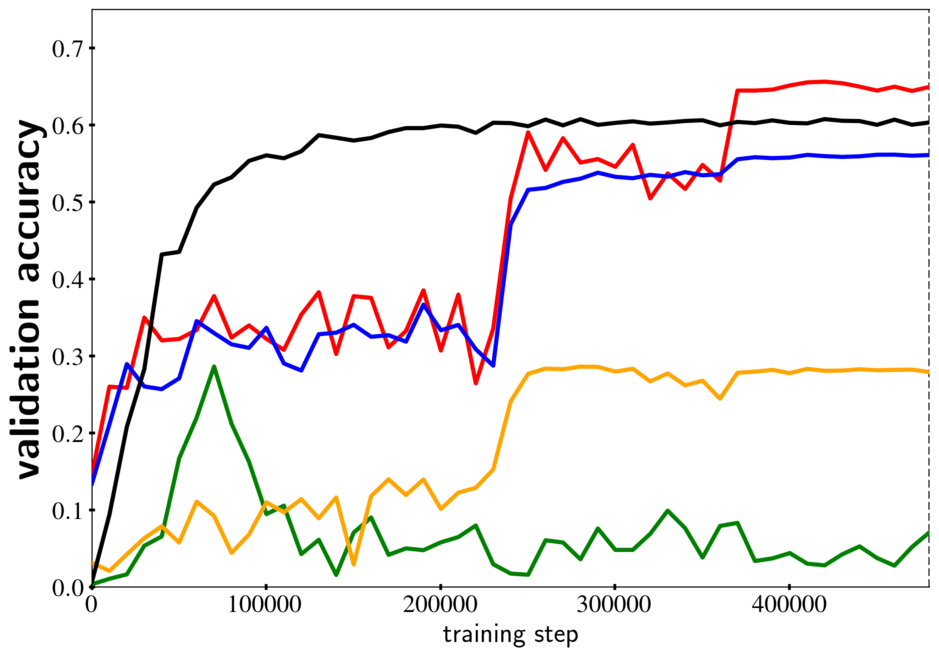}\quad
	\includegraphics[width=\figwidth\linewidth,height=\figwh\linewidth]{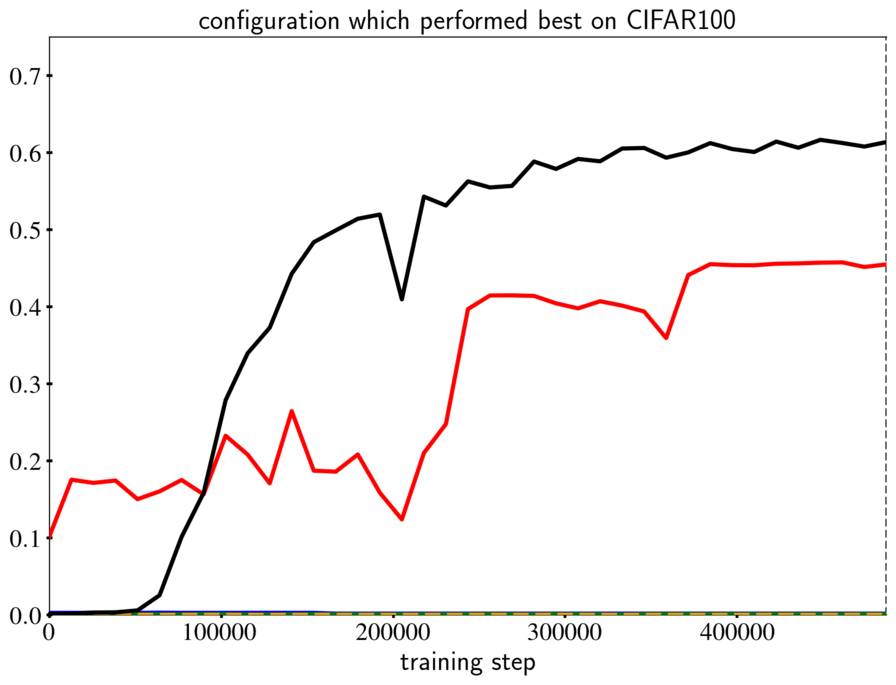} \quad
	\includegraphics[width=\figwidthc\linewidth,height=\figwh\linewidth]{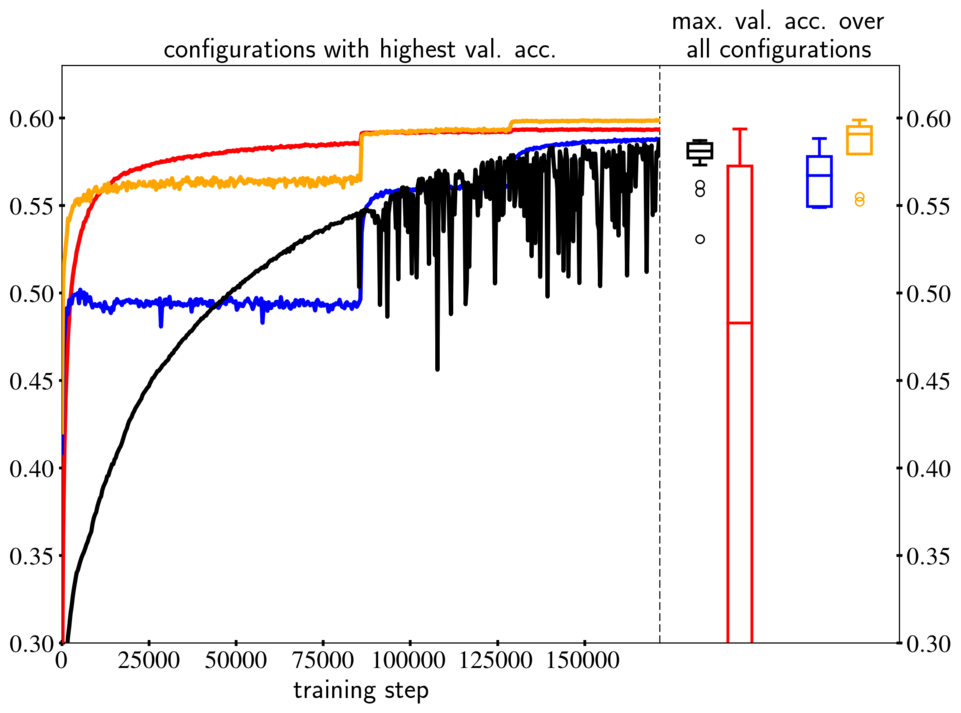}

	\includegraphics[width=\figwidth\linewidth,height=\figwh\linewidth]{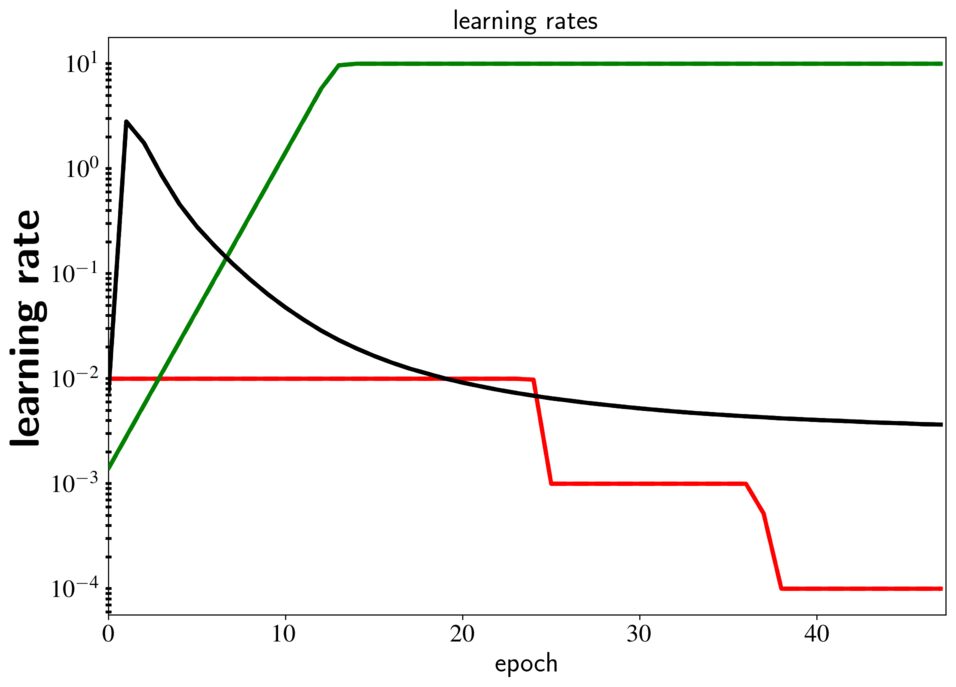}\quad
	\includegraphics[width=\figwidth\linewidth,height=\figwh\linewidth]{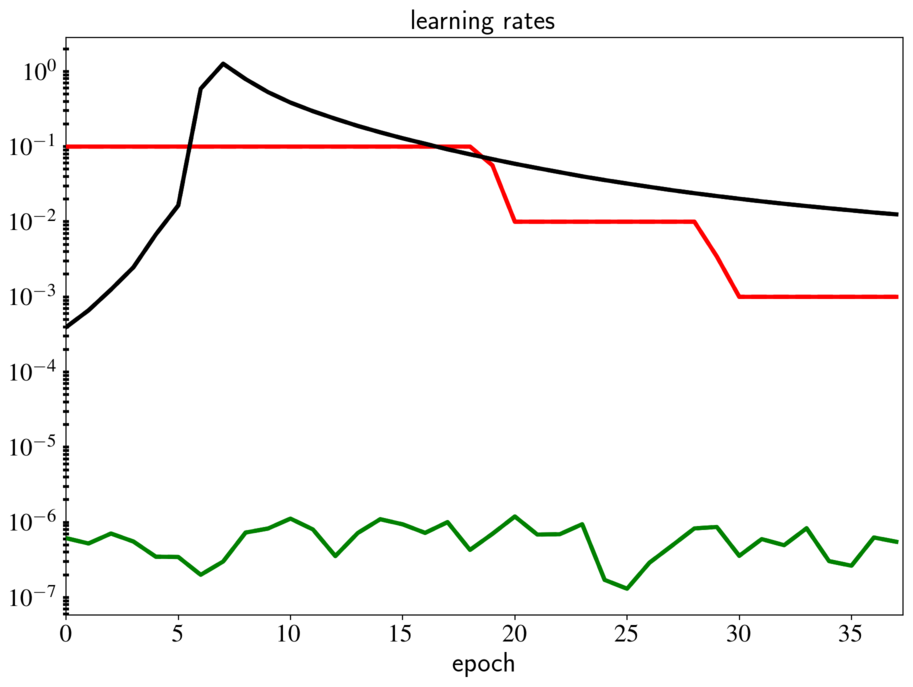} \quad
	\includegraphics[width=\figwidthc\linewidth,height=\figwh\linewidth]{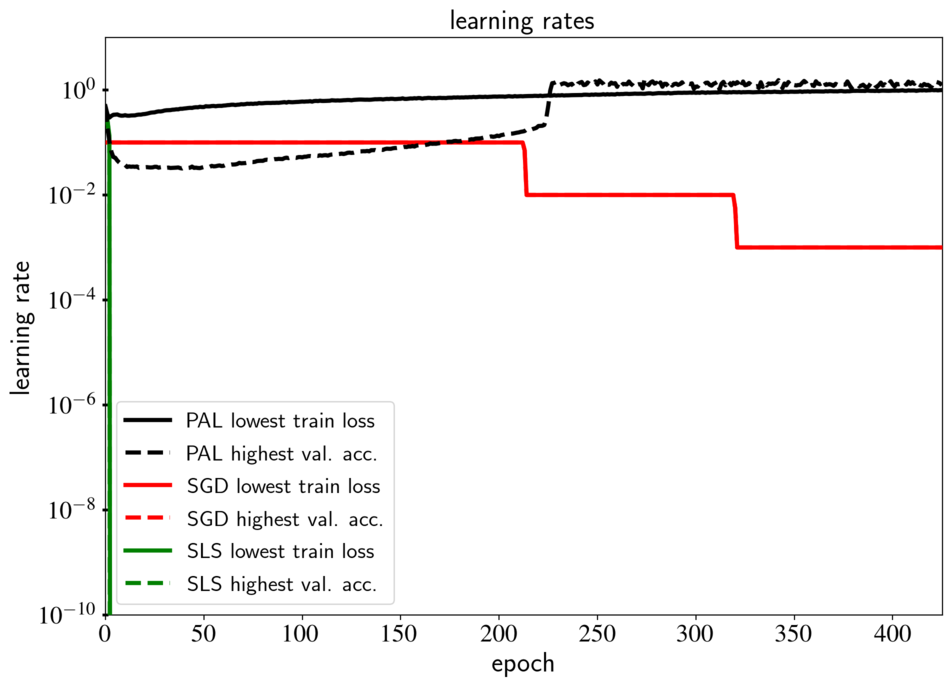}	
	
	\caption[]{Comparison of \pal\ to \textit{SGD}, \textit{SLS}, \textit{ADAM}, \textit{RMSProp} on training loss, validation accuracy and learning rates on \textbf{Imagenet}, and a simple RNN, trained on the \textbf{Tolstoi} War and Peace dataset. Learning rates are averaged over epochs. For Imagenet the best hyperparameter configuration from the CIFAR-100 evaluation were used to test hyperparameter transferability.}
	\label{fig.IMAGENETplots}
\end{figure*}
\FloatBarrier
\subsection{Wall-clock time comparison}
\label{subsec:time_per_epoch}
\begin{table}[h]
	\caption{Required seconds per epoch of \pal, \textit{SLS}, \textit{ALIG}, \textit{SGDHD}, \textit{COCOB} and \textit{SGD} on CIFAR-10. RMSP and ADAM reach a similar speed as SGD. The comparison was performed on a Nvidia Geforce GTX 1080 TI. \pal\ and \textit{SLS} perform slower, since they have to measure additional losses, whereas the additional operations of \textit{ALIG}, \textit{SGDHD}, \textit{COCOB}  tend to be cheap.}
	\centering
	\begin{tabular}{lllllllll}	
		\toprule	
		network&\vtop{\hbox{\strut seconds /  }\hbox{\strut epoch \pal}}&\vtop{\hbox{\strut  }\hbox{\strut  \textit{SLS} }}&\vtop{\hbox{\strut  }\hbox{\strut  \textit{SGD}}}&\vtop{\hbox{\strut  }\hbox{\strut  \textit{ALIG}}}&\vtop{\hbox{\strut   }\hbox{\strut  \textit{SGDHD}}}&\vtop{\hbox{\strut   }\hbox{\strut  \textit{COCOB}}}  \\ \midrule
		ResNet32&$20.9$&$21.7$&$10.7$&$11.0$&$11.1$&$16.4$\\     
		MobilenetV2&$ 53.2$&$52.4$&$34.1$&$34.01$&$34.2$&$36.6$ \\ 
		EfficientNet&$55.5$&$52.2$&$30.7$&$31.2$&$32.2$&$37.5$ \\  
		DenseNet40&$88.8$&$87.5$&$59.7$&$61.3$&$64.6$&$61.4$ \\   
		\bottomrule	
	\end{tabular}
	\label{tbl:time_per_epoch}
\end{table}
\vfill
\subsection{SLS ResNet34 test case re-implementation}
\label{subsec:slsre-impl}
In the shown experiments and in contrast to the evaluation  of \textit{SLS} in \cite{backtracking_line_search_NIPS}, we used Tensorflow default Xavier weight initialization \cite{xavier_weight_intializaiton} versus PyTorch default Lecun initialization \cite{lecun_weight_intializaiton}. In addition, we used L2 regularisation versus no regularization. Furthermore, default implementations of networks for both frameworks have small differences. All in all, those differences usually influence the optimizer performance only marginally, as given by the fact that all other investigated optimizers perform well. However, in this case of \textit{SLS} we see significant differences.\\
To prove that our implementation of \textit{SLS} is correct, we re-implemented \cite{backtracking_line_search_NIPS}'s ResNet34 test case on CIFAR-10 in Tensorflow and achieved similar results as \cite{backtracking_line_search_NIPS}. SLS  shows well performance and is not significantly overfitting as it does in in Section \ref{sec:results}. 

\begin{figure}[h!]
	\newcommand\picscale{0.4}
	\newcommand\anglemin{40}
	\newcommand\anglemax{110}
	\newcommand\angles{125000}
	\newcommand\lossmin{0}
	\newcommand\lossmax{10}
	\newcommand\evalmin{0}
	\newcommand\evalmax{1}
	\newcommand\epochs{280}
	\newcommand\picheight{0.5\linewidth}
	\newcommand\legendwidth{9cm}
	\newcommand\legendheight{3cm}
\centering
		\tikzsetnextfilename{pal_sls_reengineering_1}
		\begin{tikzpicture}[scale=\picscale] 
		\begin{axis}[
		width=\linewidth, 
		height=\picheight,
		grid=major, 
		grid style={dashed,gray!30}, 
		xlabel=epoch, 
		xlabel style={font=\LARGE},
		ylabel= validation accuracy,
		ylabel style={font=\LARGE},
		xmin=0,xmax=179,
		ymin=\evalmin,ymax=\evalmax,
		legend pos=south east, 
		x tick label style={rotate=0,anchor=near xticklabel,font=\LARGE}, 
		y tick label style={font=\LARGE},
		p1/.style={draw=blue,line width=2pt},
		title= validation accuracy CIFAR-10,
		title style={font=\Huge},
		]
		\addplot [p1] table[x=Step,y=Value,col sep=comma] {plot_data/sls_comparison/sls_eval_acc_cifar10.csv}; 

		\legend{SLS}
		\end{axis}
		\end{tikzpicture}\quad \tikzsetnextfilename{pal_sls_reengineering_2}
				\begin{tikzpicture}[scale=\picscale] 
				\begin{axis}[
				ymode=log,
				width=\linewidth, 
				height=\picheight,
				grid=major, 
				grid style={dashed,gray!30}, 
				xlabel=epoch, 
				xlabel style={font=\LARGE},
				ylabel= training loss,
				ylabel style={font=\LARGE},
				xmin=0,xmax=179,
				ymin=\lossmin,ymax=\lossmax,
				legend style={at={(-0.7,1)},anchor=north,font=\huge,minimum height= \legendheight,draw=none}, 
				x tick label style={rotate=0,anchor=near xticklabel,font=\LARGE}, 
				y tick label style={font=\LARGE},
				p1/.style={draw=blue,line width=2pt},
				title= training loss CIFAR-10,
				title style={font=\Huge},
				]
				\addplot [p1] table[x=Step,y=Value,col sep=comma] {plot_data/sls_comparison/sls_train_loss_cifar10.csv}; 
				\end{axis}
			\end{tikzpicture}\\
		\tikzsetnextfilename{pal_sls_reengineering_3}
		\begin{tikzpicture}[scale=\picscale] 
		\begin{axis}[
		width=\linewidth, 
		height=\picheight,
		grid=major, 
		grid style={dashed,gray!30}, 
		xlabel=epoch, 
		xlabel style={font=\LARGE},
		ylabel= validation accuracy,
		ylabel style={font=\LARGE},
		xmin=0,xmax=179,
		ymin=\evalmin,ymax=\evalmax,
		legend style=={anchor=south east}, 
		x tick label style={rotate=0,anchor=near xticklabel,font=\LARGE}, 
		y tick label style={font=\LARGE},
		p1/.style={draw=blue,line width=2pt},
		title= validation accuracy CIFAR-100,
		title style={font=\Huge},
		]
		\addplot [p1] table[x=Step,y=Value,col sep=comma] {plot_data/sls_comparison/sls_eval_acc_cifar100.csv}; 

		\end{axis}
	\end{tikzpicture}\quad
			\tikzsetnextfilename{pal_sls_reengineering_4}
			\begin{tikzpicture}[scale=\picscale] 
			\begin{axis}[
			ymode=log,
			width=\linewidth, 
			height=\picheight,
			grid=major, 
			grid style={dashed,gray!30}, 
			xlabel=epoch, 
			xlabel style={font=\LARGE},
			ylabel= training loss,
			ylabel style={font=\LARGE},
			xmin=0,xmax=179,
			ymin=\lossmin,ymax=\lossmax,
			legend style={at={(-0.7,1)},anchor=north,font=\huge,minimum height= \legendheight,draw=none}, 
			x tick label style={rotate=0,anchor=near xticklabel,font=\LARGE}, 
			y tick label style={font=\LARGE},
			p1/.style={draw=blue,line width=2pt},
			title= training loss CIFAR-100,
			title style={font=\Huge},
			]
			\addplot [p1] table[x=Step,y=Value,col sep=comma] {plot_data/sls_comparison/sls_train_loss_cifar100.csv}; 
			\end{axis}
		\end{tikzpicture}

\caption{On the re-implemented ResNet34 test case of \cite{backtracking_line_search_NIPS} SLS  shows well performance and is not significantly overfitting as it does in in Section \ref{sec:results}}
\label{fig:evalacc}
\end{figure}
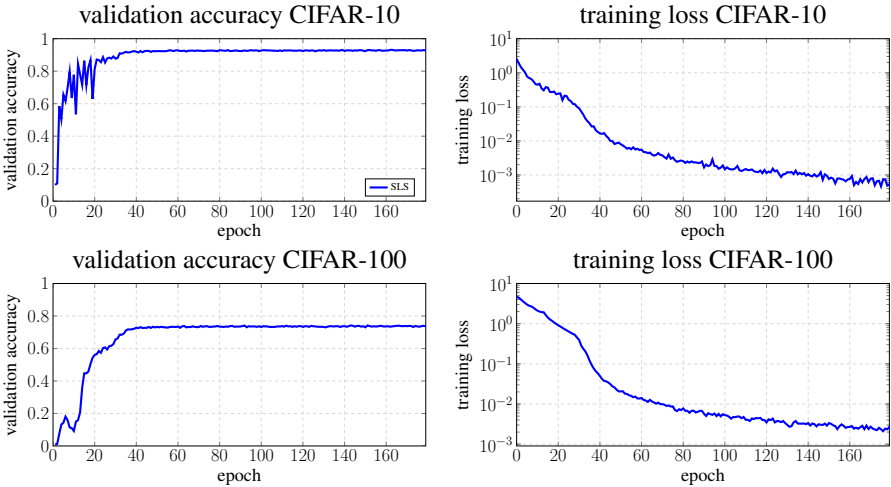
\vfill
\pagebreak

\vfill
\pagebreak
\subsection{Parabolic property in adapted directions:}
\label{fig:Parabolic property in adapted}
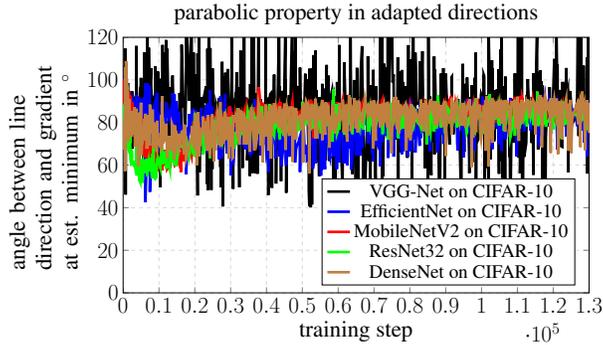
\begin{figure}[h]
	\newcommand\picscale{0.5}
	\newcommand\anglemin{0}
	\newcommand\anglemax{120}
	\newcommand\angles{130000}
	\newcommand\lossmin{0}
	\newcommand\lossmax{1}
	\newcommand\evalmin{0}
	\newcommand\evalmax{1}
	\newcommand\epochs{280}
	\newcommand\picheight{0.6\textwidth}
	\newcommand\legendwidth{9cm}
	\newcommand\legendheight{3cm}
	\centering

				\tikzsetnextfilename{pal_angles_continuous_momentum}	
				\begin{tikzpicture}[scale=\picscale] 
				\begin{axis}[
				width=\textwidth, 
				height=\picheight,
				grid=major, 
				grid style={dashed,gray!30}, 
				xlabel=training step, 
				xlabel style={font=\LARGE},
				ylabel=angle between line direction and gradient at est. minimum in $^\circ$,
				ylabel style={font=\LARGE,text width=0.4\textwidth},
				xmin=0,xmax=\angles,
				ymin=\anglemin,ymax=\anglemax,
				title={parabolic property in adapted directions},
				legend pos= south east,
				legend style={font=\Large},
				x tick label style={rotate=0,anchor=near xticklabel,font=\LARGE}, 
				y tick label style={font=\LARGE},
				p1/.style={draw=blue,line width=2pt},
				p2/.style={draw=red,line width=2pt},
				p3/.style={draw=green,line width=2pt},
				p4/.style={draw=brown,line width=2pt},
				p5/.style={draw=black,line width=2pt},
				title style={font=\LARGE},
				]
				\addplot [p5] table[x=Step,y=Value,col sep=comma] {plot_data/angle_data/cifar10_mom/vgg_run-PAL_1_PAL_45000_150000_128_1_0_1_3.16_1_0_0_log-tag-_data_angle_per_step.csv}; 
				\addplot [p1] table[x=Step,y=Value,col sep=comma] {plot_data/angle_data/cifar10_mom/run-EFFICIENTNET_models_PAL_1_PAL_45000_150000_128_0.1_0.4_0.8_3.16_1_0_0_log-tag-_data_angle_per_step.csv}; 
				\addplot [p2] table[x=Step,y=Value,col sep=comma] {plot_data/angle_data/cifar10_mom/run-MOBILENETV2_models_PAL_1_PAL_45000_150000_128_0.1_0.4_0.8_3.16_1_0_0_log-tag-_data_angle_per_step.csv}; 
				\addplot [p3] table[x=Step,y=Value,col sep=comma] {plot_data/angle_data/cifar10_mom/resnet-run-models_PAL_1_PAL_45000_150000_128_0.316_0.4_1_3.16_1_0_0_log-tag-_data_angle_per_step.csv}; 							
				\addplot [p4] table[x=Step,y=Value,col sep=comma] {plot_data/angle_data/cifar10_mom/run-DENSENET40_models_PAL_1_PAL_45000_150000_128_0.0316_0.4_1_3.16_1_0_0_log-tag-_data_angle_per_step.csv}; 
		 		\legend{VGG-Net on CIFAR-10, EfficientNet on CIFAR-10, MobileNetV2 on CIFAR-10, ResNet32 on CIFAR-10, DenseNet on CIFAR-10, }
				\end{axis}
				\end{tikzpicture}
		\caption{Angles between the line direction and the gradient at the estimated minimum measured on the same batch are plotted over a whole training process on several networks on CIFAR-10. This figure clarifies, that parabolic observation is also valid if a \textbf{direction adaptation factor of} $\mathbf{0.4}$ is applied. Measuring step sizes and update step adaptations factors (see Sections \ref{sec:paremeter_update_rule},\ref{subsec:features}) were set to fit loss along the line decently.}
		\label{fig:angles_continuous_mom}
\end{figure}
\FloatBarrier
\subsection{Influence of dynamic step sizes and the direction adaptation}
\label{sec:abldation_direction_adaptation}
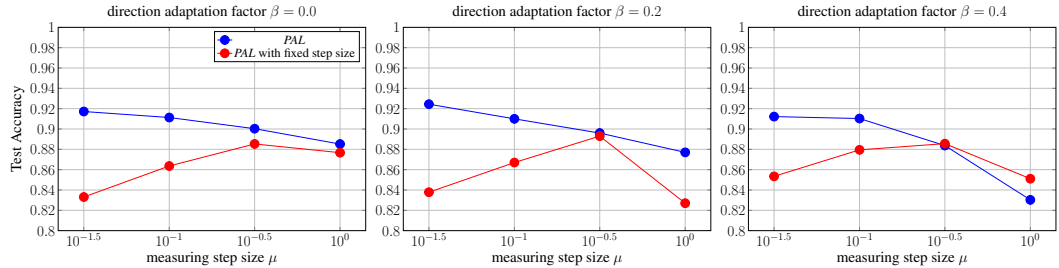
\begin{figure*}[h!]
	\newcommand\picscale{0.3}
	\newcommand\anglemin{40}
	\newcommand\anglemax{110}
	\newcommand\angles{0.8E5}
	\newcommand\lossmin{0}
	\newcommand\lossmax{1}
	\newcommand\evalmin{0}
	\newcommand\evalmax{1}
	\newcommand\epochs{160}
	\newcommand\picheight{0.7\linewidth}
	\newcommand\legendwidth{10cm}
	\newcommand\legendheight{3cm}
	
	\newcommand\marksize{5pt}
	\newcommand\marker{oplus*}
	\begin{center}
		\tikzsetnextfilename{pal_cg_ablation_study_1}
		\begin{tikzpicture}[scale=\picscale] 
		\begin{axis}[
		legend style={font=\Large},
		width=\linewidth, 
		height=\picheight,
		grid=major, 
		xlabel=measuring step size $\mu$, 
		xlabel style={font=\LARGE},
		ylabel=Test Accuracy,
		ylabel style={font=\LARGE},
		ymin=0.8,ymax=1,
		xmode=log,
		xtick={0.0316,0.1,0.316,1.0},
		x tick label style={rotate=0,anchor=near xticklabel,font=\LARGE}, 
		y tick label style={font=\LARGE},
		p1/.style={color=blue,mark size=\marksize,mark=\marker},
		p2/.style={color=red,mark size=\marksize,mark=\marker},
		p3/.style={color=black,mark size=\marksize,mark=\marker},
		title={direction adaptation factor $\beta = 0.0$},
		title style={font=\LARGE},
		]
		\addplot [p1] table[x=Step,y=TestAcc,col sep=comma] {plot_data/cg_ablation_study/beta0_dynamic.csv}; 
		\addplot [p2] table[x=Step,y=TestAcc,col sep=comma] {plot_data/cg_ablation_study/beta0_fixed.csv};

		\legend{\pal, \pal\ with fixed step size}
		\end{axis}
		\end{tikzpicture}
				\tikzsetnextfilename{pal_cg_ablation_study_2}
				\begin{tikzpicture}[scale=\picscale] 
				\begin{axis}[
				width=\linewidth, 
				height=\picheight,
				grid=major, 
				xlabel=measuring step size $\mu$, 
				xlabel style={font=\LARGE},
				ylabel style={font=\LARGE},
				ymin=0.8,ymax=1,
				xmode=log,
				xtick={0.0316,0.1,0.316,1.0},
				x tick label style={rotate=0,anchor=near xticklabel,font=\LARGE}, 
				y tick label style={font=\LARGE},
				p1/.style={color=blue,mark size=\marksize,mark=\marker},
				p2/.style={color=red,mark size=\marksize,mark=\marker},
				p3/.style={color=black,mark size=\marksize,mark=\marker},
				title={direction adaptation factor $\beta = 0.2$},
				title style={font=\LARGE},
				]
				\addplot [p1] table[x=Step,y=TestAcc,col sep=comma] {plot_data/cg_ablation_study/beta0_2_dynamic.csv}; 
				\addplot [p2] table[x=Step,y=TestAcc,col sep=comma] {plot_data/cg_ablation_study/beta0_2_fixed.csv}; 

				\end{axis}
				\end{tikzpicture}
				\tikzsetnextfilename{pal_cg_ablation_study_3}
						\begin{tikzpicture}[scale=\picscale] 
						\begin{axis}[
						width=\linewidth, 
						height=\picheight,
						grid=major, 
						xlabel=measuring step size $\mu$, 
						xlabel style={font=\LARGE},
						ylabel style={font=\LARGE},
						ymin=0.8,ymax=1,
						xmode=log,
						xtick={0.0316,0.1,0.316,1.0},
						x tick label style={rotate=0,anchor=near xticklabel,font=\LARGE}, 
						y tick label style={font=\LARGE},
						p1/.style={color=blue,mark size=\marksize,mark=\marker},
						p2/.style={color=red,mark size=\marksize,mark=\marker},
						p3/.style={color=black,mark size=\marksize,mark=\marker},
						title={direction adaptation factor $\beta = 0.4$},
						title style={font=\LARGE},
						]
						\addplot [p1] table[x=Step,y=TestAcc,col sep=comma] {plot_data/cg_ablation_study/beta0_4_dynamic.csv}; 
						\addplot [p2] table[x=Step,y=TestAcc,col sep=comma] {plot_data/cg_ablation_study/beta0_4_fixed.csv};
						\end{axis}
						\end{tikzpicture}
		\end{center}
		\caption{Analysis of the influences of dynamic step sizes and the direction adaptation factor $\beta$.}
		\label{fig:cg_ablation_study}
\end{figure*}

This section analyses, whether \pal's performance originates from dynamically chosen step sizes or from the the non-linear conjugate gradient like update step adaptation. We consider EfficientNets trained on CIFAR-10, since for those the update step adaptation factor $\beta$ is needed to achieve optimal results. We consider the following 6 scenarios: \textbf{1,2)} \pal\ without update step adaptation ($\beta=0$) and with and without dynamic step sizes (Figure \ref{fig:cg_ablation_study} left). \textbf{3,4)} \pal\ with a update step adaptation of $0.2$ and with and without dynamic step sizes (Figure \ref{fig:cg_ablation_study} middle). \textbf{5,6)}  \pal\ with a update step adaptation of $0.4$  with and without fixed step sizes (Figure \ref{fig:cg_ablation_study} right). The case with fixed step sizes result in  in normalized SGD (NSGD) with a momentum factor $\beta$. As fixed update step size we use the measuring step size $\mu$.

The results show that dynamic step sizes increase the performance always if direction adaptation is not applied and if it is applied in 6 out of 8 cases. Direction adaptation can increase or decrease the performance in both, the dynamic and the fixed step size cases. 
The best performance is achieved with a direction adaptation factor of $0.2$ and a measuring step size of $10^{-1.5}$, which shows that both factors influence the best results in this scenario.

\vfill

\pagebreak
\subsection{Sensitivity analysis:}
\label{sec:sens_analysis}
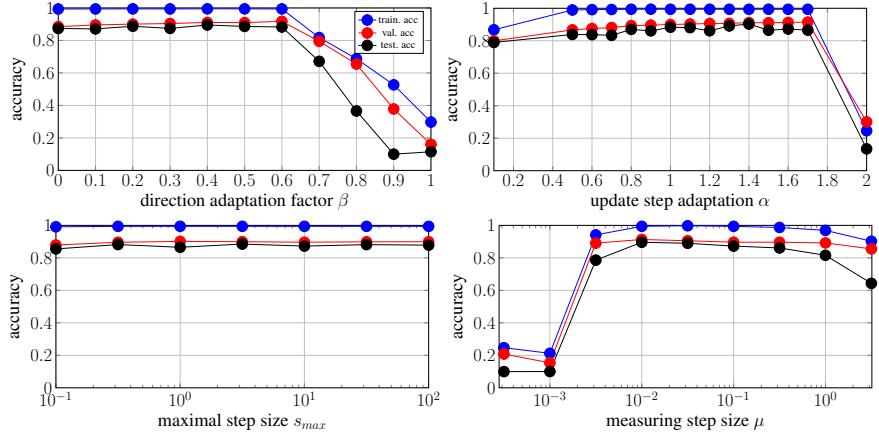
\begin{figure*}[h!]
	\newcommand\picscale{0.4}
	\newcommand\anglemin{40}
	\newcommand\anglemax{110}
	\newcommand\angles{0.8E5}
	\newcommand\lossmin{0}
	\newcommand\lossmax{1}
	\newcommand\evalmin{0}
	\newcommand\evalmax{1}
	\newcommand\epochs{160}
	\newcommand\picheight{0.5\linewidth}
	\newcommand\legendwidth{10cm}
	\newcommand\legendheight{3cm}
	
	\newcommand\marksize{5pt}
	\newcommand\marker{oplus*}
	\begin{center}
		
		\tikzsetnextfilename{pal_sensitivity_analysis_1}
		\begin{tikzpicture}[scale=\picscale] 
		\begin{axis}[
		width=\linewidth, 
		height=\picheight,
		grid=major, 
		xlabel=direction adaptation factor $\beta$, 
		xlabel style={font=\LARGE},
		ylabel=accuracy,
		ylabel style={font=\LARGE},
		xmin=0,xmax=1,
		ymin=0.0,ymax=1,
		x tick label style={rotate=0,anchor=near xticklabel,font=\LARGE}, 
		y tick label style={font=\LARGE},
		p1/.style={color=blue,mark size=\marksize,mark=\marker},
		p2/.style={color=red,mark size=\marksize,mark=\marker},
		p3/.style={color=black,mark size=\marksize,mark=\marker},
		title style={font=\Huge},
		]
		\addplot [p1] table[x=Step,y=TrainAcc,col sep=comma] {plot_data/sensitivity_analysis/cg.csv}; 
		\addplot [p2] table[x=Step,y=EvalAcc,col sep=comma] {plot_data/sensitivity_analysis/cg.csv}; 
		\addplot [p3] table[x=Step,y=TestAcc,col sep=comma] {plot_data/sensitivity_analysis/cg.csv}; 
		\legend{train. acc, val. acc, test. acc}
		\end{axis}
	\end{tikzpicture}\tikzsetnextfilename{pal_sensitivity_analysis_2}
		\begin{tikzpicture}[scale=\picscale] 
		\begin{axis}[
		width=\linewidth, 
		height=\picheight,
		grid=major, 
		xlabel= update step adaptation $\alpha$, 
		xlabel style={font=\LARGE},
		ylabel=accuracy,
		ylabel style={font=\LARGE},
		xmin=0.1,xmax=2,
		ymin=0.0,ymax=1,
		x tick label style={rotate=0,anchor=near xticklabel,font=\LARGE}, 
		y tick label style={font=\LARGE},
		p1/.style={color=blue,mark size=\marksize,mark=\marker},
		p2/.style={color=red,mark size=\marksize,mark=\marker},
		p3/.style={color=black,mark size=\marksize,mark=\marker},
		title style={font=\Huge},
		]
		\addplot [p1] table[x=Step,y=TrainAcc,col sep=comma] {plot_data/sensitivity_analysis/loose_approx_correct.csv}; 
		\addplot [p2] table[x=Step,y=EvalAcc,col sep=comma] {plot_data/sensitivity_analysis/loose_approx_correct.csv}; 
		\addplot [p3] table[x=Step,y=TestAcc,col sep=comma] {plot_data/sensitivity_analysis/loose_approx_correct.csv}; 
		\end{axis}
		\end{tikzpicture}\\\tikzsetnextfilename{pal_sensitivity_analysis_3}
		\begin{tikzpicture}[scale=\picscale] 
		\begin{axis}[
		width=\linewidth, 
		height=\picheight,
		grid=major, 
		xlabel=maximal step size $s_{max}$, 
		xlabel style={font=\LARGE},
		ylabel=accuracy,
		ylabel style={font=\LARGE},
		xmin=0.1,xmax=100,
		ymin=0.0,ymax=1,
		xmode=log,
		x tick label style={rotate=0,anchor=near xticklabel,font=\LARGE}, 
		y tick label style={font=\LARGE},
		p1/.style={color=blue,mark size=\marksize,mark=\marker},
		p2/.style={color=red,mark size=\marksize,mark=\marker},
		p3/.style={color=black,mark size=\marksize,mark=\marker},
		title style={font=\Huge},
		]
		\addplot [p1] table[x=Step,y=TrainAcc,col sep=comma] {plot_data/sensitivity_analysis/maximal_step.csv}; 
		\addplot [p2] table[x=Step,y=EvalAcc,col sep=comma] {plot_data/sensitivity_analysis/maximal_step.csv}; 
		\addplot [p3] table[x=Step,y=TestAcc,col sep=comma] {plot_data/sensitivity_analysis/maximal_step.csv};

		\end{axis}
		\end{tikzpicture}\tikzsetnextfilename{pal_sensitivity_analysis_4}
		\begin{tikzpicture}[scale=\picscale] 
		\begin{axis}[
		width=\linewidth, 
		height=\picheight,
		grid=major, 
		xlabel=measuring step size $\mu$, 
		xlabel style={font=\LARGE},
		ylabel=accuracy,
		ylabel style={font=\LARGE},
		xmin=0.00028,xmax=3.2,
		ymin=0.0,ymax=1,
		xmode=log,
		x tick label style={rotate=0,anchor=near xticklabel,font=\LARGE}, 
		y tick label style={font=\LARGE},
		p1/.style={color=blue,mark size=\marksize,mark=\marker},
		p2/.style={color=red,mark size=\marksize,mark=\marker},
		p3/.style={color=black,mark size=\marksize,mark=\marker},
		title style={font=\Huge},
		]
		\addplot [p1] table[x=Step,y=TrainAcc,col sep=comma] {plot_data/sensitivity_analysis/measuring_step.csv}; 
		\addplot [p2] table[x=Step,y=EvalAcc,col sep=comma] {plot_data/sensitivity_analysis/measuring_step.csv}; 
		\addplot [p3] table[x=Step,y=TestAcc,col sep=comma] {plot_data/sensitivity_analysis/measuring_step.csv}; 
		\end{axis}
		\end{tikzpicture}
		\end{center}
		\caption{Sensitivity analysis for PAL on a ResNet32 trained on CIFAR-10. The baseline parameters are: $\mu=0.1,\beta=0.2,\alpha=1.0,s_{max}=10$. It shows that $\beta$ should be chosen $\leq 0.6$. $\alpha$ has a low sensitivity, but with a value of $1.4$ it reaches best performance. $s_{max}$ has a low sensitivity and all investigated values perform similarly. $\mu$ should be chosen between $10^{-2}$ and $10^{-0.5}$.}
		\label{fig:sensitivity analysis}
\end{figure*}

\begin{figure}[h!]
\centering
	\newcommand\picscale{0.5}
	\newcommand\anglemin{40}
	\newcommand\anglemax{110}
	\newcommand\angles{0.8E5}
	\newcommand\lossmin{0}
	\newcommand\lossmax{1}
	\newcommand\evalmin{0}
	\newcommand\evalmax{1}
	\newcommand\epochs{160}
	\newcommand\picheight{0.5\linewidth}
	\newcommand\legendwidth{10cm}
	\newcommand\legendheight{3cm}
	
	\newcommand\marksize{5pt}
	\newcommand\marker{oplus*}
\tikzsetnextfilename{pal_sensitivity_analysis_mu}
\begin{tikzpicture}[scale=\picscale] 
		\begin{axis}[
		width=\linewidth, 
		height=\picheight,
		grid=major, 
		xlabel=measuring step size $\mu$, 
		xlabel style={font=\LARGE},
		ylabel=val. accuracy,
		ylabel style={font=\LARGE},
		xmin=0.00028,xmax=3.2,
		ymin=0.0,ymax=1,
		xmode=log,
		legend style={cells={anchor=center},at={(0.7,0.6)}},
		x tick label style={rotate=0,anchor=near xticklabel,font=\LARGE}, 
		y tick label style={font=\LARGE},
		p1/.style={color=blue,mark size=\marksize,mark=\marker},
		p2/.style={color=red,mark size=\marksize,mark=\marker},
		p3/.style={color=black,mark size=\marksize,mark=\marker},
		p4/.style={color=green,mark size=\marksize,mark=\marker},
		title style={font=\Huge},
		]
		\addplot [p2] table[x=Step,y=EvalAcc,col sep=comma] {plot_data/sensitivity_analysis/mu/measuring_step_resnet.csv}; 
		\addplot [p1] table[x=Step,y=EvalAcc,col sep=comma] {plot_data/sensitivity_analysis/mu/measuring_step_densenet.csv};
		\addplot [p3] table[x=Step,y=EvalAcc,col sep=comma] {plot_data/sensitivity_analysis/mu/measuring_step_mobilenet.csv};
		\addplot [p4] table[x=Step,y=EvalAcc,col sep=comma] {plot_data/sensitivity_analysis/mu/measuring_step_efficientnet.csv};
		\legend{ResNet,DenseNet,MobileNet,EfficientNet}
		\end{axis}
\end{tikzpicture}
\caption{Sensitivity of the measuring step size $\mu$ of PAL for several models on CIFAR-10. \pal\ shows low sensitivity.}
\label{fig:sensitivity analysis_mu}
\end{figure}
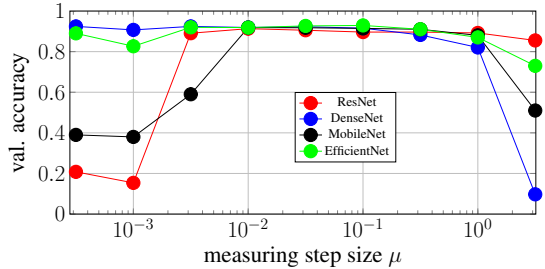

All in all \pal\ tends to have a low hyperparameter sensitivity as shown in Figure \ref{fig:sensitivity analysis}. Since $\mu$ is the most sensitive hyperparameter we analyzed its sensitivity over several further models trained on CIFAR-10 (see Figure \ref{fig:sensitivity analysis_mu}).

\FloatBarrier
\clearpage
\pagebreak
\subsection{Comparison to Probabilistic Line-Search (PLS):}
\begin{figure}[h]
	\newcommand\picscale{0.5}
	\newcommand\anglemin{40}
	\newcommand\anglemax{110}
	\newcommand\angles{150000}
	\newcommand\lossmin{0}
	\newcommand\lossmax{10}
	\newcommand\evalmin{0}
	\newcommand\evalmax{1}
	\newcommand\epochs{280}
	\newcommand\picheight{0.4\textwidth}
	\newcommand\legendwidth{9cm}
	\newcommand\legendheight{3cm}
	\tikzsetnextfilename{pal_pls_comparison_1}
	\begin{tikzpicture}[scale=\picscale] 
		\begin{axis}[
			ymode=log,
			width=\textwidth*0.85, 
			height=\picheight,
			grid=major, 
			grid style={dashed,gray!30}, 
			xlabel=training step, 
			xlabel style={font=\LARGE},
			ylabel= step size,
			ylabel style={font=\LARGE},
			xmin=0,xmax=150000,
			ymin=0,ymax=15,
			legend style={font=\large}, 
			legend pos=north east, 
			x tick label style={rotate=0,anchor=near xticklabel,font=\LARGE}, 
			y tick label style={font=\LARGE},
			ylabel shift={0.2cm},
			p1/.style={draw=black,line width=2pt},
			p2/.style={draw=blue,line width=2pt},
			title style={font=\Huge},
			]
			\addplot [p1] table[x=Step,y=Value,col sep=comma] {pls_data/pls_step_size.csv}; 
			\addplot [p2] table[x=Step,y=Value,col sep=comma] {pls_data/pal_step_size.csv}; 
			\legend{PLS,PAL}
		\end{axis}
	\end{tikzpicture}\hspace{0.5cm} \tikzsetnextfilename{pal_pls_comparison_2}
	\begin{tikzpicture}[scale=\picscale] 
		\begin{axis}[
			width=\textwidth*0.85, 
			height=\picheight,
			grid=major, 
			grid style={dashed,gray!30}, 
			xlabel=epoch, 
			xlabel style={font=\LARGE},
			ylabel= val. accuracy,
			ylabel style={font=\LARGE},
			ylabel shift={0.2cm},
			xmin=0,xmax=330,
			ymin=0.7,ymax=0.95,
			legend style={font=\large},
			legend pos=south east, 
			x tick label style={rotate=0,anchor=near xticklabel,font=\LARGE}, 
			y tick label style={font=\LARGE},
			p1/.style={draw=black,line width=2pt},
			p2/.style={draw=blue,line width=2pt},
			title style={font=\Huge},
			]
			\addplot [p1] table[x=Step,y=Value,col sep=comma] {pls_data/pls_eval_acc.csv}; 
			\addplot [p2] table[x=Step,y=Value,col sep=comma] {pls_data/pal_eval_acc.csv}; 
			\legend{PLS,PAL}
		\end{axis}
	\end{tikzpicture}
	\caption{Comparison of PAL to Probabilistic Line Search \cite{probabilisticLineSearch}}
	
	\label{plscomparison}
\end{figure}
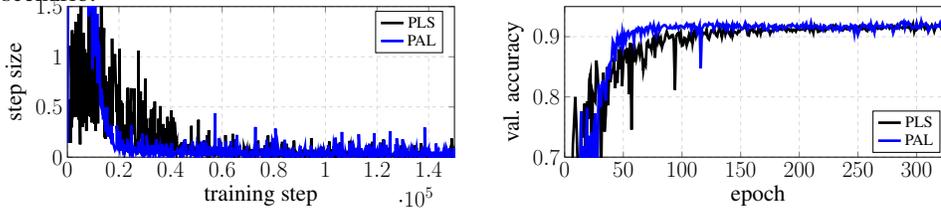

We used a empirically improved and only existing implementation of \textit{PLS} \cite{probabilisticLineSearch} for Tensorflow 1 \cite{probabilisticLineSearchImpl}. However, the sum of squared gradients has to be
derived manually for each layer, which is a considerable amount of work for modern architectures.
Consequently, we limit our comparison to a ResNet-32 trained on CIFAR-10. Figure \ref{plscomparison} shows that \textit{PAL} and \textit{PLS} perform similarly in this scenario. Default hyperparameters were used.

\FloatBarrier
\subsection{Further experimental design details}
\label{sec:further_experiment_information}
\subsubsection{Training Procedure}
On CIFAR-10 and CIFAR-100 we trained 150k steps. On Imagenet each network was trained for 500k steps. We performed a piecewise constant learning rate decay by dividing the learning rate by 10 at 50\% and 75\% of the steps.\\
The training set to evaluation set split was 45k to 15k for CIFAR-10 and CIFAR-100.
At the time of writing, the default Tensorflow classes do not support the reuse of the same randomly sampled numbers for multiple inferences, therefore, we implemented and used our own Dropout \cite{Dropout} layer.\\
To get a fair comparison of the optimizers capabilities, we compare on the training loss, the validation accuracy and the test accuracy metrics. For all metrics we provide the median and the quartiles to analyze the hyperparameter sensitivity.
For each hyperparameter combination we averaged our results over 3 runs using the seeds 1, 2 and 3 for reproducibility. All in all, we trained over 4500 networks with Tensorflow 1.15 \cite{Tensorflow} on Nvidia Geforce GTX 1080 TI graphic cards. 

\subsubsection{Data Augmentation}
\noindent On CIFAR-10  we performed the following augmentations \cite{resnet}:\\
4 pixel padding and cropping, horizontal image flipping with probability 0.5. \\
On Imagenet we applied an initial random crop to 224x224 pixels. In addition, we applied lighting as described in \cite{alexnet}.
For CIFAR-10, CIFAR-100 all images were normalized by channel-wise mean and variance. For the Tolstoi War and Peace dataset we omitted augmentation.
\pagebreak
\subsubsection{Hyperparameter grid search}
\noindent For our evaluation we used all combinations out of the following common used hyperparameters. The batch size is always 128 except for DenseNets trained with \textit{ALIG}, \textit{SGDHD} and \textit{COCOB} for which we encountered memory overflows and had to reduce the batch size to 100. Weight decay is always $10^{-4}$. \\ On Imagenet, such a comprehensive grid search was not possible. In this case we compared with the best hyperparameter combinations found on Cifar-100.\\
\textit{ADAM}:\\
\begin{tabular}{lll}
	 hyperparameter & symbol & values\\
	 \hline
	 learning rate  &    $\lambda$     & $\{ 1,0.1,0.01, 0.001, 0.0001 \}$ \\
	first momentum  & $\beta_1$  &            $\{ 0.9,0.95\}$           \\
	second momentum & $\beta_2$  &           $\{ 0.999\}$           \\
	    epsilon     & $\epsilon$ &       $\{ 1e-8\}$ \\
\end{tabular}

\noindent We did not vary the first or second momentum much, since \cite{adam} states that the values chosen are already good defaults.\\\\
\textit{SGD}:\\
\begin{tabular}{lll}
	\centering
	hyperparameter & symbol & values\\
	\hline
learning rate &$\lambda$ &$\{ 0.1, 0.01, 0.001, 0.0001\}$\\
momentum& $\alpha $ &$\{0.85, 0.9, 0.95\}$\\
\end{tabular}\\\\\\
\textit{RMSProp}:\\
\begin{tabular}{lll}
	\centering
	hyperparameter & symbol & values\\
	\hline
learning rate& $\lambda $&$ \{ 0.1, 0.01, 0.001, 0.0001\}$\\
discounting factor & $f $&$ \{0.9, 0.95\}$\\
epsilon &$\epsilon $&$ \{ 1e-8\}$\\
\end{tabular}\\\\\\
\textit{PAL}:\\
\begin{tabular}{lll}
	\centering
	hyperparameter & symbol & values\\
	\hline
measuring step size &$\mu $&$ \{10^0,10^{-0.5},10^{-0.1},10^{-0.15}\}$ \\
direction adaptation factor& $\beta $&$ \{ 0, 0.4\} $ \\
update step adaptation& $\alpha $&$ \{1,\frac{1}{0.8}\}$ \\
maximum step size& $s_{max} $&$ \{10^{0.5} (\approx 3.16)\}$ \\
\end{tabular}
\\In our implementation we worked with a inverse update step adaptation $\gamma=\frac{1}{\alpha}$.

\textit{SLS}:\\
\begin{tabular}{lll}
	hyperparameter   & symbol      & values                            \\ \hline
	initial step size & $\mu$       & $\{ 0.1,1\}$ \\
	step size decay   & $\beta$     & $\{ 0.9,0.99\}$                   \\
	step size reset   & $\gamma$    & $\{ 2.0,2.5\}$                      \\
	Armijo constant   & $c$         & $\{ 0.1,0.01\}$                       \\
	maximum step size & $\mu_{max}$ & $\{ 10.0\}$                       \\
\end{tabular}

\textit{ALIG}:\\
\begin{tabular}{lll}
	\centering
	hyperparameter & symbol & values\\
	\hline
maximal learning rate& $\lambda $&$ \{10 ,1.0, 0.1, 0.01\}$\\
momentum & $\beta$&$ \{0.85, 0.9, 0.95\}$\\
\end{tabular}\\

\textit{COCOB}:\\
\begin{tabular}{lll}
	\centering
	hyperparameter & symbol & values\\
	\hline
restriction factor& $\alpha $&$ \{25,50,75,100,125,150,175,200\}$\\
\end{tabular}\\

\textit{SGDHD}:\\
\begin{tabular}{lll}
	\centering
	hyperparameter & symbol & values\\
	\hline
learning rate& $\lambda $&$ \{0.1,0.01,0.001\}$\\
hyper gradient learning rate & $\beta$&$ \{0.1, 0.01, 0.001, 0.0001\}$\\
\end{tabular}\\

\vfil
\FloatBarrier
\pagebreak
\subsection{Detailed numerical results}
\label{sec:exp_table}
			\tiny 
	\newcolumntype{L}[1]{>{\raggedright\arraybackslash}m{4.2cm}|}	
	\begin{longtable}{lll >{$}l<{$} >{$}l<{$} >{$}l<{$} >{$}l<{$} >{$}l<{$} >{$}l<{$}}	
			\label{tbl:metrics}\\
		\caption{Performance comparison of \pal, \textit{RMSProp}, \textit{ADAM}, \textit{COCOB}, \textit{SGDHD}, \textit{ALIG} and \textit{SGD}. All hyperparameter combinations given in Appendix \ref{sec:further_experiment_information} were evaluated for each architecture. Results are averaged over 3 runs starting from different random seeds, except for training on ImageNet, for which results were not averaged. Note that tests on Imagenet were performed with the best hyperparameters found on CIFAR-100 to test the transferability of hyperparameters. Medians an Quartiles describe the distribution of results over reasonable hyper-parameter ranges.}\\ 
		\toprule
		dataset  & network      & optimizer         &\multicolumn{2}{l}{\text{training loss}} & \multicolumn{2}{l}{\text{validation accuracy}} & \multicolumn{2}{l}{\text{test accuracy}} \\
		&       &          &\text{min} &\text{median; p25; p75}  &\text{max} &\text{median; p25; p75}  &\text{max} &\text{median; p25; p75}  \\
		
		\midrule

		CIFAR-10 & EfficientNet & COCOB  & 0.659 & 0.824;0.739;0.855 &  0.857  &  0.837;0.832;0.845  &  0.843  &  0.824;0.818;0.832 \\ 
		 
		&  & ALIG  & 0.279 & 0.89;0.464;1.911 &  0.906  &  0.805;0.451;0.895  &  0.893  &  0.757;0.297;0.878 \\ 
		 
		&  & SGDHD  & 2.002 & 6.239;4.357;7.803 &  0.834  &  0.657;0.18;0.74  &  0.828  &  0.647;0.179;0.731 \\ 
		 
		&  & SLS  & 2.837 & 5.596;4.681;6.292 &  0.653  &  0.357;0.211;0.443  &  0.643  &  0.357;0.216;0.442 \\ 
		 
		&  & RMSP  & 0.154 & 0.637;0.333;1.261 &  \mathbf{0.93}  &  0.864;0.658;0.902  &  0.919  &  0.854;0.648;0.889 \\ 
		 
		&  & ADAM  & 0.155 & 0.818;0.292;2.275 &  0.926  &  0.841;0.211;0.907  &  0.919  &  0.83;0.1;0.896 \\ 
		 
		&  & SGD  & 0.165 & 2.287;0.343;4.221 &  \mathbf{0.93}  &  0.872;0.794;0.915  &  \mathbf{0.921}  &  0.862;0.784;0.906 \\ 
		 
		&  & PAL  & \mathbf{0.137} & \mathbf{0.244};0.186;0.388 &  0.927  &  \mathbf{0.912};0.906;0.921  &  0.916  &  \mathbf{0.902};0.889;0.908 \\ 
 
		\bottomrule
		CIFAR-10 & MobileNetV2 & COCOB  & 0.232 & \mathbf{0.282};0.257;0.295 &  0.879  &  0.87;0.866;0.876  &  0.865  &  0.852;0.848;0.865 \\ 
		
		&  & ALIG  & 0.183 & 0.938;0.347;1.926 &  0.914  &  0.695;0.233;0.888  &  0.897  &  0.528;0.1;0.851 \\ 
		
		&  & SGDHD  & 0.698 & 2.234;1.835;4.366 &  0.886  &  0.75;0.298;0.807  &  0.877  &  0.737;0.295;0.791 \\ 
		
		&  & SLS  & 1.387 & 2.462;2.011;2.584 &  0.667  &  0.443;0.407;0.504  &  0.595  &  0.4;0.343;0.437 \\ 
		
		&  & RMSP  & \mathbf{0.085} & 0.493;0.337;0.918 &  0.938  &  0.872;0.675;0.895  &  0.929  &  0.865;0.664;0.882 \\ 
		
		&  & ADAM  & 0.095 & 0.477;0.314;1.861 &  0.939  &  0.874;0.309;0.896  &  0.93  &  0.864;0.289;0.886 \\ 
		
		&  & SGD  & 0.149 & 0.878;0.204;1.552 &  \mathbf{0.947}  &  \mathbf{0.907};0.87;0.933  &  \mathbf{0.94}  &  \mathbf{0.899};0.859;0.925 \\ 
		
		&  & PAL  & 0.15 & 0.377;0.205;0.531 &  0.92  &  0.905;0.896;0.91  &  0.905  &  0.886;0.877;0.896 \\ 
		\bottomrule
		CIFAR-10 & DenseNet40 & COCOB  & 0.228 & 0.234;0.23;0.24 &  0.907  &  0.903;0.901;0.904  &  0.894  &  0.889;0.885;0.892 \\ 
		 
		&  & ALIG  & 0.188 & 0.604;0.227;2.903 &  0.918  &  0.848;0.438;0.902  &  0.902  &  0.784;0.336;0.875 \\ 
		 
		&  & SGDHD  & 1.094 & 2.279;1.349;2.908 &  0.775  &  0.341;0.099;0.696  &  0.762  &  0.1;0.1;0.26 \\ 
		 
		&  & SLS  & \mathbf{0.065} & \mathbf{0.115};0.104;0.189 &  0.91  &  0.904;0.897;0.905  &  0.901  &  \mathbf{0.893};0.89;0.897 \\ 
		 
		&  & RMSP  & 0.147 & 0.398;0.256;0.915 &  0.927  &  0.879;0.737;0.915  &  0.92  &  0.867;0.717;0.909 \\ 
		 
		&  & ADAM  & 0.138 & 0.749;0.274;1.028 &  0.922  &  0.777;0.611;0.91  &  0.913  &  0.806;0.605;0.907 \\ 
		 
		&  & SGD  & 0.147 & 0.794;0.396;1.746 &  \mathbf{0.932}  &  0.855;0.537;0.914  &  \mathbf{0.93}  &  0.847;0.528;0.91 \\ 
		 
		&  & PAL  & 0.099 & 0.217;0.165;0.343 &  0.925  &  \mathbf{0.907};0.894;0.919  &  0.916  &  0.882;0.861;0.9 \\

		\bottomrule
		
		CIFAR-10 & ResNet32 & COCOB  & 0.125 & 0.128;0.127;0.129 &  0.888  &  0.886;0.885;0.887  &  0.878  &  0.872;0.871;0.874 \\ 
		 
		&  & ALIG  & 0.122 & 0.658;0.279;1.485 &  0.892  &  0.815;0.47;0.881  &  0.866  &  0.71;0.367;0.852 \\ 
		 
		&  & SGDHD  & 0.35 & 0.464;0.413;0.701 &  0.864  &  0.835;0.791;0.843  &  0.837  &  0.796;0.766;0.827 \\ 
		 
		&  & SLS  & \mathbf{0.005} & \mathbf{0.006};0.005;0.827 &  0.871  &  0.856;0.758;0.869  &  0.846  &  0.824;0.657;0.839 \\ 
		 
		&  & RMSP  & 0.105 & 0.199;0.129;0.498 &  0.922  &  0.884;0.804;0.904  &  0.915  &  0.877;0.792;0.896 \\ 
		 
		&  & ADAM  & 0.105 & 0.332;0.133;1.004 &  0.917  &  0.875;0.677;0.881  &  0.914  &  0.868;0.654;0.873 \\ 
		 
		&  & SGD  & 0.098 & 0.131;0.118;0.322 &  \mathbf{0.939}  &  \mathbf{0.899};0.85;0.924  &  \mathbf{0.933}  &  \mathbf{0.893};0.838;0.92 \\ 
		 
		&  & PAL  & 0.05 & 0.105;0.075;0.195 &  0.921  &  0.893;0.887;0.906  &  0.903  &  0.88;0.849;0.888 \\

		\bottomrule

		CIFAR-100 & DenseNet40 & COCOB  & 0.739 & 0.761;0.75;0.772 &  0.642  &  0.633;0.631;0.637  &  0.646  &  0.632;0.629;0.637 \\ 
		 
		&  & ALIG  & 0.488 & 2.125;0.988;3.128 &  0.637  &  0.508;0.391;0.623  &  0.616  &  0.48;0.264;0.605 \\ 
		 
		&  & SGDHD  & 1.78 & 2.6;2.179;3.465 &  0.566  &  0.418;0.274;0.504  &  0.55  &  0.296;0.159;0.497 \\ 
		 
		&  & SLS  & 1.367 & 1.908;1.446;1.96 &  0.719  &  0.593;0.572;0.698  &  0.612  &  0.479;0.422;0.554 \\ 
		 
		&  & RMSP  & 0.348 & 1.238;0.78;1.972 &  0.716  &  0.583;0.481;0.634  &  0.712  &  0.588;0.482;0.631 \\ 
		 
		&  & ADAM  & 0.326 & 1.114;0.859;3.53 &  0.715  &  0.601;0.165;0.637  &  0.712  &  0.599;0.226;0.641 \\ 
		 
		&  & SGD  & 0.376 & 0.713;0.431;2.154 &  \mathbf{0.75}  &  0.633;0.489;0.709  &  \mathbf{0.753}  &  0.634;0.498;0.708 \\ 
		 
		&  & PAL  & \mathbf{0.275} & \mathbf{0.376};0.312;0.459 &  0.73  &  \mathbf{0.686};0.66;0.705  &  0.717  &  \mathbf{0.676};0.642;0.695 \\

		\bottomrule
		CIFAR-100 & EfficientNet & COCOB  & 0.802 & 0.817;0.807;0.822 &  0.594  &  0.583;0.581;0.59  &  0.596  &  0.582;0.58;0.588 \\ 
		 
		&  & ALIG  & 0.57 & 2.4;0.995;4.085 &  0.612  &  0.494;0.169;0.6  &  0.599  &  0.458;0.115;0.597 \\ 
		 
		&  & SGDHD  & 3.545 & 6.528;5.519;8.917 &  0.529  &  0.337;0.178;0.463  &  0.513  &  0.342;0.179;0.468 \\ 
		 
		&  & SLS  & 3.731 & 6.713;6.348;6.857 &  0.474  &  0.212;0.208;0.227  &  0.375  &  0.203;0.149;0.208 \\ 
		 
		&  & RMSP  & 0.422 & 1.823;1.253;2.968 &  0.675  &  0.517;0.383;0.588  &  0.678  &  0.521;0.382;0.59 \\ 
		 
		&  & ADAM  & 0.45 & 1.394;1.312;4.606 &  0.684  &  0.518;0.025;0.619  &  0.684  &  0.524;0.01;0.621 \\ 
		 
		&  & SGD  & 0.42 & 2.44;0.633;5.214 &  \mathbf{0.712}  &  0.579;0.473;0.661  &  \mathbf{0.709}  &  0.579;0.476;0.658 \\ 
		 
		&  & PAL  & \mathbf{0.372} & \mathbf{0.471};0.409;0.772 &  0.693  &  \mathbf{0.666};0.638;0.676  &  0.69  &  \mathbf{0.664};0.63;0.671 \\

		\bottomrule
		CIFAR-100 & MobileNetV2 & COCOB  & 0.486 & \mathbf{0.513};0.492;0.536 &  0.644  &  0.63;0.626;0.637  &  0.644  &  0.63;0.623;0.638 \\ 
		 
		&  & ALIG  & 0.323 & 2.396;0.817;4.247 &  0.661  &  0.41;0.034;0.623  &  0.652  &  0.229;0.01;0.602 \\ 
		 
		&  & SGDHD  & 1.485 & 3.307;2.425;7.002 &  0.593  &  0.476;0.39;0.545  &  0.589  &  0.456;0.385;0.525 \\ 
		 
		&  & SLS  & 3.857 & 5.086;5.031;5.64 &  0.332  &  0.2;0.099;0.203  &  0.197  &  0.081;0.052;0.126 \\ 
		 
		&  & RMSP  & 0.198 & 1.518;0.718;3.368 &  0.728  &  0.593;0.43;0.635  &  0.727  &  0.593;0.431;0.634 \\ 
		 
		&  & ADAM  & 0.218 & 1.873;0.776;4.524 &  0.729  &  0.528;0.025;0.593  &  0.729  &  0.533;0.02;0.595 \\ 
		 
		&  & SGD  & 0.4 & 0.974;0.473;2.151 &  \mathbf{0.733}  &  0.657;0.57;0.7  &  \mathbf{0.736}  &  0.659;0.573;0.701 \\ 
		 
		&  & PAL  & \mathbf{0.181} & 0.602;0.314;1.571 &  0.726  &  \mathbf{0.666};0.574;0.689  &  0.722  &  \mathbf{0.664};0.509;0.681 \\

		\bottomrule
		CIFAR-100 & ResNet32 & COCOB  & 0.498 & 0.569;0.524;0.673 &  0.609  &  0.608;0.607;0.608  &  0.605  &  0.602;0.599;0.604 \\ 
		 
		&  & ALIG  & 0.537 & 1.932;0.995;3.572 &  0.597  &  0.491;0.19;0.58  &  0.587  &  0.414;0.144;0.549 \\ 
		 
		&  & SGDHD  & 0.881 & 1.359;1.06;1.772 &  0.601  &  0.539;0.472;0.586  &  0.599  &  0.517;0.431;0.571 \\ 
		 
		&  & SLS  & 2.62 & 2.808;2.78;2.82 &  0.399  &  0.388;0.384;0.392  &  0.363  &  0.305;0.274;0.33 \\ 
		 
		&  & RMSP  & 0.519 & 1.019;0.807;2.083 &  0.661  &  0.599;0.455;0.651  &  0.656  &  0.603;0.455;0.65 \\ 
		 
		&  & ADAM  & 0.402 & 1.772;0.768;3.038 &  0.659  &  0.513;0.262;0.564  &  0.658  &  0.519;0.255;0.567 \\ 
		 
		&  & SGD  & 0.375 & \mathbf{0.474};0.4;1.522 &  \mathbf{0.697}  &  0.614;0.494;0.672  &  \mathbf{0.694}  &  0.616;0.502;0.667 \\ 
		 
		&  & PAL  & \mathbf{0.339} & 0.485;0.369;1.424 &  0.662  &  \mathbf{0.636};0.546;0.652  &  0.663  &  \mathbf{0.621};0.512;0.647 \\

		\bottomrule
		\pagebreak
		\bottomrule
TOLSTOI& RNN & COCOB  & 1.506 & 1.56;1.533;1.593 &  0.589  &  0.58;0.573;0.584  &  0.582  &  0.572;0.566;0.577 \\ 
 
&  & ALIG  & 1.501 & 1.562;1.528;1.766 &  0.591  &  0.579;0.523;0.586  &  0.584  &  0.571;0.513;0.577 \\ 
 
&  & SGDHD  & 2.282 & 2.433;2.379;2.445 &  0.375  &  0.338;0.336;0.348  &  0.369  &  0.334;0.332;0.344 \\ 
 
&  & SLS  & 3.128 & 3.149;3.136;3.156 &  0.169  &  0.159;0.158;0.165  &  0.168  &  0.158;0.157;0.164 \\ 
 
&  & RMSP  & \mathbf{1.475} & \mathbf{1.509};1.492;1.556 &  \mathbf{0.599}  &  \mathbf{0.591};0.579;0.595  &  \mathbf{0.592}  &  \mathbf{0.583};0.572;0.587 \\ 
 
&  & ADAM  & 1.516 & 1.655;1.596;1.681 &  0.588  &  0.567;0.55;0.578  &  0.581  &  0.561;0.543;0.571 \\ 
 
&  & SGD  & 1.496 & 1.872;1.56;2.675 &  0.594  &  0.483;0.278;0.573  &  0.587  &  0.476;0.275;0.566 \\ 
 
&  & PAL  & 1.528 & 1.569;1.547;1.588 &  0.587  &  0.581;0.577;0.586  &  0.579  &  0.571;0.556;0.575 \\ 
 
		\bottomrule
		
		ImageNet & ResNet50 & COCOB  & 2.0 & - &  0.51  & - &  0.518  &  -\\ 
		 
		&  & ALIG  & 1.854 & - &  0.539  &  -  &  0.512  &  - \\ 
		 
		&  & SGDHD  & 2.742 & - &  0.498  &  -  &  0.495  &  - \\
		
	    &  & RMSP  & 9.485 & -&  0.286  & -  &  0.28  &  - \\ 
		 
		&  & ADAM  & 1.863 & - &  0.562  &  -  &  0.559  &  - \\ 
		 
		&  & SLS  & 3.808 &- &  0.286  & -  &  0.069  &  - \\ 
		 
		&  & SGD  & 1.123 & - &  \mathbf{0.656}  &  -  &  \mathbf{0.65}  &  - \\ 
		 
		&  & PAL  & \mathbf{0.773} & - &  0.608  &  -  &  0.608  &  -\\ 
		\bottomrule
		ImageNet & DenseNet121 & COCOB  & 6.9 & - &  0.006  & - &  0.006  &  - \\ 
		 
		&  & ALIG  & 2.142 &- &  0.533  & - &  0.512  &  -\\ 
		 
		&  & SGDHD  & 2.939 & - &  0.343  & - &  0.362  & - \\ 
		&  & RMSP  & 6.901 & - &  0.0  &  -  &  0.0  &  -\\ 
		 
		&  & ADAM  & 6.901 & - &  0.001  &  -  &  0.0  &  - \\ 
		 
		&  & SLS  & 7.768 & - &  0.001  &  -  &  0.001  &  - \\ 
		 
		&  & SGD  & 3.308 & - &  0.458  &  -  &  0.452  & - \\ 
		 
		&  & PAL  & \mathbf{1.228} & - &  \mathbf{0.617}  &  -&  \mathbf{0.611}  &  - \\ 
		\bottomrule

	\end{longtable}
			\normalsize
\vfil
\section{Binary Line Search}
\label{sec:binary_line_search}
\noindent The optimal binary line search we compared \pal\ against. Since the line decreases in negative gradient direction, at first a extrapolation phase performs as many steps forward as the loss does not increase. Afterwards a binary search is performed. This approach is valid if the underlying line is convex. For simple readability we chose Python 3.6 syntax.
\lstset{breaklines=true, breakatwhitespace=true,basicstyle=\footnotesize\ttfamily}
\lstset{numbers=right, numberstyle=\scriptsize}
\begin{lstlisting}[language=Python, linewidth=0.95\linewidth]
Input:max_num_of_search_steps
def binary_line_search(last_loss, step, counter, is_extrapolate):
    if counter == max_num_of_search_steps:
        return last_loss
    counter += 1
    if is_extrapolate:
        current_loss = do_step_on_line(step)
        if current_loss < last_loss:
            return binary_line_search(current_loss, step,counter, is_extrapolate)
        else:
            is_extrapolate = False
            do_step_on_line(-step,get_loss=False) 
    if not is_extrapolate: 
        loss_right = do_step_on_line(0.5*step, True)
        if loss_right < last_loss:
            return binary_line_search(loss_right, 0.5*step,counter, is_extrapolate)
        loss_left = do_step_on_line(-1*step, True)
        if loss_left < last_loss:
            return binary_line_search(loss_left, 0.5*step,counter, is_extrapolate)
        do_step_on_line(0.5*step,get_loss=False) 
        if loss_right >= last_loss and loss_left >= last_loss:
            return binary_line_search(loss_left, 0.5*step, counter, is_extrapolate)
	else:	
		# this state is not possible
\end{lstlisting}
\vfil

\end{document}